\documentclass[lettersize,journal]{IEEEtran}
\usepackage{amsmath,amsfonts}
\usepackage{algorithmic}
\usepackage{algorithm}
\usepackage{array}
\usepackage{textcomp}
\usepackage{stfloats}
\usepackage{url}
\usepackage{verbatim}
\usepackage{graphicx}
\usepackage{cite}
\usepackage{colortbl}
\usepackage{etoolbox}
\usepackage{multirow}
\usepackage{tikz}
\usepackage[colorlinks,
linkcolor=blue,
anchorcolor=blue,
citecolor=blue]{hyperref}
\usepackage{amssymb}
\usepackage{booktabs}
\usepackage{xspace}

\usepackage[capitalize]{cleveref}
\crefname{section}{Sec.}{Secs.}
\Crefname{section}{Section}{Sections}
\Crefname{table}{Table}{Tables}
\crefname{table}{Tab.}{Tabs.}

\definecolor{cdark}{RGB}{0, 0, 0}
\definecolor{cpurblue}{RGB}{194, 207, 242}
\definecolor{cblue}{RGB}{66, 120, 245}
\definecolor{cgrass}{RGB}{123, 252, 3}
\definecolor{ccyan}{RGB}{8, 189, 171}
\definecolor{cbrown}{RGB}{161, 100, 56}
\definecolor{cgreen}{RGB}{26, 110, 53}
\definecolor{csky}{RGB}{148, 250, 255}
\definecolor{cyellow}{RGB}{237, 187, 36}
\definecolor{corange}{RGB}{245, 130, 69}
\definecolor{cgrey}{RGB}{157, 163, 163}
\definecolor{crose}{RGB}{235, 101, 157}
\definecolor{cpink}{RGB}{255, 212, 212}
\definecolor{clime}{RGB}{191, 255, 0}
\definecolor{cgold}{RGB}{255, 223, 0}
\definecolor{colivedrab}{RGB}{107, 142, 35}
\definecolor{cviolet}{RGB}{143, 0, 255}
\definecolor{cred}{RGB}{255, 0, 0}
\definecolor{cblack}{RGB}{0, 0, 0}
\definecolor{cdarkred}{RGB}{139, 0, 0}
\definecolor{cslateblue}{RGB}{106, 90,  205}
\definecolor{cfirebrick}{RGB}{178, 34,  34}

\newcommand{\csticksolid}[1]{
    \tikz[baseline]{\draw[color=#1, line width=1.5mm] (0,.5ex)--++(.6,0);}}

\newcommand{\cstickdashed}[1]{
    \tikz[baseline]{
        \draw[color=#1, dashed, line width=1.2mm] (0,.5ex)--++(.6,0);
        \node[color=#1, inner sep=0pt, scale=1.4] at (.263,.5ex) {$\blacktriangle$};
    }}
\newcommand{\cstickdotted}[1]{
    \tikz[baseline]{
        \draw[color=#1, dashed, line width=1.2mm] (0,.5ex)--++(.6,0);
        \node[color=#1, inner sep=0pt, scale=1.1] at (.26,.5ex) {$\blacksquare$};
    }}

\newcommand{\textitgray}[1]{\textit{\textcolor{gray}{#1}}}

\definecolor{tabfirst}{rgb}{1, 0.7, 0.7}
\definecolor{tabsecond}{rgb}{1, 0.85, 0.7}
\definecolor{tabthird}{rgb}{1, 1, 0.7}

\makeatletter
\DeclareRobustCommand\onedot{\futurelet\@let@token\@onedot}
\def\@onedot{\ifx\@let@token.\else.\null\fi\xspace}

\def\etal{\emph{et al}\onedot}
\let\@authorsaddresses\@empty
\makeatother

\newcommand{\PAR}[1]{\vspace{0.1cm}\noindent{\bf #1} }

\begin{document}

\title{SSMB: Self-Supervised Local Feature Detection under Motion Blur}

\author{Zhenjun Zhao, Fabio Bellavia, Wenting Wang, Fan Zhu, Jiajun Wu,\\Suryansh Kumar, Mingqiang Wei, Haoang Li, Javier Civera
\thanks{Z. Zhao and J. Civera are with University of Zaragoza, Zaragoza, Spain.}
\thanks{F. Bellavia is with University of Palermo, Palermo, Italy.}
\thanks{W. Wang is with The Chinese University of Hong Kong, Hong Kong, China.}
\thanks{F. Zhu is with Tohoku University, Sendai, Japan.}
\thanks{J. Wu is with Central South University, Changsha, China.}
\thanks{S. Kumar is with Texas A\&M University, College Station, TX, USA.}
\thanks{M. Wei is with Nanjing University of Aeronautics and Astronautics, Nanjing, China.}
\thanks{H. Li is with The Hong Kong University of Science and Technology (Guangzhou), Guangzhou, China.}}

\maketitle

\begin{abstract}
Keypoint detection under motion blur remains a significant challenge, as blur distorts local image structure and degrades the repeatability of feature localization.
Existing approaches either rely on computationally expensive deblur-then-detect pipelines that may introduce restoration artifacts, or learn to regress the image positions of handcrafted keypoints extracted on sharp images, which reflects the assumptions of the handcrafted detector rather than what is truly repeatable under blur.
We present SSMB, a deblur-free, self-supervised keypoint detector for motion-blurred images that requires neither handcrafted detectors nor external pseudo-labels.
SSMB introduces the Local Discriminability Enhancement (LDE) module, which restores fine-grained local discriminability after global feature mixing.
Training is performed in two stages.
First, geometric pretraining on synthetic shapes bootstraps spatially discriminative keypoint detection without any external detector, just from the rendered geometry.
Second, blur-aware training on real sharp-blur image pairs learns blur-invariant detection through a multi-component self-supervised objective that enforces cross-domain consistency, geometric alignment, and spatial coverage.
Extensive evaluations on keypoint detection, image matching, relative pose estimation, and visual localization under motion blur demonstrate that SSMB establishes a new state-of-the-art among sparse keypoint detectors, consistently outperforming both supervised and self-supervised baselines across all tasks.
Code, models, and datasets will be publicly available upon paper acceptance.
\end{abstract}

\section{Introduction}\label{sec:introduction}

Accurately detecting salient keypoints across images is a fundamental component in many computer vision applications, including Simultaneous Localization and Mapping (SLAM), Structure-from-Motion (SfM), camera calibration, image retrieval, and visual localization~\cite{campos2021orb,li2023hong,Sattler2018CVPR,meng2026dream,schonberger2016structure,li2025slam,sattler2012image,zhu2026mygo,toft2020long,gu2026ulf,sarlin2019coarse,zhao2026advances}.
An effective keypoint detector should produce features that are well distributed across the image, highly repeatable across viewpoints and time, and robust to photometric and geometric variation.
While remarkable progress has been made on sharp images~\cite{LoweDavid2004DistinctiveIF,DeTone2018SuperPointSI,Laguna2019KeyNetKD,Dusmanu2019D2NetAT,Revaud2019R2D2RA,tyszkiewicz2020disk,pakulev2023ness,Zhao2023ALIKED,gleize2023silk,bellavia2024image,edstedt2024dedodev2}, motion blur remains a major and largely overlooked challenge.
Arising from camera shake, object motion, or long exposure times in low-light conditions, motion blur smears local image structure across spatial regions.
This severely degrades the repeatability of keypoint detection, compromising the feature correspondences required by downstream vision tasks.

A natural approach to addressing motion blur is the \emph{deblur-then-detect} pipeline, which first restores a sharp image with a deblurring network and then applies any existing feature detector.
However, deblurring networks~\cite{Nah2017DeepMC,Kupyn2019DeblurGANv2D,Tao2018ScaleRecurrentNF} are computationally demanding and rarely run in real time.
Moreover, their restored images often contain artifacts, particularly under severe blur, which propagate into the detection stage and degrade the reliability of the extracted features.

In contrast, a deblur-free detector that operates directly on blurred images avoids the computational overhead of image restoration while eliminating artifact propagation, providing a more efficient and robust solution.
In this paper, we focus on this deblur-free paradigm.
A recent step in this direction is BALF~\cite{zhao2024balf}, which introduces a Multi-Layer-Perceptron (MLP)-based detector trained directly on blurred images with real-time inference capabilities.
However, BALF is trained to regress the SIFT~\cite{LoweDavid2004DistinctiveIF} keypoints detected on the corresponding sharp images, which fundamentally limits its ability to learn general blur-invariant features. 
Since SIFT is handcrafted for sharp images rather than motion-blurred ones, the learned detector inevitably reproduces its detection preferences instead of discovering the structures that are genuinely repeatable under blur.
More generally, supervising with pseudo-labels, whether from SIFT or any other handcrafted detector, constrains the learned representation to imitate the behavior of the chosen detector rather than directly learning the cues that characterize repeatable keypoints in blurred images.

\begin{figure*}[t]
    \scriptsize
    \centering
    \hspace{3mm}
    \begin{minipage}[c]{0.325\linewidth}
        \centering
        \begin{minipage}[t]{\linewidth}
            \centering
            \includegraphics[width=\linewidth]{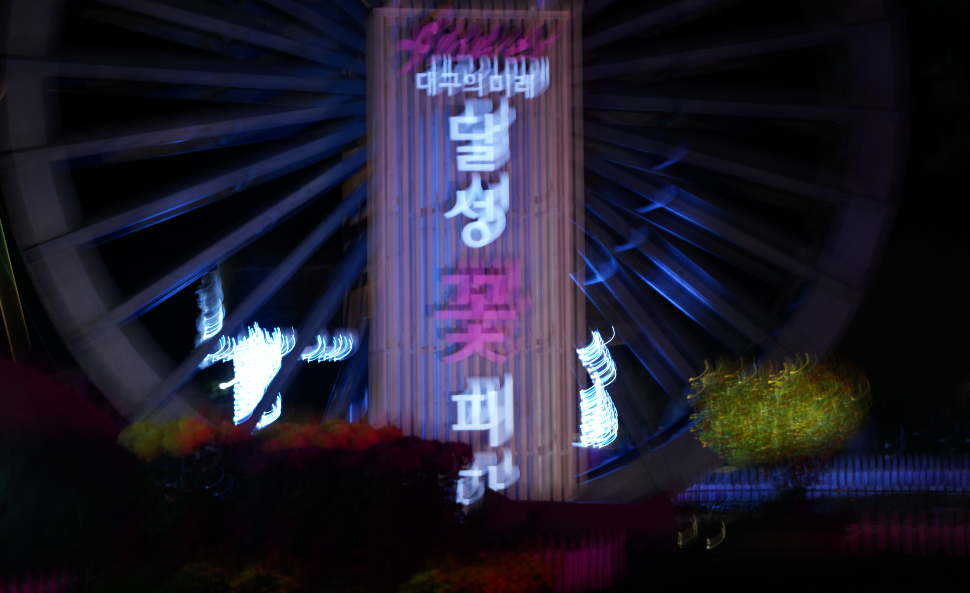}\\[1mm]
            Input
        \end{minipage}\\[3mm]
        \begin{minipage}[t]{\linewidth}
            \centering
            \includegraphics[width=\linewidth]{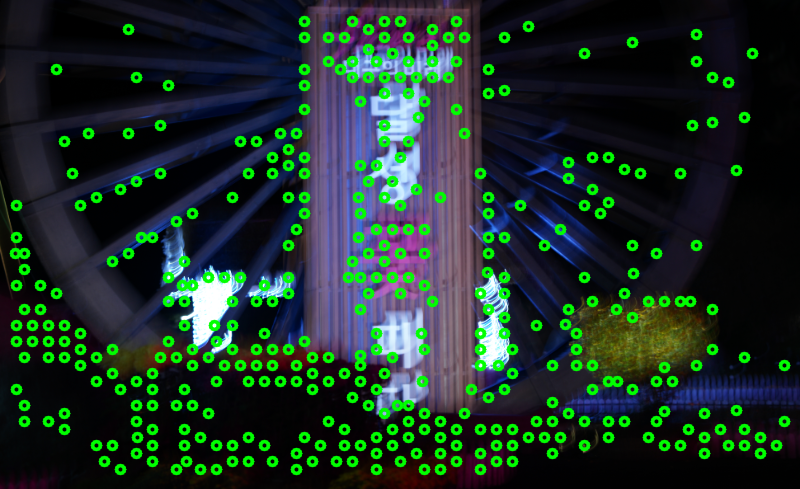}\\[1mm]
            \textbf{SSMB (Ours)}
        \end{minipage}
    \end{minipage}
    \begin{minipage}[c]{0.625\linewidth}
        \centering
        \includegraphics[width=\linewidth]{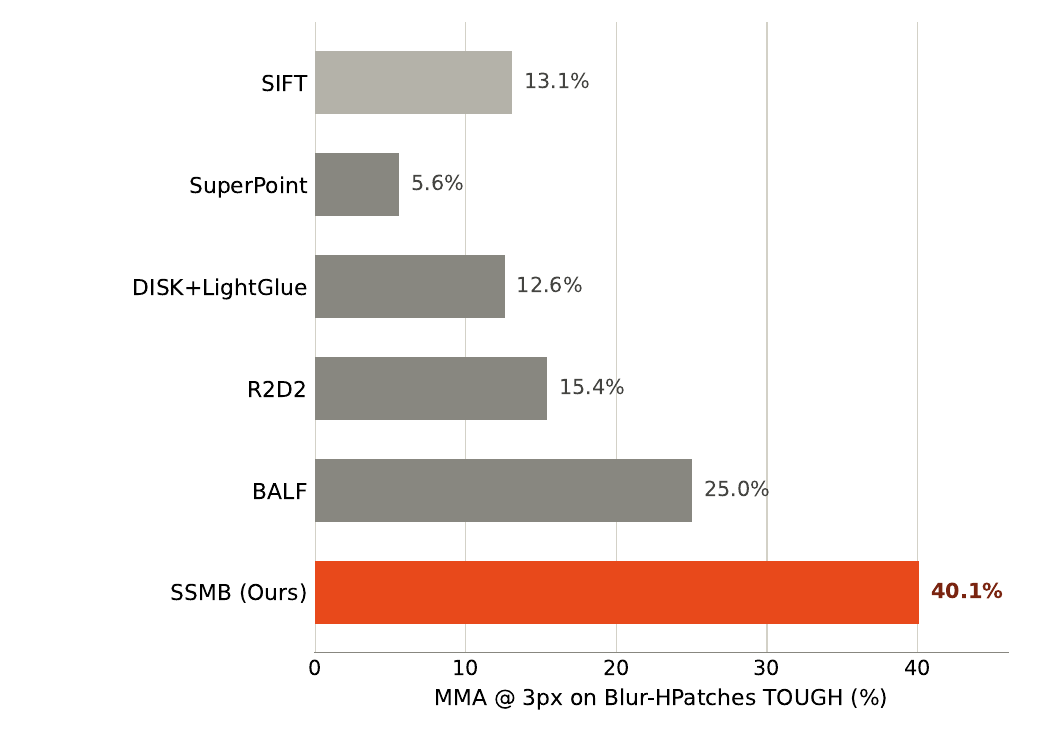}
    \end{minipage}
    \caption{\textbf{SSMB under motion blur.}
    \emph{Left:} a real-world blurred image from RealBlur~\cite{Rim2020RealWorldBD} (top) and the keypoints (green circles) detected by SSMB on it (bottom), without any deblurring preprocessing.
    \emph{Right:} image matching accuracy (Mean Matching Accuracy (MMA)~\cite{Dusmanu2019D2NetAT} at a 3-pixel threshold) on the Blur-HPatches \textsc{Tough} split~\cite{zhao2024balf} under motion blur.
    SSMB, our self-supervised sparse detector trained without handcrafted or external supervision, substantially outperforms all sparse baselines.}
    \label{fig:teaser}
\end{figure*}

In this paper, we propose \textbf{SSMB}, a deblur-free keypoint detector for motion-blurred images that requires neither handcrafted detectors nor external pseudo-labels.
Since SSMB must learn on its own which local structures are relevant, without any external signal to guide it, we introduce a novel \textbf{Local Discriminability Enhancement (LDE)} module, which explicitly preserves and reinforces these local cues throughout the network, enabling effective self-supervised learning in the presence of blur.
Additionally, we propose a \textbf{two-stage training pipeline}.
At the first stage, that we denote as \emph{geometric pretraining}, SSMB learns to detect spatially discriminative keypoints from synthetic images of simple geometric forms, in which reliable corner labels are derived directly from the rendered geometry without any external detector.
At the second stage, that we denote as \emph{blur-aware training}, SSMB is trained on real sharp-blur pairs from the GoPro dataset~\cite{Nah2017DeepMC} using a multi-component self-supervised objective. This objective aligns detections across the sharp and blurred domains, anchors them to geometrically consistent locations through online homographic adaptation~\cite{DeTone2018SuperPointSI}, and promotes a well-distributed set of keypoints across the image, preventing the network from collapsing to a small set of dominant responses.
In our experiments we observed that, without this spatial spreading term, self-supervised training degenerates into a trivial solution in which detection concentrates on a handful of salient locations, leaving most of the image without any meaningful response.
SSMB reliably detects keypoints on a real-world blurred image and achieves the highest image matching accuracy among sparse methods under motion blur, as illustrated in \cref{fig:teaser}.

In summary, our main contributions are as follows:
\begin{itemize}
    \item We introduce the Local Discriminability Enhancement (LDE) module, which restores fine-grained local discriminability after global feature mixing, a necessary condition for self-supervised training to succeed under blur.
    \item We propose a two-stage training pipeline with geometric pretraining on synthetic data followed by blur-aware training on real sharp-blur pairs, and show that geometric pretraining is an essential prerequisite for effectively bootstrapping self-supervised learning under blur.
    \item We design a multi-component self-supervised loss combining homographic adaptation, blur consistency, position consistency, and spatial diversity terms, and identify spatial diversity as the key term that prevents the network from collapsing to a degenerate solution.
\end{itemize}

Extensive evaluations on keypoint detection, image matching, relative pose estimation, and visual localization under motion blur demonstrate that SSMB establishes a new state of the art among sparse keypoint detectors, consistently outperforming both supervised and self-supervised sparse baselines, and even surpassing detector-free matching methods at low pixel thresholds.
\section{Related Work}\label{sec:related_work}

\subsection{Hand-crafted Local Feature Detection}

Hand-crafted keypoint detectors identify salient image structures using operators that were manually designed to capture certain local image statistics.
The Harris corner detector~\cite{Harris1988ACC} identifies corners from first-order image derivatives, while the Hessian detector~\cite{beaudet1978rotational} detects blob-like structures from second-order derivatives.
SIFT~\cite{LoweDavid2004DistinctiveIF} became the most widely adopted due to its robustness to scale and rotation, and it has long served as a source of pseudo-labels for supervised learning-based keypoint detectors.

\subsection{Learning-Based Local Feature Detection}

Learning-based methods overcome the limitations of hand-crafted operators by training neural networks to detect repeatable keypoints directly from data.
SuperPoint~\cite{DeTone2018SuperPointSI} introduces a self-supervised framework that bootstraps keypoint labels from synthetic geometry via homographic adaptation on real images, producing a joint detector and descriptor without manual annotation.
Subsequent methods explored increasingly powerful network architectures and learning objectives. Key.Net~\cite{Laguna2019KeyNetKD} combines hand-crafted filters with learned Convolutional Neural Network (CNN) features for multi-scale detection.
D2-Net~\cite{Dusmanu2019D2NetAT} presents a single CNN that jointly detects and describes features by identifying local maxima across both spatial and channel dimensions of deep feature maps.
R2D2~\cite{Revaud2019R2D2RA} simultaneously predicts keypoint locations and descriptors using dilated convolutions, training the detector to be both repeatable and reliable.
DISK~\cite{tyszkiewicz2020disk} trains an end-to-end detector and descriptor pipeline using reinforcement learning rewards derived from matching performance, bypassing the non-differentiability of sparse keypoint selection.
REKD~\cite{lee2022self} proposes a self-supervised rotation-equivariant detector.
NeSS-ST~\cite{pakulev2023ness} combines hand-crafted keypoints with a learned neural stability score to select high-quality feature points.
ALIKED~\cite{Zhao2023ALIKED} proposes sparse deformable descriptors that learn geometrically adaptive supporting features for each keypoint, emphasizing precise keypoint localization accuracy.
DeDoDe v2~\cite{edstedt2024dedodev2} decouples detection from description for improved pose estimation.
XFeat~\cite{potje2024cvpr} prioritizes real-time efficiency with a lightweight architecture suitable for resource-limited devices.

\subsection{Detector-Free Matching Methods}

An alternative to sparse keypoint detection is dense or semi-dense feature matching, where correspondences are established directly between image pairs without explicit keypoint detection.
LoFTR~\cite{sun2021loftr} pioneered this paradigm by leveraging a transformer architecture~\cite{Vaswani2017AttentionIA} to compute dense correlation maps between image patches, enabling semi-dense matching even in weakly textured regions, where sparse detectors typically fail.
Subsequent methods, such as MatchFormer~\cite{wang2022matchformer} and ASpanFormer~\cite{chen2022aspanformer}, improve upon LoFTR with interleaved attention and adaptive span mechanisms, respectively.
More recently, RoMa~\cite{edstedt2024roma} combines frozen DINOv2~\cite{oquab2023dinov2} features with CNN features and a transformer-based matching decoder to achieve dense correspondences, while RoMa v2~\cite{edstedt2025roma} improves both the matching architecture and training distribution to establish the current state of the art in dense matching.
Despite their impressive performance, detector-free methods have important practical limitations.
Since correspondences are computed jointly for each image pair, they cannot produce reusable keypoints or descriptors, making them poorly suited to retrieval-based visual localization pipelines~\cite{sarlin2019coarse}, which are designed around pre-computed, reusable keypoint maps of the environment and become prohibitively expensive when every image pair must be matched from scratch.
Furthermore, their inference cost is also substantially higher than that of sparse detectors, as they require processing the full image pair at inference time.
Finally, although these methods significantly advanced the state of the art on standard benchmarks, neither sparse nor detector-free approaches are designed nor trained to handle motion blur, which distorts the local image structure on which both paradigms rely.

\subsection{Keypoint Detection on Motion-Blurred Images}

\PAR{Image deblurring.}
A common strategy for handling motion blur is to first restore a sharp image using a deblurring network and then apply an existing keypoint detector.
Early deep learning approaches estimate the blur kernels from the degraded image and restore the blur-less image through learned deconvolution~\cite{Sun2015LearningAC}. More recent methods address deblurring in an end-to-end manner, formulating it as an image-to-image translation problem.
Nah~\etal~\cite{Nah2017DeepMC} introduce a multi-scale CNN with a coarse-to-fine strategy for blind deblurring.
Kupyn~\etal propose DeblurGAN~\cite{Kupyn2018DeblurGANBM} and its follow-up DeblurGAN-v2~\cite{Kupyn2019DeblurGANv2D}, employing adversarial training to improve perceptual quality.
SRN-DeblurNet~\cite{Tao2018ScaleRecurrentNF} uses a scale-recurrent architecture for progressive multi-scale restoration.
While these methods produce visually plausible results in most cases, they introduce restoration artifacts, in particular under severe blur, and require significant computational resources that preclude real-time operation.

\PAR{Deblur-free detection.}
Rather than restoring the image before detection, a more principled approach is to design a detector that operates directly on blurred images, without any intermediate restoration.
To the best of our knowledge, BALF~\cite{zhao2024balf} is the first method to explicitly follow this paradigm, proposing a pure MLP-based architecture trained on sharp-blur image pairs with SIFT keypoints on sharp images as supervision.
Although BALF achieves real-time inference and strong performance on blurred benchmarks, its supervision is fundamentally tied to a hand-crafted detector.
Consequently, its learned representation is constrained to replicate SIFT's detections rather than discovering structures that are genuinely repeatable under blur.
In contrast, our SSMB removes the need for external supervision entirely by learning directly from geometric and cross-domain consistency constraints in a self-supervised framework.
\section{Methodology}\label{sec:method}

\subsection{Overview}

We propose \textbf{SSMB}, a deblur-free, self-supervised keypoint detector specifically designed for motion-blurred images.
In contrast to BALF~\cite{zhao2024balf}, which learns to regress SIFT keypoints extracted from sharp images, SSMB is trained without any handcrafted or external pseudo-labels, eliminating the SIFT dependency that limits generalization.
Without an external signal such as SIFT to guide it, however, SSMB must learn on its own which local structures are worth attending to.
To this end, we introduce a novel \textbf{Local Discriminability Enhancement (LDE)} module that explicitly preserves and reinforces these local cues throughout the network, making self-supervised learning under blur effective in the first place.

\begin{figure*}[t]
    \centering
    \includegraphics[width=0.95\textwidth]{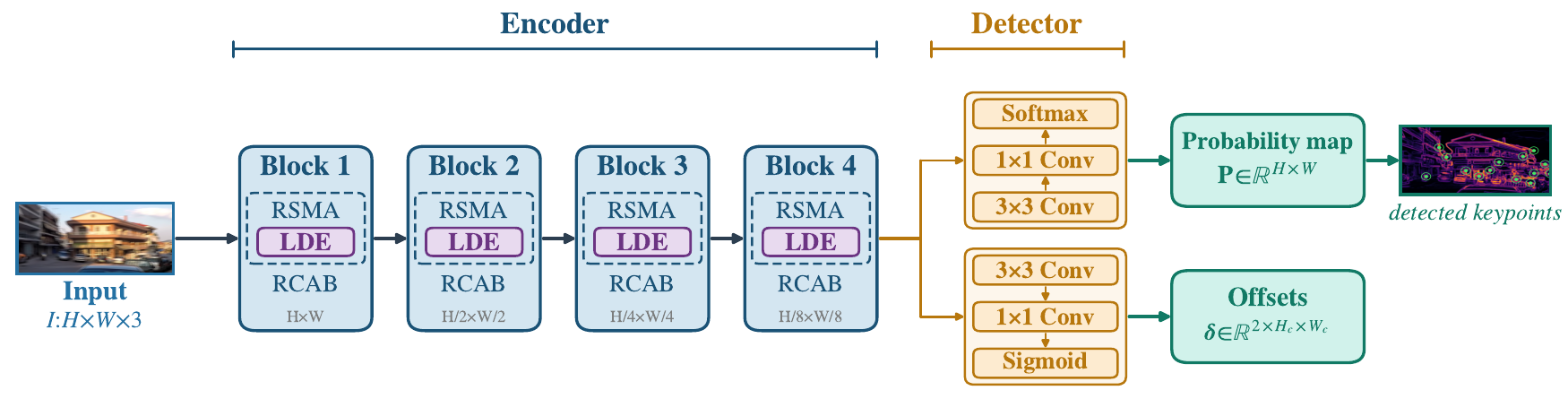}
    \caption{\textbf{Network architecture of SSMB.}
    Given an input image $I \in \mathbb{R}^{H\times W\times 3}$, the \emph{encoder} progressively downsamples the feature map through four cascaded blocks, from $H\times W$ to $\frac{H}{8}\times\frac{W}{8}$, while expanding the channel dimension.
    Each block contains an RSMA layer, within which the Local Discriminability Enhancement (LDE) module is inserted between its two internal branches, and a subsequent RCAB layer.
    The \emph{detector head} then takes the encoder output and branches into two independent towers: a probability tower (two convolutions followed by a Softmax) that produces a keypoint probability map $\mathbf{P}\in\mathbb{R}^{H\times W}$, and an offset tower (two convolutions followed by a Sigmoid) that produces sub-pixel position offsets $\boldsymbol{\delta}\in\mathbb{R}^{2\times H_c\times W_c}$ ($H_c = H/8$, $W_c = W/8$).
    Keypoints are localized at the cells where $\mathbf{P}$ peaks, refined by $\boldsymbol{\delta}$.
    }
    \label{fig:architecture}
\end{figure*}

\cref{fig:architecture} illustrates the network architecture.
SSMB consists of two main components: \textbf{\textit{(i)}} an \emph{MLP-based encoder} with an LDE module inserted in each of its four blocks, and \textbf{\textit{(ii)}} a \emph{detector head} that produces a dense keypoint probability map and sub-pixel localization offsets.

Training is performed in two self-supervised stages, illustrated in \cref{fig:training}.
In the first stage (\emph{geometric pretraining}), the model learns to detect spatially discriminative keypoints from synthetic geometric shapes, where reliable corner annotations are derived directly from the rendered geometry.
In the second stage (\emph{blur-aware training}), the model leverages real sharp-blur image pairs from the GoPro dataset~\cite{Nah2017DeepMC} and a multi-component self-supervised loss that enforces cross-domain detection consistency and spatial spreading over the whole image space.

\subsection{Network Architecture}

\PAR{MLP-based encoder.}
The encoder adopts the multi-axis gated MLP design used in BALF~\cite{zhao2024balf}, which is built upon the MAXIM architecture~\cite{Tu2022MAXIMMM}.
It consists of four cascaded blocks.
Each block contains a linear projection layer, a \emph{ResidualSplitHeadMultiAxisGmlpLayer} (RSMA), a \emph{ResidualChannelAttentionBlock} (RCAB), and a max-pooling downsampler (disabled in the last block).
The encoder progressively reduces spatial resolution from $H \times W$ to $\frac{H}{8} \times \frac{W}{8}$ while expanding channel dimensions from 3 to $\{32, 64, 128, 256\}$.
The RSMA layer splits the feature map into two branches processed in parallel: a \emph{GridGmlpLayer} for global spatial mixing via dilated grid partitioning, and a \emph{BlockGmlpLayer} for local spatial mixing via dense block partitioning~\cite{Tu2022MAXIMMM}.
This dual-path design captures both long-range structural context and local texture patterns, which is particularly important for motion-blurred inputs where pixel intensities are spread across neighborhoods.
Intuitively, grid partitioning subsamples spatially distant tokens into each group, giving each token a coarse, image-wide receptive field at low computational cost, while block partitioning groups spatially adjacent tokens, preserving fine local detail.
We refer readers to the original MAXIM paper~\cite{Tu2022MAXIMMM} for a complete architectural specification.

\begin{figure}[t]
    \centering
    \includegraphics[width=0.76\columnwidth]{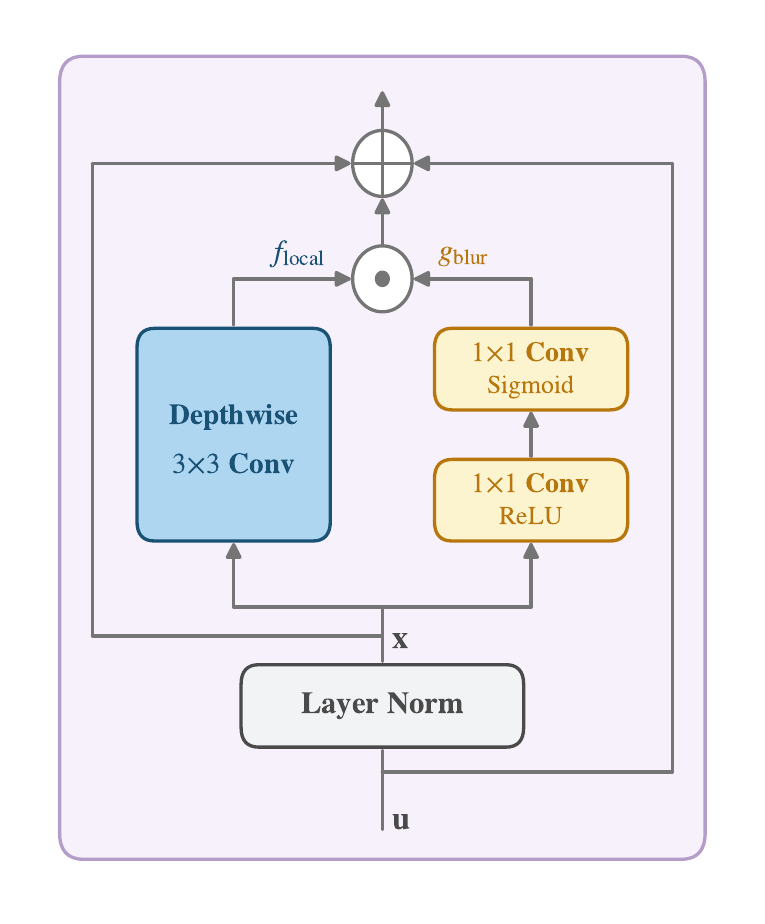}
    \caption{\textbf{Architecture of the LDE module.}
    The input feature $\mathbf{u}$ is first normalised by Layer Norm to produce $\mathbf{x} = \text{LN}(\mathbf{u})$.
    Two parallel branches then operate on $\mathbf{x}$: a depthwise $3{\times}3$ convolution $f_{\mathrm{local}}$ that captures local gradient structure, and a lightweight, feature-dependent gating branch $g_{\mathrm{blur}}$ consisting of a two-layer pointwise convolution bottleneck with a ReLU in between, followed by a Sigmoid activation.
    The branch outputs are fused by element-wise multiplication ($\odot$) and added to $\mathbf{x}$ via an inner residual, and then to the original input $\mathbf{u}$ via an outer residual, yielding
    $\text{LDE}(\mathbf{u}) = \mathbf{u} + \text{LN}(\mathbf{u}) + f_{\text{local}}(\text{LN}(\mathbf{u})) \odot g_{\text{blur}}(\text{LN}(\mathbf{u}))$.}
    \label{fig:lde}
\end{figure}

\PAR{Local Discriminability Enhancement (LDE).}
Global MLP mixing improves contextual understanding but can dilute the fine-grained local discriminability that precise keypoint localization depends on.
This risk is particularly acute for SSMB, since it is trained without any external signal to anchor the network toward geometrically meaningful locations.
To compensate for that, we introduce the LDE module, inserted within the RSMA layer immediately after the \emph{GridGmlpLayer} and before the \emph{BlockGmlpLayer}.
LDE operates on the intermediate feature tensor $\mathbf{u} \in \mathbb{R}^{B \times H \times W \times C}$ (the global-mixing branch output).
It first applies layer normalization, then extracts local gradient features via a depthwise convolution, gated by a learned, feature-dependent channel attention, as illustrated in \cref{fig:lde}:
\begin{equation}
    \text{LDE}(\mathbf{u}) = \mathbf{u} + \text{LN}(\mathbf{u}) + f_{\text{local}}(\text{LN}(\mathbf{u})) \odot g_{\text{blur}}(\text{LN}(\mathbf{u})),
    \label{eq:lde}
\end{equation}
where $\text{LN}(\cdot)$ denotes layer normalization, $f_{\text{local}}(\cdot)$ is a depthwise convolution capturing local spatial patterns, and $g_{\text{blur}}(\cdot)$ is a two-layer pointwise convolution bottleneck that adaptively weights the local enhancement based on channel statistics.
Both operations are applied in the spatial format via permutation, with a residual connection preserving the input features.
This design introduces a negligible parameter overhead ($\approx 50$K) while meaningfully improving keypoint localization under blur.
We visually confirm the necessity of this design in \cref{sec:ablation}, where removing LDE causes the predicted probability map to collapse toward near-zero almost everywhere.

\PAR{Detector head.}
Following~\cite{DeTone2018SuperPointSI,zhao2024balf}, the detector head takes the encoder output $\mathcal{F} \in \mathbb{R}^{B \times C \times H_c \times W_c}$ (where $H_c = H/8$, $W_c = W/8$) and splits into two independent branches.
The probability branch produces the keypoint probability map $\mathbf{P} \in \mathbb{R}^{B \times H \times W}$: two convolutional layers followed by a Softmax yield logits $\mathbf{L} \in \mathbb{R}^{B \times 65 \times H_c \times W_c}$ over $8 \times 8 = 64$ pixel positions per cell plus one dustbin channel, which are then converted to a full-resolution probability map via pixel-shuffle after discarding the dustbin channel.
The offset branch produces sub-pixel position offsets $\boldsymbol{\delta} \in \mathbb{R}^{B \times 2 \times H_c \times W_c}$: two convolutional layers followed by a Sigmoid predict the $(x,y)$ offset within each cell.
Keypoints are localized at the cells where $\mathbf{P}$ peaks, refined by $\boldsymbol{\delta}$.

\subsection{Self-Supervised Training}

\begin{figure*}[t]
    \centering
    \includegraphics[width=0.95\textwidth]{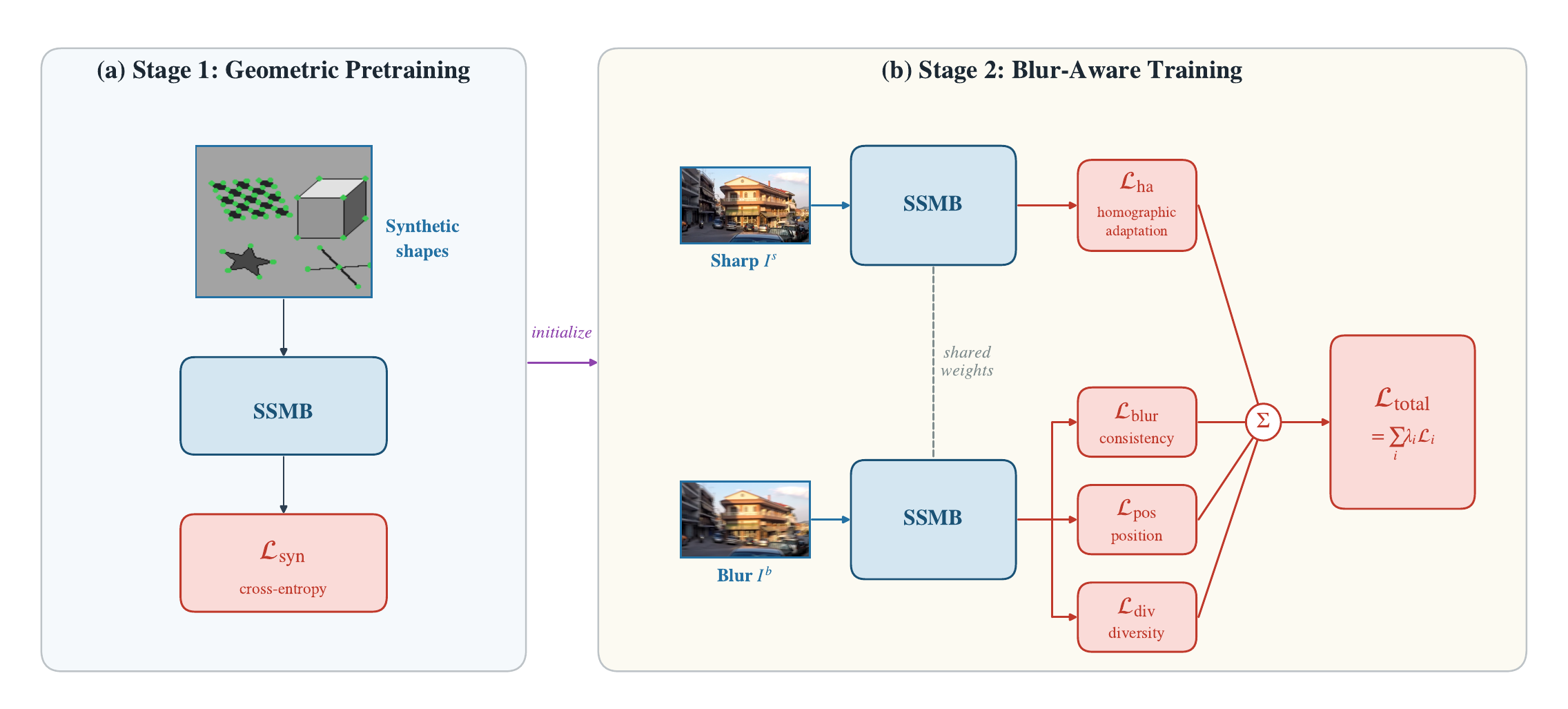}
    \caption{\textbf{Two-stage self-supervised training pipeline of SSMB.}
    \textbf{(a)}~\textbf{Stage~1 (geometric pretraining)}: SSMB is pretrained on synthetic geometric shapes with corner ground truth rendered directly from the synthetic images (no handcrafted or learned detector involved), supervised by the cross-entropy loss $\mathcal{L}_{\text{syn}}$.
    \textbf{(b)}~\textbf{Stage~2 (blur-aware training)}: a single SSMB network with shared weights, initialized from the Stage~1 checkpoint, processes sharp $I^s$ and blurred $I^b$ GoPro image pairs.
    The sharp-image branch is supervised by the homographic-adaptation loss $\mathcal{L}_{\text{ha}}$.
    The blur-image branch is supervised by the blur consistency loss $\mathcal{L}_{\text{blur}}$, the position consistency loss $\mathcal{L}_{\text{pos}}$, and the spatial diversity loss $\mathcal{L}_{\text{div}}$.
    All four terms are combined into the total training loss $\mathcal{L}_{\text{total}} = \sum_i \lambda_i \mathcal{L}_i$, $i \in \{\text{ha},\text{blur},\text{pos},\text{div}\}$.
    }
    \label{fig:training}
\end{figure*}

SSMB is trained entirely without manual keypoint annotations, in the two self-supervised stages shown in \cref{fig:training}.
We describe each stage and the loss functions used in each.

\subsubsection{Geometric Pretraining}

This first stage initializes the detector with meaningful spatial discriminability before any blur-specific training.
Without this initialization, we observe \emph{score map collapse}: all spatial locations receive nearly identical scores (Pearson correlation $\approx 1.0$ across different images), making subsequent blur-aware training ineffective.

\PAR{Synthetic pretraining.}
We generate synthetic images on-the-fly by randomly composing geometric primitives (checkerboards, line segments, polygons, ellipses, and star patterns) on grayscale backgrounds, inspired by the synthetic pre-training paradigm of SuperPoint~\cite{DeTone2018SuperPointSI}.
Since the geometry is fully controlled, reliable corner labels can be derived directly from the rendered images without requiring any external supervision or handcrafted annotations.
For each synthetic image, we produce a label map $\mathbf{Y} \in \{0, \ldots, 64\}^{H_c \times W_c}$: each $8 \times 8$ cell is assigned the index of the corner-occupied pixel, or the dustbin class (64) if no corner is present.
The network is trained with a masked cross-entropy (CE) loss:
\begin{equation}
    \mathcal{L}_{\text{syn}} = \frac{1}{|\mathcal{V}|} \sum_{(i,j) \in \mathcal{V}}
    \text{CE}\bigl(\mathbf{L}_{ij},\, \mathbf{Y}_{ij}\bigr),
    \label{eq:syn_loss}
\end{equation}
where $\mathcal{V}$ denotes the set of valid (non-border) cells and $\mathbf{L}_{ij}$ are the model logits for cell $(i,j)$.

\subsubsection{Blur-Aware Training}

In this second stage, we train the model on real sharp-blur image pairs from the GoPro dataset~\cite{Nah2017DeepMC}.
A single SSMB network performs two forward passes per training iteration, one on the sharp image $I^s$ and one on its paired blurred image $I^b$, sharing all weights.
The sharp forward pass serves a dual role. It generates the homographic adaptation pseudo-label target and provides the consistency supervision signal for the blur forward pass.
The total training loss combines four terms:
\begin{equation}
    \mathcal{L} = \lambda_{\text{ha}} \mathcal{L}_{\text{ha}}
                + \lambda_{\text{blur}} \mathcal{L}_{\text{blur}}
                + \lambda_{\text{pos}} \mathcal{L}_{\text{pos}}
                + \lambda_{\text{div}} \mathcal{L}_{\text{div}}.
    \label{eq:total_loss}
\end{equation}

\PAR{Homographic adaptation loss $\mathcal{L}_{\text{ha}}$.}
For each sharp image in the batch, we perform online Homographic Adaptation (HA)~\cite{DeTone2018SuperPointSI}.
Specifically, we generate $N_H$ random homographies $\{\mathbf{H}_k\}_{k=1}^{N_H}$, warp the sharp image, and run the current detector on each warped image to obtain a probability map $P_k$.
We then aggregate these warped probability maps back to the original image coordinates via:
\begin{equation}
    \hat{P}(i,j) = \frac{\sum_{k=1}^{N_H} \mathbf{H}_k^{-1} \bigl[ P_k(i,j) \bigr] \cdot \mathcal{M}_k(i,j)}
                        {\sum_{k=1}^{N_H} \mathcal{M}_k(i,j) + \epsilon},
    \label{eq:ha_agg}
\end{equation}
where $\mathcal{M}_k$ is the validity mask for homography $k$, and homographies covering less than 50\% of the image are discarded.
The aggregated map is discretized into per-cell pseudo-labels $\hat{\mathbf{Y}}$, and the HA loss supervises the sharp-image forward pass:
\begin{equation}
    \mathcal{L}_{\text{ha}} = \frac{1}{|\mathcal{V}|}
    \sum_{(i,j) \in \mathcal{V}} \text{CE}\bigl(\mathbf{L}^s_{ij},\, \hat{\mathbf{Y}}_{ij}\bigr),
    \label{eq:ha_loss}
\end{equation}
where $\mathbf{L}^s_{ij}$ are the cell $(i,j)$'s logits from the sharp-image forward pass and $\hat{\mathbf{Y}}_{ij}$ are their respective HA pseudo-labels.
This loss anchors the network's detection response to stable, geometry-consistent locations.

\PAR{Blur consistency loss $\mathcal{L}_{\text{blur}}$.}
The core self-supervised signal for blur robustness enforces that the blurred image produces keypoints at the same locations as its paired sharp image.
We treat the sharp image's per-cell $\arg\max$ prediction as a hard target for the blurred image:
\begin{equation}
    \mathcal{L}_{\text{blur}} = \frac{1}{|\mathcal{V}|} \sum_{(i,j) \in \mathcal{V}} \text{CE}\bigl(\mathbf{L}^b_{ij},\,
    \underbrace{\arg\max_{c}\, \mathbf{L}^s_{ij}}_{\text{detached}}\bigr),
    \label{eq:blur_loss}
\end{equation}
where $\mathbf{L}^b_{ij}$ and $\mathbf{L}^s_{ij}$ are the cell $(i,j)$'s logits from the blurred- and sharp-image forward passes, respectively.
``detached'' indicates that gradients are not backpropagated through the sharp $\arg\max$ target, so it is treated as a fixed pseudo-label rather than being optimized jointly with the blurred-image prediction.
This hard consistency formulation differs from soft Kullback–Leibler (KL) divergence distillation, as it is computationally efficient and directly aligns the predicted keypoint cells rather than the full distribution.

\PAR{Position consistency loss $\mathcal{L}_{\text{pos}}$.}
The detector head also predicts sub-pixel offsets $\boldsymbol{\delta} \in [0,1]^{2 \times H_c \times W_c}$ within each cell.
To ensure that the blurred image predicts the same sub-pixel positions as the sharp image, we apply an $\ell_2$ regression loss:
\begin{equation}
    \mathcal{L}_{\text{pos}} = \frac{1}{H_c W_c}
    \bigl\| \boldsymbol{\delta}^s - \boldsymbol{\delta}^b \bigr\|_F^2,
    \label{eq:pos_loss}
\end{equation}
where $\boldsymbol{\delta}^s$ and $\boldsymbol{\delta}^b$ are the position offsets from the sharp and blur forward passes, respectively.

\PAR{Spatial diversity loss $\mathcal{L}_{\text{div}}$.}
Without explicit coverage enforcement, the network tends to collapse to detecting only the most salient locations (typically image centers or high-contrast edges), leaving most of the image with near-zero response.
We prevent this by partitioning the feature map into a $G_h \times G_w$ grid of spatial cells and penalizing low maximum response in any cell:
\begin{equation}
    \mathcal{L}_{\text{div}} = -\frac{1}{G_h G_w}
    \sum_{m=1}^{G_h G_w} \log\bigl(\max_{p \in \mathcal{C}_m} \hat{P}^b_p + \epsilon\bigr),
    \label{eq:div_loss}
\end{equation}
where $\hat{P}^b$ is the probability map (dustbin excluded) from the blurred forward pass, $\mathcal{C}_m$ indexes the pixels within the $m$-th grid cell, and $\epsilon = 10^{-6}$ ensures numerical stability.
This loss encourages the detector to maintain at least one confident keypoint in every spatial region, promoting well-distributed detections across the image.

The loss weights are set to $\lambda_{\text{ha}} = 0.5$, $\lambda_{\text{blur}} = 0.5$, $\lambda_{\text{pos}} = 0.1$, and $\lambda_{\text{div}} = 0.005$. Despite the small weight, $\mathcal{L}_{\text{div}}$ proves to be the most critical component (see \cref{sec:ablation}).
Without it, detection again collapses onto a handful of salient locations even after the geometric pretraining stage.

\subsection{Implementation Details}

\PAR{Stage 1 training.}
We pretrain on $25,000$ synthetic geometric images for $10$ epochs (batch size $8$, $320 \times 320$ crops, Adam optimizer with initial lr $= 10^{-5}$, linear warmup over $4$ epochs).

\PAR{Stage 2 training.}
We train on the GoPro dataset~\cite{Nah2017DeepMC} ($2,912$ sharp-blur pairs from $30$ sequences) for $36$ epochs starting from the Stage~1 checkpoint, with batch size $8$, $320 \times 320$ random crops, $N_H = 100$ homographies per image for online HA, spatial diversity grid $G_h = G_w = 8$, and rotation augmentation up to $\pm 90^\circ$.
The Adam optimizer uses initial lr $= 10^{-5}$, decaying at $60\%$ and $80\%$ of total epochs by a factor of $0.1$.

\PAR{Data augmentation.}
During Stage 2, each sharp-blur pair undergoes synchronized geometric augmentation (random homography, horizontal flip) followed by independent color jitter (brightness $\pm 0.4$, contrast $\pm 0.3$, saturation $\pm 0.3$, hue $\pm 0.1$) and random grayscale conversion ($p=0.1$).
All training is conducted on a single NVIDIA GeForce RTX 3090.
\section{Experiments}\label{sec:experiments}

\subsection{Datasets}

\PAR{Blur-HPatches.}
As no specific benchmark exists for evaluating keypoint detectors directly on motion-blurred images, we use the Blur-HPatches dataset introduced in~\cite{zhao2024balf}.
It is derived from the original HPatches dataset~\cite{Balntas2017HPatchesAB}, which provides $116$ sequences ($59$ viewpoint changes and $57$ illumination changes), each with one reference image and five target images accompanied by ground-truth homographies.
For each image, motion blur is synthesized by convolving with a randomly generated point spread function.
Three difficulty levels are defined by increasing blur kernel size and motion irregularity: \textsc{Easy}, \textsc{Hard}, and \textsc{Tough}.
We evaluate SSMB under two settings: \emph{blur-to-sharp} (reference sharp, target blurred) and \emph{blur-to-blur} (both images blurred).
The original HPatches dataset (all images sharp) is additionally used to assess performance on clean images.

\PAR{Deblur-HPatches.}
Also introduced in~\cite{zhao2024balf}, this dataset applies two state-of-the-art deblurring networks, SRN-DeblurNet~\cite{Tao2018ScaleRecurrentNF} and DeblurGAN-v2~\cite{Kupyn2019DeblurGANv2D}, to the Blur-HPatches images without any fine-tuning.
It is used to evaluate whether a deblur-then-detect pipeline can match the performance of a deblur-free approach.

\PAR{ArchViz.}
For relative pose estimation, we use the ArchViz dataset~\cite{liu2021mba}, generated using the Unreal Engine to simulate rapid back-and-forth camera movements.
It provides ground-truth camera poses and paired sharp and blurred images, along with texture-less regions and repetitive patterns that make pose estimation particularly challenging.
The evaluation set consists of $3,321$ image pairs at $768 \times 480$ resolution, with up to $2,048$ keypoints per image.

\PAR{Aachen Day-Night.}
For visual localization, we use the Aachen Day-Night benchmark~\cite{Sattler2018CVPR}, which covers $824$ daytime and $98$ nighttime query images against a 3D map built from $4,328$ daytime reference images.
Following the established protocol for evaluating motion-blur robustness on real photographic benchmarks~\cite{liu2021mba,zhao2024balf,wang2025mba}, we synthesize motion blur on both query and database images using the same procedure as Blur-HPatches.
We use the original Aachen v1.0 benchmark to maintain consistency with this protocol.

\subsection{Evaluation Metrics}

\PAR{Repeatability.}
We use the repeatability metric~\cite{Mikolajczyk2003APE} to evaluate keypoint detection quality.
For an image pair, repeatability is the ratio of mutually corresponding keypoints to the smaller of the two detection counts.
Correspondences are identified via overlap error $\epsilon_{\text{IoU}}$ between keypoint regions~\cite{zhang2017learning}, with a match accepted if $\epsilon_{\text{IoU}} < 0.4$ (region overlap greater than $60\%$).
We use $1,000$ keypoints per image.

\PAR{MMA.}
For image matching, we report the Mean Matching Accuracy (MMA)~\cite{Dusmanu2019D2NetAT}, defined as the ratio of correctly matched keypoints at pixel thresholds of $3$, $5$, and $10$ pixels, using up to $2,048$ keypoints per image.

\PAR{Pose AUC.}
For relative pose estimation, we report the Area Under the Cumulative Curve (AUC) of the pose error at thresholds $5^\circ$, $10^\circ$, $20^\circ$, and $30^\circ$, following~\cite{sarlin20superglue,sun2021loftr}.
The pose error is the maximum angular error in rotation and translation.
We solve the essential matrix using the five-point algorithm~\cite{nister2004efficient} with RANSAC~\cite{Fischler1981RandomSC}, with the inlier threshold tuned per method on the test data, following~\cite{lindenberger2023lightglue}, since evaluating RANSAC itself is not the focus of this comparison.

\PAR{Localization accuracy.}
For visual localization, we report the percentage of queries correctly localized within error thresholds $(0.25\text{m}, 2^\circ)$, $(0.5\text{m}, 5^\circ)$, and $(1.0\text{m}, 10^\circ)$.

\subsection{Baseline Methods}

We compare SSMB against a comprehensive set of methods spanning three categories.

\PAR{Sparse methods.}
We compare against classical SIFT~\cite{LoweDavid2004DistinctiveIF}, DoG~\cite{LoweDavid2004DistinctiveIF}, and a comprehensive set of learning-based methods: Key.Net~\cite{Laguna2019KeyNetKD}, SuperPoint~\cite{DeTone2018SuperPointSI}, LF-Net~\cite{Ono2018LFNetLL}, D2-Net~\cite{Dusmanu2019D2NetAT}, R2D2~\cite{Revaud2019R2D2RA}, DISK~\cite{tyszkiewicz2020disk}, SiLK~\cite{gleize2023silk}, REKD~\cite{lee2022self}, NeSS-ST~\cite{pakulev2023ness}, ALIKED~\cite{Zhao2023ALIKED}, DeDoDe v2~\cite{edstedt2024dedodev2}, XFeat~\cite{potje2024cvpr}, and BALF~\cite{zhao2024balf}.
For downstream tasks  (image matching, pose estimation, and visual localization), each method is paired with its native descriptor (or HardNet~\cite{Mishchuk2017WorkingHT}/HyNet~\cite{hynet2020} when a native descriptor is unavailable) and matched via non-learned matching (mutual nearest neighbor, MNN, or Dual-Softmax), except where noted otherwise.
SuperPoint, DISK, and ALIKED are additionally evaluated with LightGlue~\cite{lindenberger2023lightglue} as a learned matcher.

\PAR{Detector-free methods.}
For image matching, pose estimation, and visual localization, we include LoFTR~\cite{sun2021loftr}, MatchFormer~\cite{wang2022matchformer}, ASpanFormer~\cite{chen2022aspanformer}, and RoMa v2~\cite{edstedt2025roma}.
These methods operate on full image pairs and represent a complementary paradigm to sparse detectors.
They are reported separately and highlights are applied within sparse methods only, as the two paradigms are not directly comparable in terms of computational cost and target downstream tasks.

For evaluation in downstream tasks, SSMB keypoints are paired with HardNet~\cite{Mishchuk2017WorkingHT} or HyNet~\cite{hynet2020} descriptors and MNN matching.

Note that the red, orange, and yellow highlights in all tables indicate the \colorbox{tabfirst}{1st}, \colorbox{tabsecond}{2nd}, and \colorbox{tabthird}{3rd} best performing sparse method for each metric, respectively.

\subsection{Keypoint Detection}\label{ssec:keypointdetection}

\begin{table}[t]
    \scriptsize
    \centering
    \renewcommand\arraystretch{1.1}
    \setlength{\tabcolsep}{4pt}
    \caption{\textbf{Keypoint detection repeatability on the Blur-HPatches dataset~\cite{zhao2024balf}.}
    For compactness, we only report the overall repeatability.
    The disaggregated results for viewpoint and illumination changes can be found in \cref{subsec:keypoint_detection_appendix} of the Appendix.}
    \begin{tabular}{lcccccc}
        \toprule
        & \multicolumn{3}{c}{\begin{tabular}[c]{@{}c@{}}Reference: Sharp\\Target: Blur\end{tabular}}
        & \multicolumn{3}{c}{\begin{tabular}[c]{@{}c@{}}Reference: Blur\\Target: Blur\end{tabular}}\\
        \cmidrule(l){2-4} \cmidrule(l){5-7}
        Method & E\begin{tiny}ASY\end{tiny} $\uparrow$ & H\begin{tiny}ARD\end{tiny} $\uparrow$ & T\begin{tiny}OUGH\end{tiny} $\uparrow$ & E\begin{tiny}ASY\end{tiny} $\uparrow$ & H\begin{tiny}ARD\end{tiny} $\uparrow$ & T\begin{tiny}OUGH\end{tiny} $\uparrow$ \\
        \midrule
        SIFT\cite{LoweDavid2004DistinctiveIF}           & 55.92 & 56.80 & 53.49 & 56.99 & 53.49 & 45.94
        \\
        Key.Net~\cite{Laguna2019KeyNetKD}               & 60.34 & 54.71 & 44.69 & \cellcolor{tabthird}62.77 & 58.17 & 49.25
        \\
        SuperPoint~\cite{DeTone2018SuperPointSI}        & \cellcolor{tabthird}65.64 & \cellcolor{tabthird}62.22 & 52.84 & 58.60 & 50.03 & 43.28
        \\
        LF-Net~\cite{Ono2018LFNetLL}                    & 63.54 & 61.19 & \cellcolor{tabthird}56.78 & 60.45 & 59.07 & 57.71
        \\
        D2-Net~\cite{Dusmanu2019D2NetAT}                & 49.71 & 47.30 & 44.32 & 51.80 & 51.05 & 50.53
        \\
        R2D2~\cite{Revaud2019R2D2RA}                    & 57.99 & 51.73 & 40.57 & 57.49 & 55.31 & 46.86
        \\
        DISK~\cite{tyszkiewicz2020disk}                 & 60.24 & 58.18 & 55.97 & 61.89 & \cellcolor{tabthird}60.87 & \cellcolor{tabthird}60.65
        \\
        REKD~\cite{lee2022self}                         & 55.15 & 53.15 & 48.52 & 52.57 & 48.54 & 44.02
        \\
        NeSS-ST~\cite{pakulev2023ness}                & 57.93 & 56.12 & 53.83 & 60.97 & 59.80 & 59.21
        \\
        ALIKED~\cite{Zhao2023ALIKED}                    & 58.88 & 54.12 & 50.36 & 62.19 & 59.49 & 57.57
        \\
        DeDoDe v2~\cite{edstedt2024dedodev2}            & 52.78 & 49.41 & 46.78 & 60.17 & 57.39 & 54.99
        \\
        XFeat~\cite{potje2024cvpr}                      & 54.82 & 51.59 & 49.56 & 59.70 & 58.29 & 57.74
        \\
        BALF~\cite{zhao2024balf}                        & \cellcolor{tabsecond}74.12 & \cellcolor{tabsecond}74.45 & \cellcolor{tabsecond}71.84 & \cellcolor{tabsecond}70.48 & \cellcolor{tabsecond}68.43 & \cellcolor{tabsecond}67.71 \\
        \midrule
        SSMB (Ours)                                & \cellcolor{tabfirst}77.24 & \cellcolor{tabfirst}77.18 & \cellcolor{tabfirst}77.16 & \cellcolor{tabfirst}74.20  & \cellcolor{tabfirst}74.33 & \cellcolor{tabfirst}74.48 \\
        \bottomrule
    \end{tabular}
    \label{tab:blur_HPatches}
\end{table}

Although SSMB is designed for motion-blurred images, we additionally verify its performance on the original, all-sharp HPatches dataset, where it remains competitive with methods explicitly optimized for sharp images.
Full results are provided in \cref{subsec:keypoint_detection_appendix} of the Appendix.

\PAR{Results on Blur-HPatches.}
\cref{tab:blur_HPatches} presents the main evaluation on motion-blurred images.
In the \emph{blur-to-sharp} setting, SSMB achieves $77.24\%$, $77.18\%$, and $77.16\%$ on \textsc{Easy}, \textsc{Hard}, and \textsc{Tough} respectively, outperforming BALF at every difficulty level.
Importantly, SSMB's performance remains remarkably stable across difficulty levels, whereas BALF drops from $74.12\%$ to $71.84\%$ and most other methods degrade substantially as blur severity increases.
This stability is a direct consequence of the blur consistency training objective, which explicitly aligns detection responses across different levels of blur degradation.
In the \emph{blur-to-blur} setting, SSMB outperforms all methods across all difficulty levels, achieving $74.20\%$, $74.33\%$, and $74.48\%$ on \textsc{Easy}, \textsc{Hard}, and \textsc{Tough} respectively.
Unlike most detectors, which degrade as blur severity increases, SSMB exhibits a slight improvement, a further consequence of the blur consistency loss.

\begin{table*}[ht]
    \scriptsize
    \centering
    \renewcommand\arraystretch{1.1}
    \setlength{\tabcolsep}{7pt}
    \caption{\textbf{Keypoint detection repeatability on the Deblur-HPatches dataset~\cite{zhao2024balf},} where blurred images are first restored using SRN-DeblurNet or DeblurGAN-v2 before detection.}
    \begin{tabular}{lcccccccccccc}
        \toprule
        & \multicolumn{6}{c}{\begin{tabular}[c]{@{}c@{}}Reference: Sharp\\Target: Deblur\end{tabular}} 
        & \multicolumn{6}{c}{\begin{tabular}[c]{@{}c@{}}Reference: Deblur\\Target: Deblur\end{tabular}} 
        \\
        \cmidrule(l){2-7} \cmidrule(l){8-13}
        & \multicolumn{3}{c}{SRN-DeblurNet~\cite{Tao2018ScaleRecurrentNF}} 
        & \multicolumn{3}{c}{DeblurGAN-v2~\cite{Kupyn2019DeblurGANv2D}} 
        & \multicolumn{3}{c}{SRN-DeblurNet~\cite{Tao2018ScaleRecurrentNF}}
        & \multicolumn{3}{c}{DeblurGAN-v2~\cite{Kupyn2019DeblurGANv2D}}
        \\ \cmidrule(l){2-4} \cmidrule(l){5-7} \cmidrule(l){8-10} \cmidrule(l){11-13}
        Method & 
        E\begin{tiny}ASY\end{tiny} $\uparrow$ & H\begin{tiny}ARD\end{tiny}  $\uparrow$ & T\begin{tiny}OUGH\end{tiny} $\uparrow$ &
        E\begin{tiny}ASY\end{tiny}  $\uparrow$& H\begin{tiny}ARD\end{tiny} $\uparrow$ & T\begin{tiny}OUGH\end{tiny} $\uparrow$ &
        E\begin{tiny}ASY\end{tiny} $\uparrow$ & H\begin{tiny}ARD\end{tiny} $\uparrow$ & T\begin{tiny}OUGH\end{tiny} $\uparrow$ &
        E\begin{tiny}ASY\end{tiny}  $\uparrow$ & H\begin{tiny}ARD\end{tiny} $\uparrow$ & T\begin{tiny}OUGH\end{tiny} $\uparrow$
        \\
        \midrule
        SIFT~\cite{LoweDavid2004DistinctiveIF} &
        56.62 & 55.36 & 53.83 &
        57.63 & 56.52 & 56.50 &
        59.75 & 58.13 & 50.63 &
        59.44 & 57.98 & 51.21
        \\
        Key.Net~\cite{Laguna2019KeyNetKD} &
        63.28 & 58.01 & 47.10 &
        63.99 & 59.16 & 49.35 &
        62.86 & 60.44 & 50.74 &
        62.73 & 60.58 & 52.96
        \\
        SuperPoint~\cite{DeTone2018SuperPointSI} &
        \cellcolor{tabthird}67.72 & \cellcolor{tabthird}64.05 & 55.26 &
        \cellcolor{tabthird}67.95 & \cellcolor{tabthird}65.86 & \cellcolor{tabthird}58.22 &
        66.38 & 63.16 & 49.52 &
        66.50 & 63.71 & 52.09
        \\
        LF-Net~\cite{Ono2018LFNetLL} & 
        62.22 & 59.90 & 54.73 &
        62.59 & 60.24 & 54.81 &
        63.06 & 62.03 & 57.28 &
        63.00 & 61.79 & 57.85
        \\
        D2-Net~\cite{Dusmanu2019D2NetAT} &
        51.81 & 49.49 & 45.94 &
        52.64 & 50.21 & 45.88 &
        53.60 & 53.00 & 50.93 &
        53.93 & 53.29 & 50.74
        \\
        R2D2~\cite{Revaud2019R2D2RA} &
        60.31 & 55.43 & 43.26 &
        60.46 & 55.68 & 45.38 &
        58.11 & 54.80 & 45.77 &
        57.95 & 55.03 & 47.86
        \\
        DISK~\cite{tyszkiewicz2020disk}  &
        63.44 & 61.74 & \cellcolor{tabthird}57.18 &
        63.25 & 61.76 & 58.12 &
        64.10 & 63.46 & \cellcolor{tabthird}59.46 &
        63.86 & 62.86 & 59.86
        \\
        REKD~\cite{lee2022self} &
        55.18 & 53.87 & 49.50 &
        55.27 & 53.82 & 50.65 &
        54.22 & 51.99 & 45.35 &
        53.86 & 51.61 & 45.59 
        \\
        NeSS-ST~\cite{pakulev2023ness} &
        61.26 & 59.61 & 55.37 &
        61.40 & 59.88 & 56.27 &
        63.11 & 62.18 & 59.33 &
        62.69 & 62.01 & 59.39 
        \\
        ALIKED~\cite{Zhao2023ALIKED} &
        66.24 & 63.51 & 54.96 &
          66.81 & 64.16 & 57.28 &
        \cellcolor{tabthird}68.45 & \cellcolor{tabthird}66.26 & 58.19 &
        \cellcolor{tabthird}68.03 & \cellcolor{tabthird}66.29 & \cellcolor{tabthird}60.78 
        \\
        DeDoDe v2~\cite{edstedt2024dedodev2} &
        56.80 & 55.15 & 47.88 &
        57.75 & 56.18 & 51.12 &
        61.67 & 60.89 & 52.91 &
        61.34 & 60.75 & 56.97 
        \\
        XFeat~\cite{potje2024cvpr} &
        58.59 & 57.50 & 52.47 &
        59.13 & 57.77 & 53.34 &
        61.68 & 61.14 & 58.01 &
        61.43 & 60.87 & 58.59 \\
        \midrule
        BALF~\cite{zhao2024balf} &
        \cellcolor{tabsecond}74.12 & \cellcolor{tabsecond}74.45 & \cellcolor{tabsecond}71.84 &
        \cellcolor{tabsecond}74.12 & \cellcolor{tabsecond}74.45 & \cellcolor{tabsecond}71.84 &
        \cellcolor{tabsecond}70.48 & \cellcolor{tabsecond}68.43 & \cellcolor{tabsecond}67.71 &
        \cellcolor{tabsecond}70.48 & \cellcolor{tabsecond}68.43 & \cellcolor{tabsecond}67.71 \\
        SSMB (Ours) &
        \cellcolor{tabfirst}77.24 & \cellcolor{tabfirst}77.18 & \cellcolor{tabfirst}77.16 &
        \cellcolor{tabfirst}77.24 & \cellcolor{tabfirst}77.18 & \cellcolor{tabfirst}77.16 &
        \cellcolor{tabfirst}74.20 & \cellcolor{tabfirst}74.33 & \cellcolor{tabfirst}74.48 &
        \cellcolor{tabfirst}74.20 & \cellcolor{tabfirst}74.33 & \cellcolor{tabfirst}74.48 \\
        \bottomrule
    \end{tabular}
	\label{tab:deblur_blur_hpatches}
\end{table*}

\PAR{Results on Deblur-HPatches.}
\cref{tab:deblur_blur_hpatches} evaluates the deblur-then-detect pipeline, where blurred images are first restored by SRN-DeblurNet or DeblurGAN-v2 and then fed to each keypoint detector.
While deblurring networks improve the performance of other detectors, neither deblurring-detector pair is able close the performance gap to SSMB.
For example, under the \textsc{Tough} \emph{deblur-to-sharp} setting, the best competing result (DeblurGAN-v2 with SuperPoint) reaches $58.22\%$, while SSMB directly on the blurred image achieves $77.16\%$.
This confirms that deblurring introduces artifacts that limit downstream keypoint detection, and that a deblur-free approach is more effective.

\subsection{Image Matching}

\begin{table*}
    \scriptsize
    \centering
    \renewcommand\arraystretch{1.1}
    \setlength{\tabcolsep}{9pt}
    \caption{\textbf{Image matching on the Blur-HPatches dataset~\cite{zhao2024balf}.}
    Mean matching accuracy (MMA) is reported at multiple pixel thresholds.
    \colorbox{tabfirst}{first}, \colorbox{tabsecond}{second}, and \colorbox{tabthird}{third} best results are highlighted within sparse methods only, as detector-free methods operate on full images and are not directly comparable to sparse detectors.
    }
    \begin{tabular}{lccccccccc}
        \toprule
        & \multicolumn{3}{c}{Overall MMA $\uparrow$} & \multicolumn{3}{c}{Illumination MMA $\uparrow$} & \multicolumn{3}{c}{Viewpoint MMA $\uparrow$} \\
        \cmidrule(l){2-4} \cmidrule(l){5-7} \cmidrule(l){8-10}
        Method & $@3$px & $@5$px & $@10$px & $@3$px & $@5$px & $@10$px & $@3$px & $@5$px & $@10$px \\
        \midrule
        \multicolumn{10}{l}{\textit{Detector-free (Dense/Semi-dense)}} \\
        LoFTR~\cite{sun2021loftr}
            & 26.40 & 35.28 & 64.96
            & 39.97 & 42.32 & 73.23
            & 13.29 & 28.48 & 56.96 \\
        MatchFormer~\cite{wang2022matchformer}
            & 38.08 & 48.86 & 73.67
            & 57.69 & 60.57 & 86.33
            & 19.14 & 37.54 & 61.45 \\
        ASpanFormer~\cite{chen2022aspanformer}
            & 22.91 & 33.20 & 65.51
            & 29.35 & 30.87 & 63.24
            & 16.69 & 35.46 & 67.70 \\
        RoMa v2~\cite{edstedt2025roma}
            & 28.51 & 54.58 & 88.54
            & 36.24 & 67.24 & 96.47
            & 21.04 & 42.34 & 80.89 \\
        \midrule
        \multicolumn{10}{l}{\textit{Sparse (detector + learned matcher)}} \\
        SuperPoint~\cite{DeTone2018SuperPointSI} + LightGlue~\cite{lindenberger2023lightglue}
            & 9.93  & 24.58 & 61.17
            & 10.88 & 26.72 & 65.55
            & 9.02  & 22.51 & 56.93 \\
        DISK~\cite{tyszkiewicz2020disk} + LightGlue~\cite{lindenberger2023lightglue}
            & 12.58 & 29.52 & 63.69
            & 15.24 & 35.01 & 71.83
            & 10.00 & 24.21 & 55.84 \\
        ALIKED~\cite{Zhao2023ALIKED} + LightGlue~\cite{lindenberger2023lightglue}
            & 11.51 & 27.36 & 62.91
            & 11.77 & 27.83 & 62.74
            & 11.27 & 26.91 & \cellcolor{tabthird}63.07 \\
        \midrule
        \multicolumn{10}{l}{\textit{Sparse (detector + non-learned matching)}} \\
        SIFT~\cite{LoweDavid2004DistinctiveIF} + MNN
            & 13.08 & 21.79 & 31.74
            & 12.85 & 21.79 & 32.42
            & 13.31 & 21.79 & 31.08 \\
        DoG~\cite{LoweDavid2004DistinctiveIF} + AffNet~\cite{AffNet2017} + HardNet~\cite{Mishchuk2017WorkingHT} + MNN
            & 7.01  & 13.50 & 24.72
            & 6.26  & 12.25 & 22.80
            & 7.73  & 14.71 & 26.58 \\
        Key.Net~\cite{Laguna2019KeyNetKD} + AffNet~\cite{AffNet2017} + HardNet~\cite{Mishchuk2017WorkingHT} + MNN
            & 10.12 & 20.68 & 41.65
            & 11.62 & 22.37 & 43.76
            & 8.68  & 19.05 & 39.61 \\
        SuperPoint~\cite{DeTone2018SuperPointSI} + MNN
            & 5.63  & 13.34 & 30.76
            & 7.21  & 16.91 & 38.03
            & 4.09  & 9.90  & 23.72 \\
        D2-Net~\cite{Dusmanu2019D2NetAT} + MNN
            & 12.79 & 27.43 & 54.80
            & 16.92 & 34.76 & 66.48
            & 8.81  & 20.35 & 43.52 \\
        R2D2~\cite{Revaud2019R2D2RA} + MNN
            & 15.39 & 28.83 & 52.28
            & 19.05 & 33.83 & 60.31
            & 11.84 & 23.99 & 44.52 \\
        DISK~\cite{tyszkiewicz2020disk} + MNN
            & 9.18  & 21.61 & 47.87
            & 11.47 & 26.56 & 56.68
            & 6.97  & 16.82 & 39.35 \\
        SiLK~\cite{gleize2023silk} + MNN
            & 2.06  & 4.13  & 7.77
            & 4.17  & 8.39  & 15.67
            & 0.02  & 0.02  & 0.15  \\
        DeDoDe v2~\cite{edstedt2024dedodev2} + Dual-Softmax
            & 10.76 & 21.71 & 46.29
            & 15.90 & 29.22 & 57.83
            & 5.80  & 14.45 & 35.13 \\
        BALF~\cite{zhao2024balf} + HardNet~\cite{Mishchuk2017WorkingHT} + MNN
            & \cellcolor{tabthird}25.03 & \cellcolor{tabthird}44.71 & \cellcolor{tabthird}72.64
            & \cellcolor{tabthird}34.58 & \cellcolor{tabthird}56.06 & \cellcolor{tabthird}84.90
            & \cellcolor{tabthird}15.80 & \cellcolor{tabthird}33.74 & 60.81 \\
        BALF~\cite{zhao2024balf} + HyNet~\cite{hynet2020} + MNN
            & 20.55 & 38.12 & 67.84
            & 27.51 & 45.90 & 78.03
            & 13.82 & 30.62 & 58.00 \\
        \cmidrule{2-10}
        SSMB (Ours) + HardNet~\cite{Mishchuk2017WorkingHT} + MNN
            & \cellcolor{tabfirst}40.10 & \cellcolor{tabfirst}51.43 & \cellcolor{tabfirst}77.22
            & \cellcolor{tabfirst}62.02 & \cellcolor{tabfirst}62.78 & \cellcolor{tabfirst}85.60
            & \cellcolor{tabfirst}18.91 & \cellcolor{tabfirst}40.46 & \cellcolor{tabfirst}69.12 \\
        SSMB (Ours) + HyNet~\cite{hynet2020} + MNN
            & \cellcolor{tabsecond}38.45 & \cellcolor{tabsecond}49.52 & \cellcolor{tabsecond}76.67
            & \cellcolor{tabsecond}59.93 & \cellcolor{tabsecond}60.23 & \cellcolor{tabsecond}84.91
            & \cellcolor{tabsecond}18.21 & \cellcolor{tabsecond}39.18 & \cellcolor{tabsecond}68.72 \\
        \bottomrule
    \end{tabular}
    \label{tab:image_matching}
\end{table*}

\PAR{Dataset and protocol.}
We evaluate image matching on the Blur-HPatches \textsc{Tough} split under the \emph{blur-to-blur} setting, using MMA at pixel thresholds of $3$, $5$, and $10$.

\PAR{Results.}
As shown in \cref{tab:image_matching}, SSMB with HardNet achieves $40.10\%$, $51.43\%$, and $77.22\%$ MMA at $3$, $5$, and $10$ pixels overall, ranking first among all sparse methods by a substantial margin.
The next best sparse method is BALF with HardNet, which reaches $72.64\%$ at $10$ pixels overall, benefiting from the same deblur-free detection paradigm but without the self-supervised training that removes SIFT dependency.
The improvement is most pronounced under illumination changes, where SSMB+HardNet achieves $62.02\%$ at $3$ pixels compared to $34.58\%$ for the next best sparse method (BALF+HardNet).
Under viewpoint changes, SSMB+HardNet achieves $69.12\%$ at $10$ pixels, also ranking first among sparse methods, ahead of BALF+HardNet at $60.81\%$.
SSMB+HyNet ranks second across all metrics, confirming that the gains stem from the detector rather than the descriptor choice.
Among detector-free methods, RoMa v2 achieves $88.54\%$ at $10$ pixels. However, this performance comes at the cost of operating on complete image pairs during inference, resulting in substantially higher computation than that of our sparse detector (see \cref{tab:efficiency}).
Note also that SSMB outperforms all detector-free methods in overall MMA at the strict threshold of $3$ pixels.

\subsection{Relative Pose Estimation}

\begin{table*}
  \scriptsize
  \centering
  \renewcommand\arraystretch{1.1}
  \setlength{\tabcolsep}{12pt}
  \caption{\textbf{Relative pose estimation on the ArchViz dataset~\cite{liu2021mba}.}
  The Area Under the Cumulative Curve (AUC) is reported at multiple angular error thresholds in blur-to-sharp and blur-to-blur settings.
  \colorbox{tabfirst}{First}, \colorbox{tabsecond}{second}, and \colorbox{tabthird}{third} best results are highlighted within sparse methods only.
  }
  \begin{tabular}{lcccccccc}
    \toprule
    & \multicolumn{4}{c}{Blur-to-Sharp} & \multicolumn{4}{c}{Blur-to-Blur} \\
    \cmidrule(l){2-5}\cmidrule(l){6-9}
    \multirow{3}[-5]{*}{Method}
    & \multicolumn{4}{c}{Pose AUC $\uparrow$}
    & \multicolumn{4}{c}{Pose AUC $\uparrow$} \\
    \cmidrule(l){2-5}\cmidrule(l){6-9}
    & $@5$\textdegree & $@10$\textdegree & $@20$\textdegree & $@30$\textdegree
    & $@5$\textdegree & $@10$\textdegree & $@20$\textdegree & $@30$\textdegree \\
    \midrule
    \multicolumn{9}{l}{\textit{Detector-free (Dense/Semi-dense)}} \\
    LoFTR~\cite{sun2021loftr}
        & 10.14 & 23.15 & 41.94 & 54.72
        & 6.64  & 19.53 & 41.54 & 52.25 \\
    MatchFormer~\cite{wang2022matchformer}
        & 2.50  & 4.32  & 10.45 & 18.39
        & 0.00  & 2.55  & 9.74  & 14.68 \\
    ASpanFormer~\cite{chen2022aspanformer}
        & 4.31  & 11.00 & 20.20 & 28.87
        & 3.56  & 7.40  & 15.85 & 22.31 \\
    RoMa v2~\cite{edstedt2025roma}
        & 21.45 & 38.50 & 59.01 & 70.76
        & 16.71 & 35.02 & 57.09 & 68.26 \\
    \midrule
    \multicolumn{9}{l}{\textit{Sparse (detector + learned matcher)}} \\
    SuperPoint~\cite{DeTone2018SuperPointSI} + LightGlue~\cite{lindenberger2023lightglue}
        & 14.86 & 26.50 & 40.72 & 51.07
        & 4.25  & 8.67  & 16.86 & 28.13 \\
    DISK~\cite{tyszkiewicz2020disk} + LightGlue~\cite{lindenberger2023lightglue}
        & 12.01 & 18.63 & 30.33 & 39.12
        & 2.47  & 9.36  & 17.14 & 22.54 \\
    ALIKED~\cite{Zhao2023ALIKED} + LightGlue~\cite{lindenberger2023lightglue}
        & 6.83  & 16.39 & 30.07 & 39.54
        & 5.99  & 14.31 & 27.56 & 36.63 \\
    \midrule
    \multicolumn{9}{l}{\textit{Sparse (detector + non-learned matching)}} \\
    SIFT~\cite{LoweDavid2004DistinctiveIF} + MNN
        & 9.28  & 19.60 & 34.44 & 45.92
        & 5.20  & 12.53 & 23.16 & 32.26 \\
    DoG~\cite{LoweDavid2004DistinctiveIF} + AffNet~\cite{AffNet2017} + HardNet~\cite{Mishchuk2017WorkingHT} + MNN
        & 8.56  & 16.29 & 27.48 & 35.58
        & 3.43  & 8.96  & 18.42 & 25.55 \\
    Key.Net~\cite{Laguna2019KeyNetKD} + AffNet~\cite{AffNet2017} + HardNet~\cite{Mishchuk2017WorkingHT} + MNN
        & 9.91  & 16.61 & 29.51 & 39.74
        & 5.53  & 13.88 & 24.36 & 33.64 \\
    SuperPoint~\cite{DeTone2018SuperPointSI} + MNN
        & 7.07  & 12.91 & 24.43 & 32.94
        & 3.73  & 6.57  & 11.28 & 17.17 \\
    D2-Net~\cite{Dusmanu2019D2NetAT} + MNN
        & 7.18  & 19.83 & 40.19 & 53.38
        & 4.90  & 17.01 & 37.16 & 50.05 \\
    R2D2~\cite{Revaud2019R2D2RA} + MNN
        & 14.26 & 23.48 & 38.37 & 48.00
        & 9.93  & 22.25 & 34.75 & 44.85 \\
    DISK~\cite{tyszkiewicz2020disk} + MNN
        & 6.98  & 14.90 & 24.59 & 31.02
        & 3.66  & 6.57  & 9.97  & 13.94 \\
    SiLK~\cite{gleize2023silk} + MNN
        & 0.00  & 0.00  & 0.67  & 3.34
        & 0.00  & 0.00  & 0.00  & 2.82  \\
    DeDoDe v2~\cite{edstedt2024dedodev2} + Dual-Softmax
        & 9.46  & 19.86 & 35.95 & 45.12
        & 5.87  & 11.10 & 17.85 & 24.07 \\
    BALF~\cite{zhao2024balf} + HardNet~\cite{Mishchuk2017WorkingHT} + MNN
        & \cellcolor{tabthird}14.89 & \cellcolor{tabthird}27.29 & \cellcolor{tabthird}47.84 & \cellcolor{tabthird}59.15
        & \cellcolor{tabfirst}12.63 & \cellcolor{tabsecond}25.17 & \cellcolor{tabthird}46.09 & \cellcolor{tabthird}58.42 \\
    BALF~\cite{zhao2024balf} + HyNet~\cite{hynet2020} + MNN
        & 14.29 & 26.57 & 45.17 & 58.44
        & 10.06 & 23.11 & 42.19 & 55.20 \\
    \cmidrule{2-9}
    SSMB (Ours) + HardNet~\cite{Mishchuk2017WorkingHT} + MNN
        & \cellcolor{tabsecond}15.08 & \cellcolor{tabsecond}27.34 & \cellcolor{tabsecond}48.67 & \cellcolor{tabsecond}63.01
        & \cellcolor{tabthird}10.99 & \cellcolor{tabfirst}25.61 & \cellcolor{tabsecond}46.95 & \cellcolor{tabsecond}60.24 \\
    SSMB (Ours) + HyNet~\cite{hynet2020} + MNN
        & \cellcolor{tabfirst}16.51 & \cellcolor{tabfirst}31.59 & \cellcolor{tabfirst}52.94 & \cellcolor{tabfirst}66.61
        & \cellcolor{tabsecond}11.07  & \cellcolor{tabthird}24.65 & \cellcolor{tabfirst}48.39 & \cellcolor{tabfirst}60.83 \\
    \bottomrule
  \end{tabular}
  \label{tab:pose_auc}
\end{table*}

\PAR{Dataset and protocol.}
We evaluate on the ArchViz dataset~\cite{liu2021mba} under both \emph{blur-to-sharp} and \emph{blur-to-blur} settings.
We use up to $2,048$ keypoints per image and report Pose AUC at $5^\circ$, $10^\circ$, $20^\circ$, and $30^\circ$ thresholds.

\PAR{Results.}
\cref{tab:pose_auc} shows that SSMB consistently achieves top-2 results among sparse keypoint detectors across both settings and all thresholds, with at least one of its two descriptor variants (HardNet or HyNet) ranking in the top two in every case.
In the \emph{blur-to-sharp} setting, SSMB+HyNet achieves an AUC of $66.61$ at $30^\circ$, the highest among all sparse methods and a substantial improvement over the next best result of $59.15$ (BALF+HardNet).
In the \emph{blur-to-blur} setting, SSMB+HyNet again leads sparse methods with $60.83$ at $30^\circ$, ahead of the next best sparse method, BALF+HardNet, at $58.42$.
The only exception to this trend occurs at the $5^\circ$ threshold in the \emph{blur-to-blur} setting, where BALF+HardNet ($12.63$) narrowly surpasses both SSMB variants ($11.07$ and $10.99$).
At this very strict threshold, pose accuracy is highly sensitive to a small number of high-quality correspondences, and both methods perform close to the noise floor of the benchmark.
The consistent performance of SSMB across both settings and the remaining seven threshold-setting combinations demonstrates that blur-robust detection translates directly to improved pose estimation accuracy.

\subsection{Visual Localization}

\begin{table}[t]
    \scriptsize
    \centering
    \renewcommand\arraystretch{1.1}
    \setlength{\tabcolsep}{-0.5pt}
    \caption{\textbf{Visual localization on the Aachen Day-Night benchmark~\cite{Sattler2018CVPR}.}
    The percentage of queries successfully localized within three error thresholds is reported.
    \colorbox{tabfirst}{First}, \colorbox{tabsecond}{second}, and \colorbox{tabthird}{third} best results are highlighted within sparse methods only.
    \textsuperscript{\dag}RoMa v2 could not be evaluated on this benchmark due to insufficient system memory during large-scale bundle adjustment ($4,328$ database images).}
    \resizebox{\linewidth}{!}{
        \begin{tabular}{lcc}
            \toprule
            & \multicolumn{2}{c}{Correctly Localized Queries (\%) $\uparrow$} \\
            \cmidrule{2-3}
            Method & Day & Night \\
            \cmidrule{2-3}
            & \multicolumn{2}{c}{(0.25m,2\textdegree) / (0.5m,5\textdegree) / (1.0m,10\textdegree)} \\
            \midrule
            \multicolumn{3}{l}{\textit{Detector-free (Dense/Semi-dense)}} \\
            LoFTR~\cite{sun2021loftr}
                & 49.3 / 63.3 / 80.1 & 21.4 / 41.8 / 67.3 \\
            MatchFormer~\cite{wang2022matchformer}
                & 6.9 / 24.0 / 76.0  & 3.1 / 14.3 / 49.0 \\
            ASpanFormer~\cite{chen2022aspanformer}
                & 43.3 / 57.2 / 76.5  & 12.4 / 29.6 / 58.4 \\
            RoMa v2~\cite{edstedt2025roma}\textsuperscript{\dag} & \textitgray{OOM} & \textitgray{OOM} \\
            \midrule
            \multicolumn{3}{l}{\textit{Sparse (detector + learned matcher)}} \\
            SuperPoint~\cite{DeTone2018SuperPointSI} + LightGlue~\cite{lindenberger2023lightglue}
                & 31.9 / 51.1 / 79.5 & 11.2 / 23.5 / 62.2 \\
            DISK~\cite{tyszkiewicz2020disk} + LightGlue~\cite{lindenberger2023lightglue}
                & 42.1 / 59.7 / 79.6 & 10.2 / 26.5 / 56.1 \\
            ALIKED~\cite{Zhao2023ALIKED} + LightGlue~\cite{lindenberger2023lightglue}
                & 39.3 / 55.2 / 77.5 & 18.4 / 28.6 / 60.2 \\
            \midrule
            \multicolumn{3}{l}{\textit{Sparse (detector + non-learned matching)}} \\
            SIFT~\cite{LoweDavid2004DistinctiveIF} + MNN
                & 49.7 / 59.8 / 68.8 & 13.3 / 18.4 / 28.6 \\
            Key.Net~\cite{Laguna2019KeyNetKD} + AffNet~\cite{AffNet2017} + HardNet~\cite{Mishchuk2017WorkingHT} + MNN
                & 35.7 / 52.3 / 74.6 & 7.1 / 23.5 / 40.8 \\
            SuperPoint~\cite{DeTone2018SuperPointSI} + MNN
                & 33.3 / 56.4 / 76.5 & 9.2 / 21.4 / 60.2 \\
            D2-Net~\cite{Dusmanu2019D2NetAT} + MNN
                & 10.6 / 26.8 / 74.4 & 3.1 / 12.2 / 57.1 \\
            R2D2~\cite{Revaud2019R2D2RA} + MNN
                & 47.5 / 60.9 / 78.4 & 20.1 / 35.7 / 60.3 \\
            DISK~\cite{tyszkiewicz2020disk} + MNN
                & 43.7 / 58.0 / 76.7 & 13.7 / 30.1 / 63.8 \\
            DeDoDe v2~\cite{edstedt2024dedodev2} + Dual-Softmax
                & 44.9 / 58.7 / 75.4 & 20.4 / 32.7 / 54.1 \\
            BALF~\cite{zhao2024balf} + HardNet~\cite{Mishchuk2017WorkingHT} + MNN
                & 51.0 / 68.0 / 89.2 & \colorbox{tabthird}{21.4} / \colorbox{tabthird}{36.7} / \colorbox{tabthird}{80.6} \\
            BALF~\cite{zhao2024balf} + HyNet~\cite{hynet2020} + MNN
                & \colorbox{tabthird}{51.2} / \colorbox{tabthird}{68.9} / \colorbox{tabthird}{89.4} & 15.3 / 30.6 / 72.4 \\
            \cmidrule{2-3}
            SSMB (Ours) + HardNet~\cite{Mishchuk2017WorkingHT} + MNN
                & \colorbox{tabsecond}{55.3} / \colorbox{tabsecond}{73.8} / \colorbox{tabsecond}{92.6}
                & \colorbox{tabsecond}{28.6} / \colorbox{tabsecond}{53.1} / \colorbox{tabsecond}{84.7} \\
            SSMB (Ours) + HyNet~\cite{hynet2020} + MNN
                & \colorbox{tabfirst}{58.5} / \colorbox{tabfirst}{76.2} / \colorbox{tabfirst}{93.0}
                & \colorbox{tabfirst}{31.6} / \colorbox{tabfirst}{59.2} / \colorbox{tabfirst}{86.7} \\
            \bottomrule
        \end{tabular}
    }
    \label{tab:visual_localization}
\end{table}

\PAR{Dataset and protocol.}
We evaluate on the Aachen Day-Night benchmark~\cite{Sattler2018CVPR} with motion blur on both query and database images.
We use the hierarchical localization framework~\cite{sarlin2019coarse} with NetVLAD~\cite{arandjelovic2016netvlad} retrieval (top-$20$ candidates), COLMAP~\cite{schonberger2016structure} for 3D reconstruction, and $4,096$ keypoints per image at $640 \times 480$ resolution.

\PAR{Results.}
As shown in \cref{tab:visual_localization}, SSMB outperforms all sparse baselines under both daytime and nighttime conditions.
For daytime queries, SSMB+HyNet achieves $58.5\%$, $76.2\%$, and $93.0\%$ at the three error thresholds, compared to $51.2\%$, $68.9\%$, and $89.4\%$ for the best competing sparse method, BALF+HyNet.
The improvement is even more pronounced for nighttime queries, where SSMB+HyNet achieves $31.6\%$, $59.2\%$, and $86.7\%$, above the next best result of $21.4\%$, $36.7\%$, and $80.6\%$, achieved by BALF+HardNet.
The larger nighttime gains are expected, as nighttime images naturally exhibit motion blur due to longer exposures, making blur-robust detection particularly beneficial in this condition.
Interestingly, BALF's stronger descriptor pairing differs by condition (HyNet for daytime, HardNet for nighttime), whereas SSMB+HyNet remains the best configuration in both conditions, suggesting that SSMB's detections are more consistently compatible with a single descriptor across illumination changes.
Among detector-free methods, LoFTR, MatchFormer, and ASpanFormer all trail SSMB at every threshold and condition (\textit{e.g.,} LoFTR achieves $21.4\%$, $41.8\%$, and $67.3\%$ at night, compared to SSMB's $31.6\%$, $59.2\%$, and $86.7\%$), despite operating on full image pairs rather than sparse keypoints.
RoMa v2 could not be evaluated on this benchmark, as its dense per-pair correspondence estimation exhausted the available system memory during bundle adjustment over this large-scale reconstruction ($4{,}328$ database images), consistent with its substantially higher per-pair computational cost reported in \cref{tab:efficiency}.
This result highlights the importance of blur-robust sparse detection even in comparison to dense matching approaches, particularly at the scale required by real-world visual localization deployments.

\subsection{Computational Efficiency}\label{ssec:efficiency}
 
\begin{table}[t]
    \scriptsize
    \centering
    \renewcommand\arraystretch{1.1}
    \setlength{\tabcolsep}{12pt}
    \caption{\textbf{Computational cost (in millisecond)} for keypoint detection on a single image, measured on an NVIDIA RTX 3090 GPU (Intel Xeon E5-2686 v4 CPU for SIFT).
    \textsuperscript{\dag}For D2-Net, detection and description cannot be separated in a single forward pass, so the reported time includes both.}
    \begin{tabular}{lcc}
        \toprule
        Method & 240$\times$320 pixels  $\downarrow$ & 480$\times$640 pixels  $\downarrow$\\
        \midrule
        \multicolumn{3}{l}{\textit{Detector-free (Dense/Semi-dense)}} \\
        LoFTR~\cite{sun2021loftr}                         & 30.35          & 75.22 \\
        MatchFormer~\cite{wang2022matchformer}            & 22.81          & 37.92 \\
        ASpanFormer~\cite{chen2022aspanformer}            & 53.22          & 108.54 \\
        RoMa v2~\cite{edstedt2025roma}                    & 486.64         & 497.96 \\
        \midrule
        \multicolumn{3}{l}{\textit{Sparse}} \\
        SIFT~\cite{LoweDavid2004DistinctiveIF}           & 12.97          & 57.18 \\
        Key.Net~\cite{Laguna2019KeyNetKD}                & 14.80           & 33.54  \\
        SuperPoint~\cite{DeTone2018SuperPointSI}         & \textbf{2.14}  & \textbf{3.40} \\
        D2-Net~\cite{Dusmanu2019D2NetAT}\textsuperscript{\dag} & 4.97     & 17.84 \\
        R2D2~\cite{Revaud2019R2D2RA}                     & 4.77           & 17.60 \\
        DISK~\cite{tyszkiewicz2020disk}                  & 4.89           & 13.18 \\
        REKD~\cite{lee2022self}                          & 7.86           & 27.18 \\
        NeSS-ST~\cite{pakulev2023ness}                 & 10.54          & 14.45 \\
        ALIKED~\cite{Zhao2023ALIKED}                     & 5.62           & 5.63  \\
        DeDoDe v2~\cite{edstedt2024dedodev2}             & 14.92          & 42.77 \\
        XFeat~\cite{potje2024cvpr}                       & 5.45           & 5.44  \\
        BALF~\cite{zhao2024balf}                         & 9.42           & 27.04 \\
        \midrule
        SSMB (Ours)                                 & 11.65          & 30.63 \\
        \bottomrule
    \end{tabular}
    \label{tab:efficiency}
\end{table}
 
\cref{tab:efficiency} reports detection-only inference time for all methods at two resolutions, $240 \times 320$ and $480 \times 640$ pixels, measured on an NVIDIA RTX 3090 GPU.
For methods that jointly predict descriptors alongside keypoints, we exclude the descriptor head from the forward pass to ensure a fair, detection-only comparison.
Detector-free methods have no separate detection step, so the reported time is for full pairwise dense/semi-dense matching on an image pair, which is not directly comparable to the detection-only timings of sparse methods.
SSMB runs at $11.65$ ms and $30.63$ ms, comfortably within real-time requirements ($33$ FPS at VGA resolution, \textit{i.e.,} $480 \times 640$ pixels) for robotic applications.
Compared to BALF, SSMB incurs a modest overhead ($30.63$ ms vs. $27.04$ ms at VGA resolution) due to the added LDE module, which is justified by the substantial gains in self-supervised training without external supervision (see \cref{tab:ablation}).

\subsection{Ablation Study}\label{sec:ablation}

We conduct ablation studies to analyze the contribution of each design choice in SSMB.
All ablations are evaluated on the Blur-HPatches \textsc{Tough} split under both \emph{blur-to-sharp} and \emph{blur-to-blur} settings, reporting repeatability.

\PAR{Loss components.}
\cref{tab:ablation} ablates each loss term by removing it while keeping all others.
Removing $\mathcal{L}_{\text{div}}$ causes the largest performance drop, from $77.16\%$ to $64.07\%$ in \emph{blur-to-sharp} and from $74.48\%$ to $65.26\%$ in \emph{blur-to-blur}, confirming that spatial diversity is the most critical component despite its small weight ($\lambda_{\text{div}} = 0.005$).
Without $\mathcal{L}_{\text{div}}$, detection again collapses onto a handful of salient locations during GoPro training, even after geometric pretraining.
Removing $\mathcal{L}_{\text{blur}}$ causes the second largest drop (to $65.81\%$ and $67.84\%$), demonstrating that explicit cross-domain consistency supervision is essential for blur robustness.
Removing $\mathcal{L}_{\text{ha}}$ leads to a moderate drop (to $69.03\%$ and $69.78\%$), as the HA loss anchors detection to geometrically stable locations on sharp images.
Removing $\mathcal{L}_{\text{pos}}$ has the smallest effect ($72.69\%$ and $73.77\%$), indicating that sub-pixel position consistency provides only a marginal benefit under the repeatability metric used throughout this evaluation.
Since this loss only enforces consistency between the sharp and blurred branches without an absolute geometric target, we additionally verify that the offset head does not collapse to a degenerate constant field.
Applying the predicted offset $\boldsymbol{\delta}$ to the final keypoint coordinates provides a modest but consistent improvement on the \textsc{Tough} split in both settings, from $75.74\%$ to $77.16\%$ in \emph{blur-to-sharp} and from $73.89\%$ to $74.48\%$ in \emph{blur-to-blur}, confirming that the offset head learns non-trivial sub-pixel corrections.

\begin{table}[t]
    \scriptsize
    \centering
    \renewcommand\arraystretch{1.1}
    \setlength{\tabcolsep}{10pt}
    \caption{\textbf{Ablation study.}
    Repeatability is reported on the Blur-HPatches \textsc{Tough} split~\cite{zhao2024balf} under blur-to-sharp and blur-to-blur settings.
    }
        \begin{tabular}{lcc}
            \toprule
            & \multicolumn{2}{c}{Repeatability (\%)} \\
            \cmidrule{2-3}
            Variant & Blur-to-Sharp & Blur-to-Blur \\
            \midrule
            \multicolumn{3}{l}{\textit{Loss components}} \\
            w/o $\mathcal{L}_{\text{ha}}$   & 69.03 & 69.78 \\
            w/o $\mathcal{L}_{\text{blur}}$ & 65.81 & 67.84 \\
            w/o $\mathcal{L}_{\text{pos}}$  & 72.69 & 73.77 \\
            w/o $\mathcal{L}_{\text{div}}$  & 64.07 & 65.26 \\
            \midrule
            \multicolumn{3}{l}{\textit{Training pipeline}} \\
            w/o Synthetic pretraining       & 59.58 & 61.44 \\
            \midrule
            \multicolumn{3}{l}{\textit{Architecture}} \\
            w/o LDE                        & 33.83 & 50.13 \\
            \midrule
            Full (ours)                     & \textbf{77.16} & \textbf{74.48} \\
            \bottomrule
        \end{tabular}
    \label{tab:ablation}
\end{table}

\PAR{Multi-stage training.}
Skipping synthetic pretraining leads to a dramatic drop from $77.16\%$ to $59.58\%$ in \emph{blur-to-sharp} repeatability and from $74.48\%$ to $61.44\%$ in \emph{blur-to-blur} repeatability.
This confirms that geometric pretraining is an essential prerequisite. Without it, the network cannot escape score map collapse during blur-aware training, and the blur consistency loss has no stable detection signal to align.

\begin{figure}[t]
    \centering
    \includegraphics[width=0.98\columnwidth]{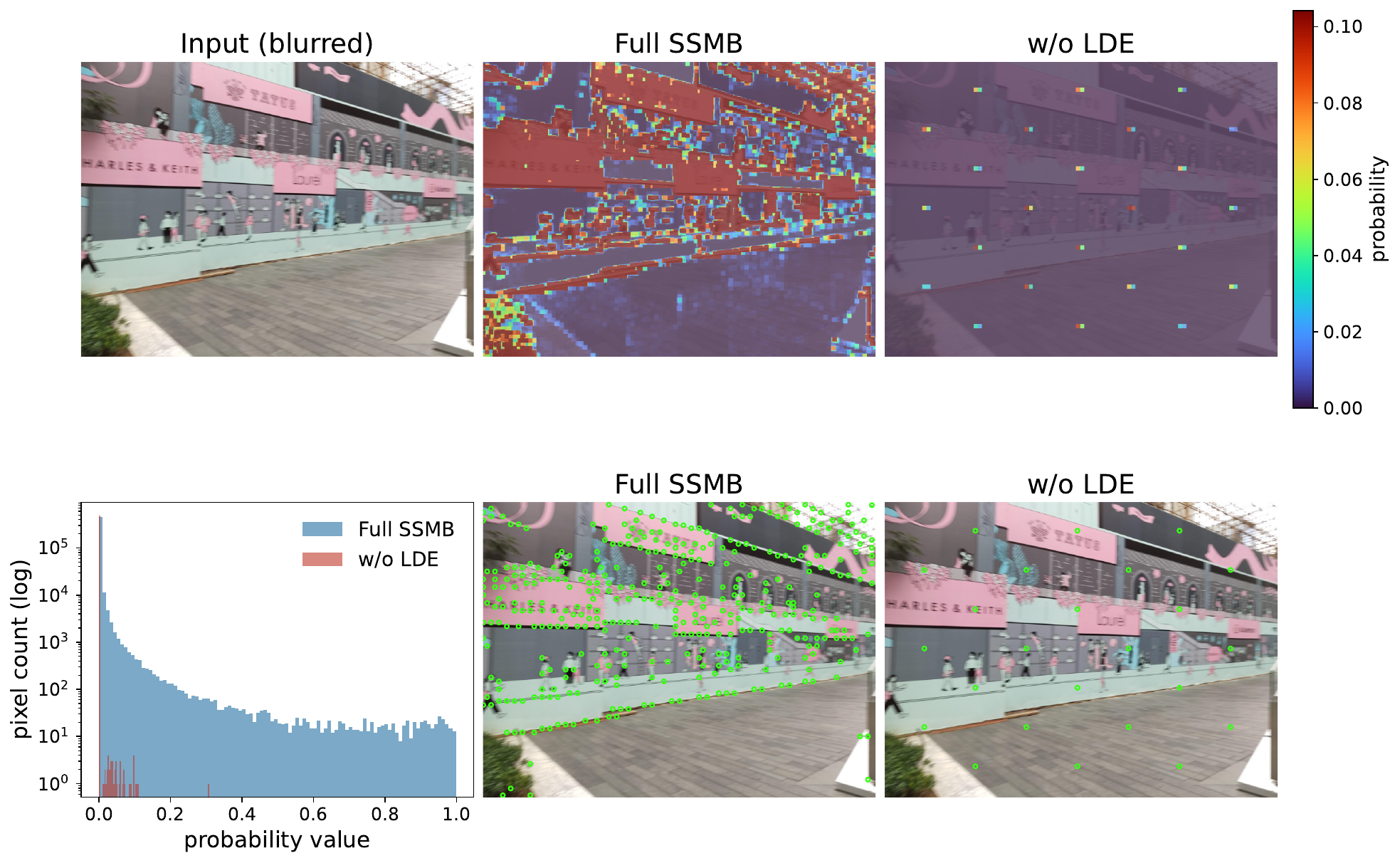}
    \caption{\textbf{Effect of the LDE module on the predicted probability map.}
    \emph{Top row:} probability heatmaps on the same real-world blurred image from the RWBI dataset~\cite{zhang2020deblurring}, overlaid on a lightened version of the input.
    Full SSMB produces dense, well-localized responses that closely trace real structure such as storefront signage, arched window frames, and shop dividers.
    Without LDE, the response collapses toward near-zero almost everywhere, with only a handful of residual peaks remaining.
    \emph{Bottom row:} the pixel-value histogram of the probability map, together with the extracted keypoints (green circles) for each variant.
    Full SSMB's keypoints densely and precisely trace the storefront geometry.
    In contrast, the w/o LDE variant yields far fewer keypoints overall, which no longer trace real structure.
    Most fall on featureless regions such as bare pavement, with a spacing that resembles a uniform grid rather than genuine edge localization, consistent with the near-flat residual field revealed by the histogram.
    This visually explains the dramatic repeatability drop observed in \cref{tab:ablation} (from $77.16\%$ to $33.83\%$ on the \emph{blur-to-sharp} \textsc{Tough} setting).
    Without local discriminability, the network can no longer anchor its predictions to repeatable geometric structure.
    }
    \label{fig:lde_ablation}
\end{figure}

\PAR{LDE module.}
Removing LDE causes a dramatic drop from $77.16\%$ to $33.83\%$ in \emph{blur-to-sharp} repeatability and from $74.48\%$ to $50.13\%$ in \emph{blur-to-blur} repeatability, at a parameter saving of only $0.05$M.
This confirms that LDE is the most impactful architectural component in SSMB.
Notably, the w/o LDE variant is architecturally identical to the BALF-style backbone~\cite{zhao2024balf}, trained with our self-supervised objective rather than SIFT supervision, showing that this backbone alone is insufficient to support stable self-supervised training under blur.
This is because, without LDE, the global MLP mixing in the \emph{GridGmlpLayer} destroys the local gradient structure that distinguishes repeatable keypoint locations under blur, leaving the network unable to identify discriminative spatial positions.
\cref{fig:lde_ablation} visualizes this effect directly.
The response collapses toward near-zero almost everywhere, and the few residual peaks that remain no longer trace real structure, falling instead on featureless regions with a spacing that resembles a uniform grid rather than genuine edge localization.

\subsection{Qualitative Results}\label{ssec:realworld}

\begin{figure*}[t]
    \centering
    {\scriptsize
    \begin{minipage}[t]{0.24\linewidth}
        \centering
        \includegraphics[width=\linewidth]{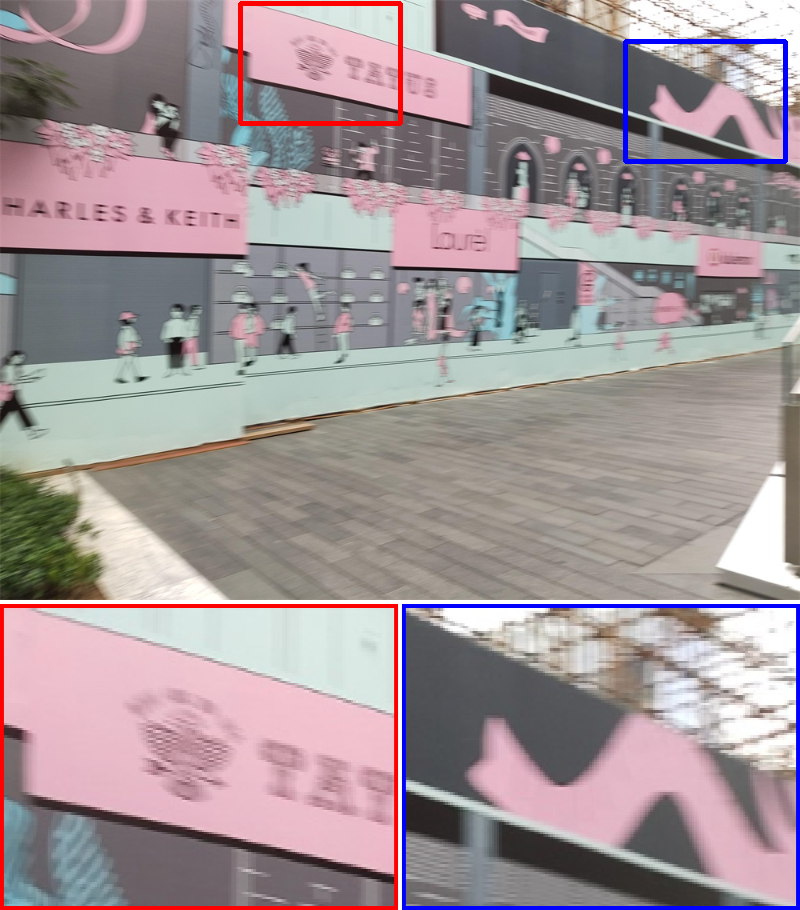}\\[1mm]
        Input
    \end{minipage}
    \hspace{0.5mm}
    \begin{minipage}[t]{0.24\linewidth}
        \centering
        \includegraphics[width=\linewidth]{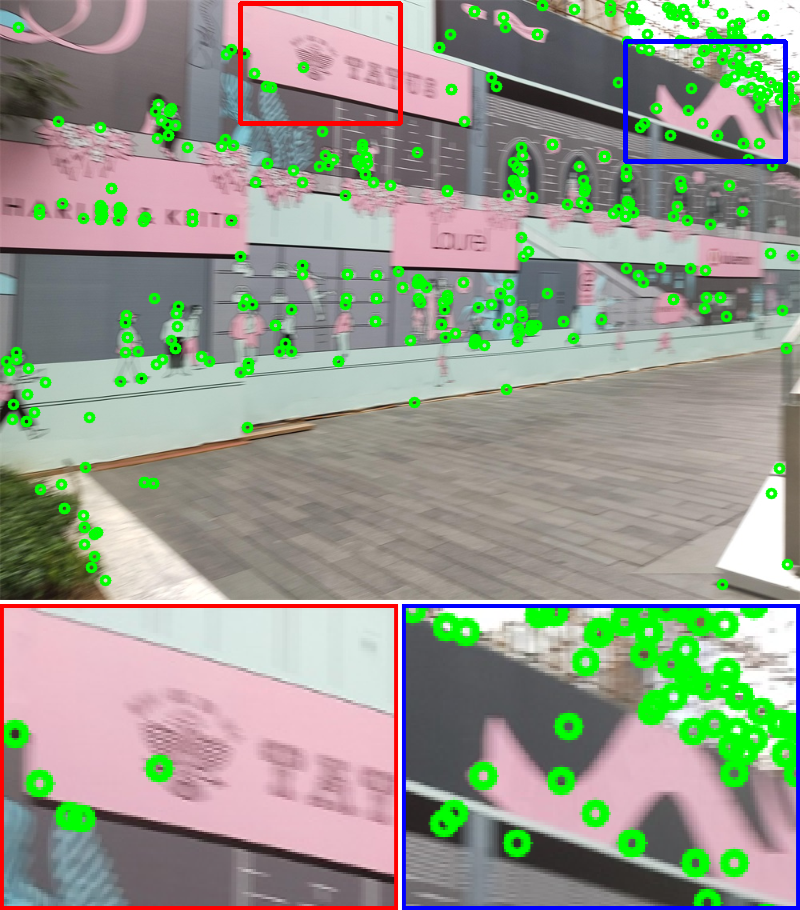}\\[1mm]
        SIFT~\cite{LoweDavid2004DistinctiveIF}
    \end{minipage}
    \hspace{0.5mm}
    \begin{minipage}[t]{0.24\linewidth}
        \centering
        \includegraphics[width=\linewidth]{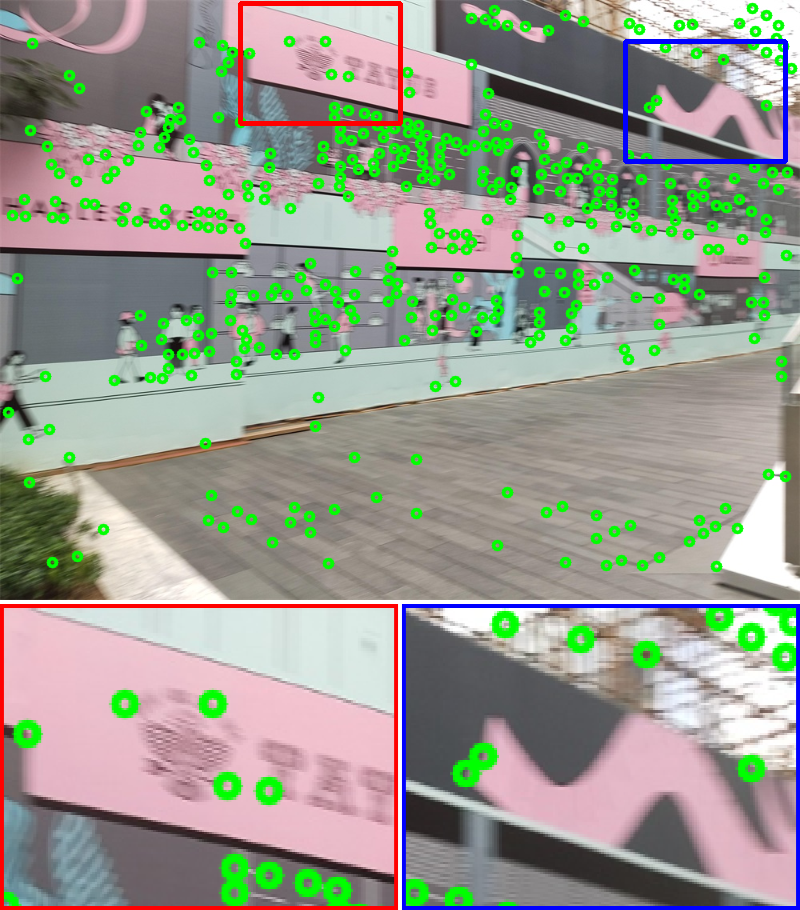}\\[1mm]
        SuperPoint~\cite{DeTone2018SuperPointSI}
    \end{minipage}
    \hspace{0.5mm}
    \begin{minipage}[t]{0.24\linewidth}
        \centering
        \includegraphics[width=\linewidth]{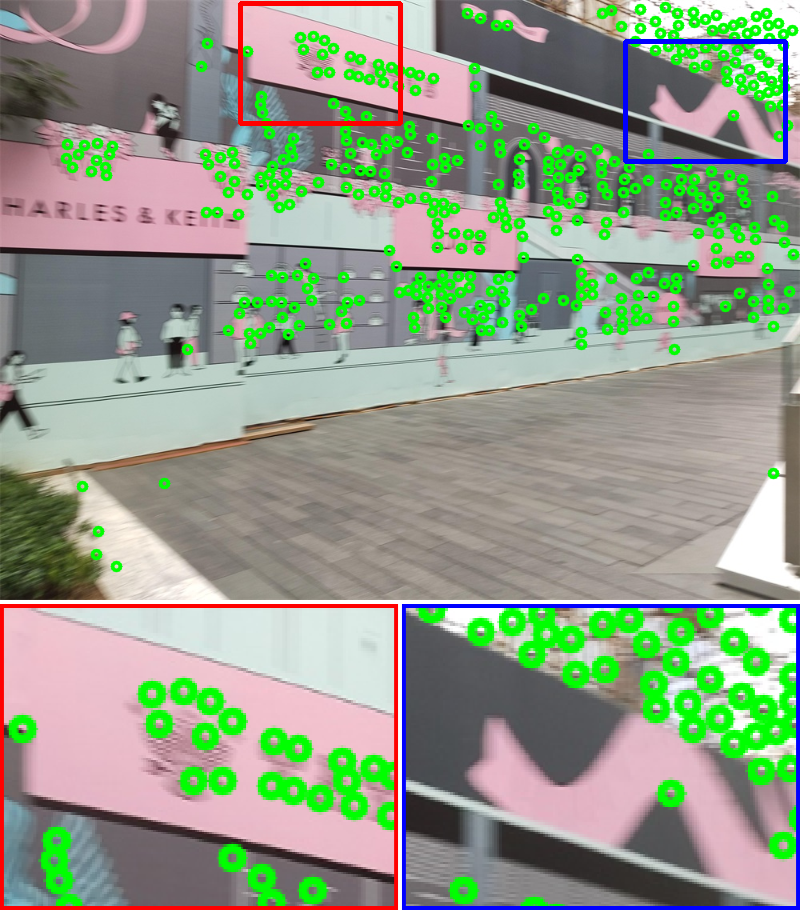}\\[1mm]
        DISK~\cite{tyszkiewicz2020disk}
    \end{minipage}
    
    \vspace{4mm}
    
    \begin{minipage}[t]{0.24\linewidth}
        \centering
        \includegraphics[width=\linewidth]{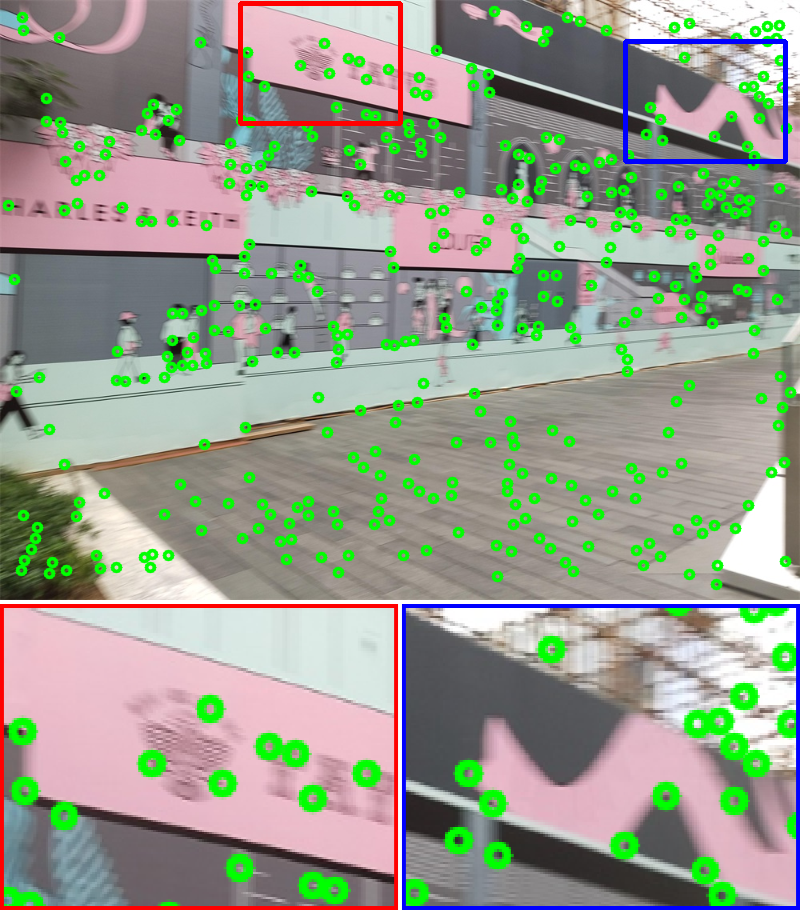}\\[1mm]
        NeSS-ST~\cite{pakulev2023ness}
    \end{minipage}
    \hspace{0.5mm}
    \begin{minipage}[t]{0.24\linewidth}
        \centering
        \includegraphics[width=\linewidth]{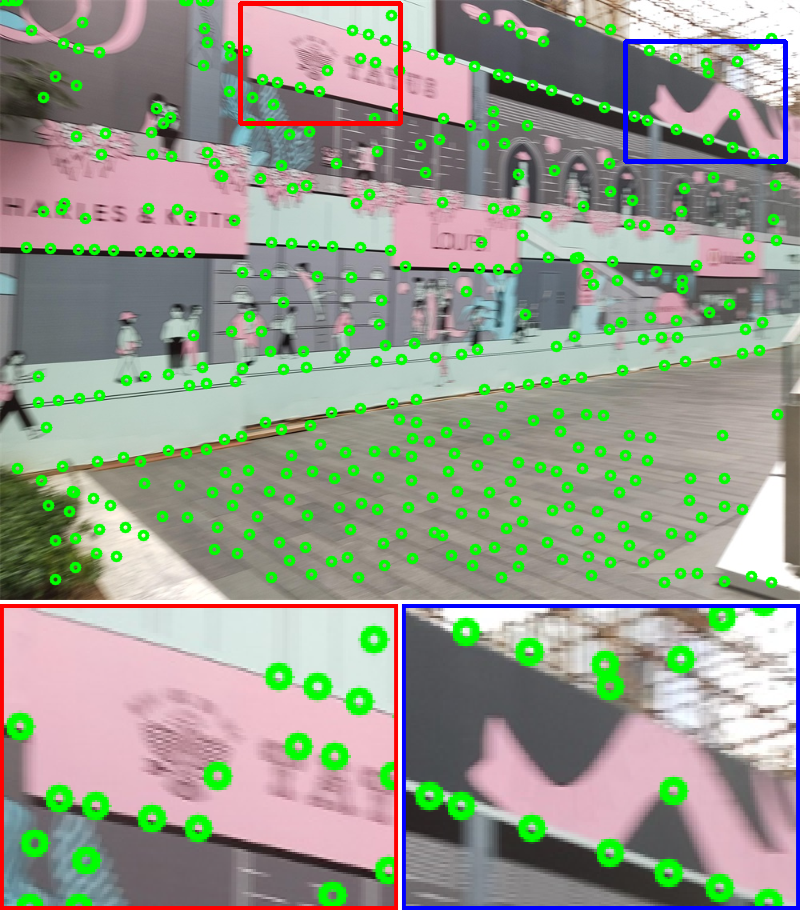}\\[1mm]
        ALIKED~\cite{Zhao2023ALIKED}
    \end{minipage}
    \hspace{0.5mm}
    \begin{minipage}[t]{0.24\linewidth}
        \centering
        \includegraphics[width=\linewidth]{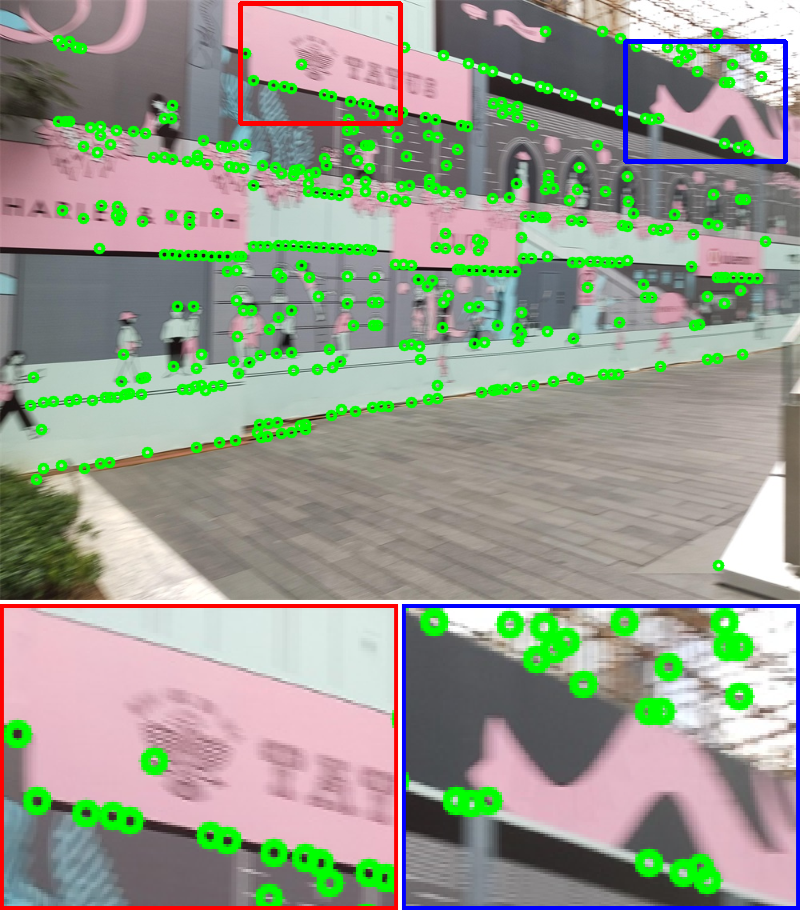}\\[1mm]
        BALF~\cite{zhao2024balf}
    \end{minipage}
    \hspace{0.5mm}
    \begin{minipage}[t]{0.24\linewidth}
        \centering
        \includegraphics[width=\linewidth]{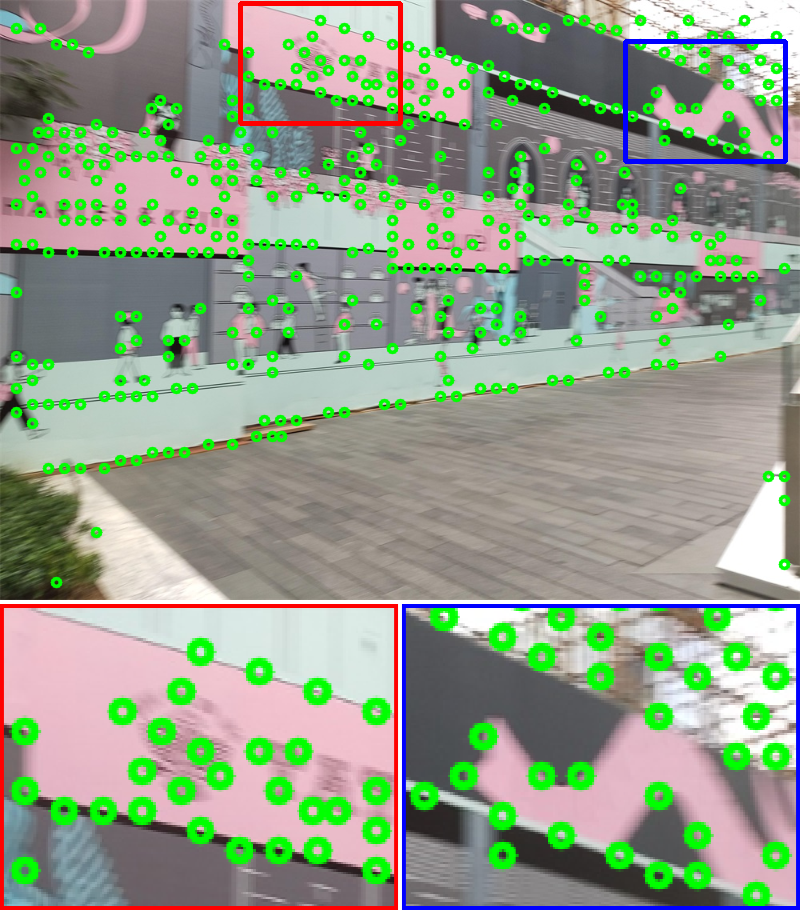}\\[1mm]
        \textbf{SSMB (Ours)}
    \end{minipage}
    }
    \caption{\textbf{Qualitative detection results on real-world blurred images from the RWBI dataset~\cite{zhang2020deblurring}.}
    SSMB detects well-localized and consistently distributed keypoints under real camera motion blur, while several baselines either miss salient structure or produce noisy responses in heavily blurred regions.
    The \textcolor{red}{red} and \textcolor{blue}{blue} boxes mark two representative regions, shown enlarged below each image for closer comparison.
    Best viewed in color.}
    \label{fig:real_detection}
\end{figure*}

\PAR{Detection on RWBI.}
\cref{fig:real_detection} presents keypoint detection results on a real-world blurred image from the RWBI dataset~\cite{zhang2020deblurring}, captured with real cameras rather than synthetic blur.
This is the same image used earlier to visualize the effect of LDE in \cref{fig:lde_ablation}.
SSMB produces well-localized keypoints that remain concentrated on genuine image structure, whereas SIFT and DISK detect comparatively sparser responses in heavily blurred regions, and SuperPoint, NeSS-ST, and ALIKED scatter detections more uniformly without consistently aligning to salient structure.
Compared to BALF, SSMB produces more accurate keypoints around text regions and other heavily blurred structures, despite being trained without any SIFT-derived supervision.
The enlarged regions highlight this difference clearly, particularly in areas with fine text or high-frequency edge structure that are otherwise easy to overlook at full-image scale.
Additional keypoint detection results on the RealBlur dataset~\cite{Rim2020RealWorldBD} are provided in \cref{fig:real_detection_appendix} of the Appendix.

\begin{figure}[t]
    \centering
    \includegraphics[width=0.95\columnwidth]{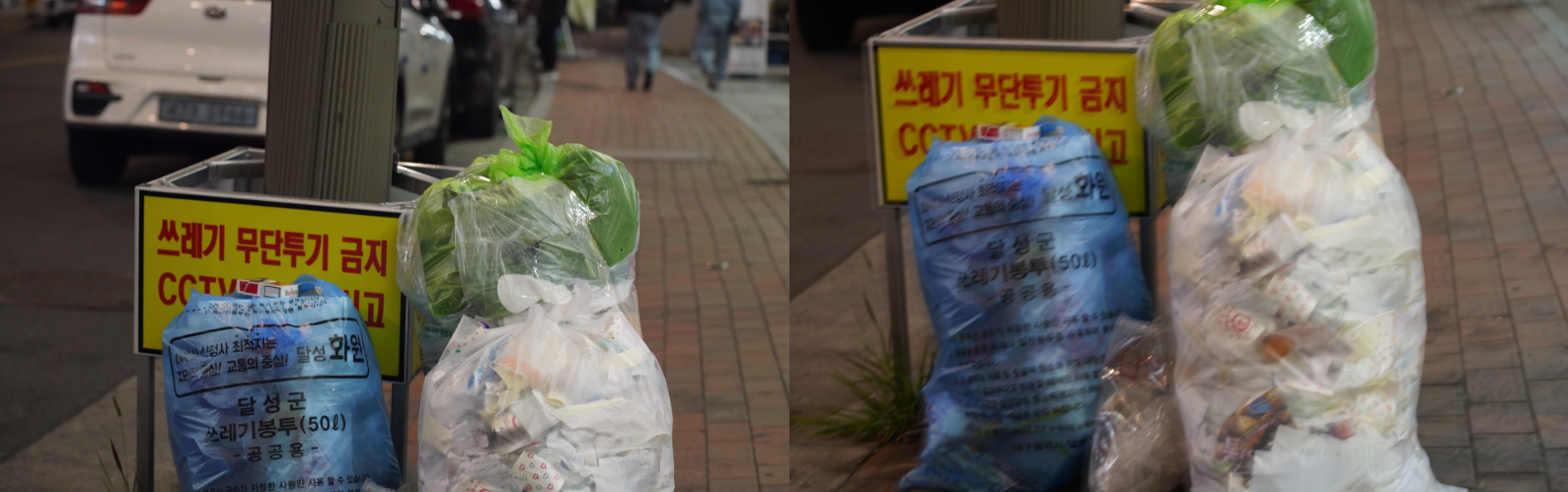}\\[1mm]
    \includegraphics[width=0.95\columnwidth]{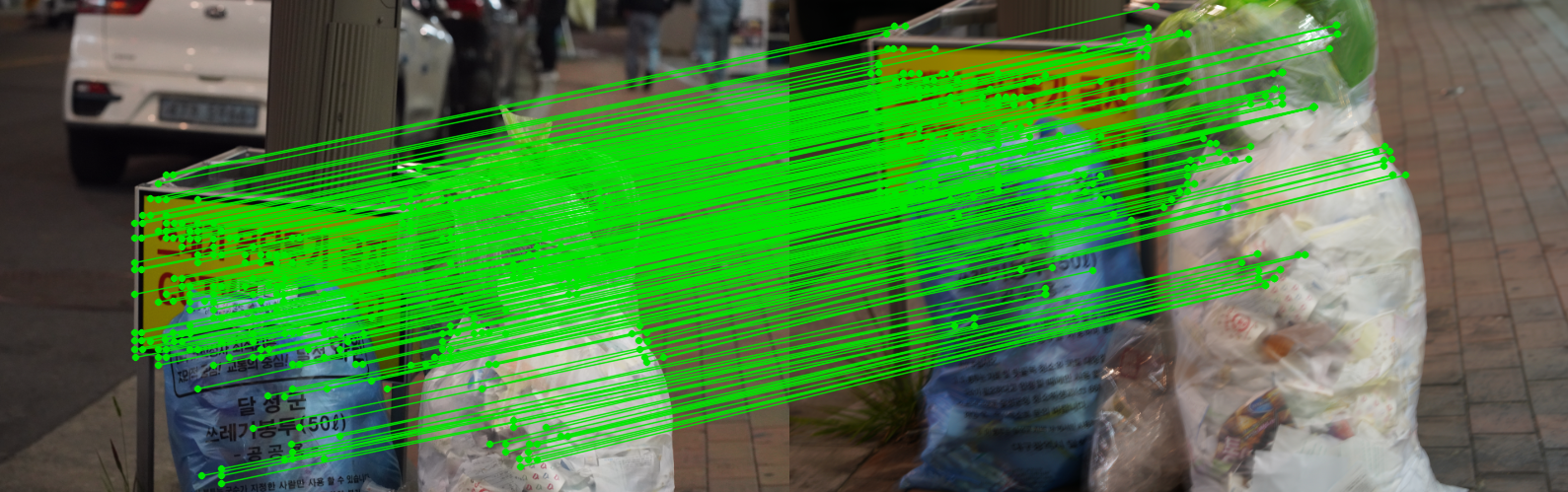}
    \caption{\textbf{Qualitative matching results on a real-world sharp-blur image pair from the RealBlur dataset~\cite{Rim2020RealWorldBD}.}
    \emph{Top:} the original image pair (left: sharp, right: blurred).
    \emph{Bottom:} SSMB detects well-localized and repeatable keypoints from both images for matching under real camera motion blur and viewpoint changes.
    Green lines indicate correct matches.
    Best viewed in color.}
    \label{fig:real_match}
\end{figure}

\PAR{Matching on RealBlur.}
\cref{fig:real_match} presents a qualitative feature matching result on a real-world sharp-blur image pair from the RealBlur dataset~\cite{Rim2020RealWorldBD}, using SSMB with HardNet~\cite{Mishchuk2017WorkingHT} and MNN matching.
The pair exhibits real camera motion blur, together with a viewpoint change between the sharp and blurred images.
SSMB produces well-localized and repeatable keypoints from both images despite the cross-domain appearance gap, yielding dense and geometrically consistent matches, confirming that the gains on synthetic benchmarks transfer to real camera blur.
Additional matching results across indoor and outdoor scenes are provided in \cref{fig:real_match_appendix} of the Appendix.
\section{Conclusion}\label{sec:conclusion}
We present SSMB, a self-supervised keypoint detector for motion-blurred images that requires no external supervision.
The key architectural contribution is the Local Discriminability Enhancement (LDE), which restores fine-grained local discriminability after global feature mixing.
The training pipeline proceeds in two stages, where geometric pretraining on synthetic shapes bootstraps spatial discriminability, and blur-aware training on real sharp-blur pairs reinforces it through a multi-component self-supervised loss.
Both stages are individually necessary and, in particular, the spatial diversity loss term proves essential for preventing the network from degenerating during training on blurred images.
Evaluations across keypoint detection, image matching, relative pose estimation, and visual localization tasks demonstrate that SSMB achieves state-of-the-art performance among sparse detectors under motion blur.
\section*{Acknowledgments}
This work was supported by Spanish grants PID2021-127685NB-I00 and PID2024-155886NB-I00 and Arag{\'o}n grant T45\_23R.

\newpage

\clearpage

\appendix
In this appendix, we provide additional analysis and experimental results that are omitted from the main paper due to space limitations:
\begin{itemize}
    \item Complete keypoint detection results on the original, all-sharp HPatches dataset, together with the disaggregated viewpoint and illumination repeatability for Blur-HPatches and Deblur-HPatches (see \cref{subsec:keypoint_detection_appendix}).
    \item Additional image matching results across the full range of pixel thresholds, complementing the single-threshold comparison reported in the main paper (see \cref{subsec:image_matching_appendix}).
    \item Additional qualitative results for relative pose estimation on the ArchViz dataset (see  \cref{subsec:relpose_appendix}).
    \item Additional qualitative detection and matching results on real-world blurred images from the RealBlur dataset (see \cref{subsec:more_real_results}).
\end{itemize}

\subsection{Keypoint Detection}\label{subsec:keypoint_detection_appendix}

Due to limited space in the main paper, we only report the overall repeatability in \cref{tab:blur_HPatches,tab:deblur_blur_hpatches} of the main paper, and omit results on the original all-sharp HPatches dataset~\cite{Balntas2017HPatchesAB} entirely.
We present the complete results here, including the viewpoint and illumination repeatability separately.

\PAR{Results on HPatches (sharp images).}
\cref{tab:sharp_hpatches} shows repeatability on the original HPatches dataset with all-sharp images.
SSMB achieves $75.17\%$ overall repeatability, ranking second among all methods and outperforming BALF ($70.28\%$).
This result is notable given that SSMB is specifically designed for blurred images and trained exclusively on GoPro blur-sharp pairs without any sharp-image-specific supervision.
Homographic adaptation provides invariance to planar image warps, but does not fully model the 3D viewpoint changes, parallax, and occlusion present in HPatches' viewpoint sequences, which explains the more modest viewpoint repeatability of $67.50\%$ compared to methods such as REKD ($80.94\%$) and ALIKED ($72.85\%$) that are specifically designed for geometric invariance on sharp images.
Meanwhile, the blur consistency objective and photometric augmentations may favor robustness to illumination and blur variations, consistent with the strong illumination repeatability of $83.11\%$.
Motion blur primarily degrades fine spatial structure rather than overall photometric appearance, so learning to be consistent across blur levels naturally generalizes to illumination changes as well.

\begin{table}[t]
    \scriptsize
    \centering
    \renewcommand\arraystretch{1.3}
    \setlength{\tabcolsep}{10pt}
    \caption{\textbf{Keypoint detection repeatability on the HPatches dataset~\cite{Balntas2017HPatchesAB}.}
    For each method, we report the average repeatability score for viewpoint changes, illumination changes, and for all image sequences.}
    \begin{tabular}{lcc|c}
        \toprule
        & \multicolumn{3}{c}{\begin{tabular}[c]{@{}c@{}}Reference: Sharp\\Target: Sharp\end{tabular}}\\
        \cmidrule{2-4}
        Method & Viewpoint $\uparrow$ & Illumination $\uparrow$ & Overall $\uparrow$
        \\ \midrule
        SIFT~\cite{LoweDavid2004DistinctiveIF}          & 60.29 & 60.44 & 60.36
        \\
        Key.Net~\cite{Laguna2019KeyNetKD}               & 68.99 & 67.47 & 68.24 \\
        SuperPoint~\cite{DeTone2018SuperPointSI}        & \cellcolor{tabthird}69.53 & 68.92 & 69.23
        \\
        LF-Net~\cite{Ono2018LFNetLL}                    & 68.41 & \cellcolor{tabthird}73.61 & 70.96
        \\
        D2-Net~\cite{Dusmanu2019D2NetAT}                & 53.99 & 62.80 & 58.32
        \\
        R2D2~\cite{Revaud2019R2D2RA}                    & 61.68 & 61.93 & 61.80
        \\
        DISK~\cite{tyszkiewicz2020disk}                 & 67.22 & 66.68 & 66.95
        \\
        REKD~\cite{lee2022self}                         & \cellcolor{tabfirst}80.94 & \cellcolor{tabsecond}76.29 & \cellcolor{tabfirst}78.65 \\
        NeSS-ST~\cite{pakulev2023ness}                & 64.45 & 64.98 & 64.71
        \\
        ALIKED~\cite{Zhao2023ALIKED}                    & \cellcolor{tabsecond}72.85 & 70.37 & \cellcolor{tabthird}71.63
        \\
        DeDoDe v2~\cite{edstedt2024dedodev2}            & 61.95 & 61.47 & 61.71
        \\
        XFeat~\cite{potje2024cvpr}                      & 59.07 & 64.61 & 61.79
        \\
        BALF~\cite{zhao2024balf}                        & 67.21 & 73.51 & 70.28
        \\
        \midrule
        SSMB (Ours)                                & 67.50 & \cellcolor{tabfirst}83.11 & \cellcolor{tabsecond}75.17 \\
        \bottomrule
    \end{tabular}
    \label{tab:sharp_hpatches}
\end{table}
\begin{table*}
    \scriptsize
    \centering
    \renewcommand\arraystretch{1.3}
    \setlength{\tabcolsep}{8pt}
    \caption{\textbf{Keypoint detection repeatability on the Blur-HPatches dataset under the blur-to-sharp setting.}
    SSMB achieves the highest illumination repeatability and ranks first in overall repeatability at the \textsc{Tough} level.}
    \begin{tabular}{l|ccc|ccc|ccc}
        \toprule
        & \multicolumn{3}{c|}{E\begin{tiny}ASY\end{tiny}} & \multicolumn{3}{c|}{H\begin{tiny}ARD\end{tiny}} & \multicolumn{3}{c}{T\begin{tiny}OUGH\end{tiny}} \\
        Method & Viewpoint $\uparrow$ & Illumination $\uparrow$ & Total $\uparrow$ & Viewpoint $\uparrow$ & Illumination $\uparrow$ & Total $\uparrow$ & Viewpoint $\uparrow$ & Illumination $\uparrow$ & Total $\uparrow$\\
        \midrule
        SIFT~\cite{LoweDavid2004DistinctiveIF} & 50.64 & 61.39 & 55.92 & 51.55 & 62.23 & 56.80 & 47.36 & 59.83 & 53.49\\
        Key.Net~\cite{Laguna2019KeyNetKD} & 58.02 & 62.74 & 60.34 & 50.54 & 59.02 & 54.71 & 39.20 & 50.38 & 44.69\\
        SuperPoint~\cite{DeTone2018SuperPointSI} & \cellcolor{tabthird}66.06 & 65.21 & \cellcolor{tabthird}65.64 & \cellcolor{tabthird}61.96 & 62.49 & \cellcolor{tabthird}62.22 & 51.60 & 54.13 & 52.84\\
        LF-Net~\cite{Ono2018LFNetLL} & 59.57 & 67.65 & 63.54 & 57.64 & 64.87 & 61.19 & 52.14 & 61.57 & \cellcolor{tabthird}56.78 \\
        D2-Net~\cite{Dusmanu2019D2NetAT} & 44.30 & 55.31 & 49.71 & 41.60 & 53.19 & 47.30 & 38.39 & 50.47 & 44.32\\
        R2D2~\cite{Revaud2019R2D2RA} & 55.10 & 60.99 & 57.99 & 47.37 & 56.25 & 51.73 & 33.46 & 47.92 & 40.57\\
        DISK~\cite{tyszkiewicz2020disk} & 60.42 & 60.05 & 60.24 & 58.43 & 57.91 & 58.18 & \cellcolor{tabthird}55.81 & 56.13 & 55.97 \\
        REKD~\cite{lee2022self} & 34.53 & \cellcolor{tabsecond}76.50 & 55.15 & 31.87 & \cellcolor{tabthird}75.18 & 53.15 & 27.87 & \cellcolor{tabthird}69.89 & 48.52 \\
        NeSS-ST~\cite{pakulev2023ness} & 57.68 & 58.19 & 57.93 & 56.09 & 56.15 & 56.12 & 53.85 & 53.81 & 53.83 \\
        ALIKED~\cite{Zhao2023ALIKED} & 60.49 & 57.21 & 58.88 & 56.19 & 51.97 & 54.12 & 52.08 & 48.58 & 50.36\\
        DeDoDe v2~\cite{edstedt2024dedodev2} & 52.71 & 52.85 & 52.78 & 49.17 & 49.66 & 49.41 & 45.94 & 47.64 & 46.78 \\
        XFeat~\cite{potje2024cvpr} & 53.56 & 56.13 & 54.82 & 50.38 & 52.83 & 51.59 & 48.15 & 51.03 & 49.56 \\
        BALF~\cite{zhao2024balf} & \cellcolor{tabfirst}72.58 & \cellcolor{tabthird}75.74 & \cellcolor{tabsecond}74.12 & \cellcolor{tabfirst}72.93 & \cellcolor{tabsecond}76.07 & \cellcolor{tabsecond}74.45 & \cellcolor{tabfirst}67.26 & \cellcolor{tabsecond}76.54 & \cellcolor{tabsecond}71.84 \\
        \midrule
        SSMB (Ours) & \cellcolor{tabsecond}66.29 & \cellcolor{tabfirst}88.57 & \cellcolor{tabfirst}77.24 & \cellcolor{tabsecond}66.23 & \cellcolor{tabfirst}88.51 & \cellcolor{tabfirst}77.18 & \cellcolor{tabsecond}66.22 & \cellcolor{tabfirst}88.48 & \cellcolor{tabfirst}77.16 \\
        \bottomrule
    \end{tabular}
	\label{tab:blur_Hpatches_blur_to_sharp}
\end{table*}

\begin{table*}
    \scriptsize
    \centering
    \renewcommand\arraystretch{1.3}
    \setlength{\tabcolsep}{8pt}
    \caption{\textbf{Keypoint detection repeatability on the Blur-HPatches dataset under the blur-to-blur setting.}
    SSMB achieves the highest illumination and overall repeatability across all difficulty levels.}
    \begin{tabular}{l|ccc|ccc|ccc}
        \toprule
        & \multicolumn{3}{c|}{E\begin{tiny}ASY\end{tiny}} & \multicolumn{3}{c|}{H\begin{tiny}ARD\end{tiny}} & \multicolumn{3}{c}{T\begin{tiny}OUGH\end{tiny}}\\
        Method & Viewpoint $\uparrow$ & Illumination $\uparrow$ & Total $\uparrow$ & Viewpoint $\uparrow$ & Illumination $\uparrow$ & Total $\uparrow$ & Viewpoint $\uparrow$ & Illumination $\uparrow$ & Total $\uparrow$\\
        \midrule
        SIFT~\cite{LoweDavid2004DistinctiveIF} & 56.67 & 57.31 & 56.99 & 53.15 & 53.85 & 53.49 & 46.23 & 46.63 & 45.94\\
        Key.Net~\cite{Laguna2019KeyNetKD} & 61.81 &  63.77 & \cellcolor{tabthird}62.77 & 56.82 & 59.57 & 58.17 & 45.68 & 52.94 & 49.25\\
        SuperPoint~\cite{DeTone2018SuperPointSI} & 58.01 & 59.22 & 58.60 & 49.69 & 50.37 & 50.03 & 42.34 & 44.25 & 43.28\\
        LF-Net~\cite{Ono2018LFNetLL} & 52.45 & 68.74 & 60.45 & 51.20 & \cellcolor{tabthird}67.21 & 59.07 & 49.68 & \cellcolor{tabthird}66.02 & 57.71 \\
        D2-Net~\cite{Dusmanu2019D2NetAT} & 46.65 & 57.14 & 51.80 & 45.84 & 56.44 & 51.05 & 45.11 & 56.13 & 50.53\\
        R2D2~\cite{Revaud2019R2D2RA} & 54.10 & 61.00 & 57.49 & 51.87 & 58.87 & 55.31 & 40.88 & 53.05 & 46.86\\
        DISK~\cite{tyszkiewicz2020disk} & 62.34 & 61.43 & 61.89 & 60.81 & 60.94 & \cellcolor{tabthird}60.87 & \cellcolor{tabthird}60.89 & 60.40 & \cellcolor{tabthird}60.65 \\
        REKD~\cite{lee2022self} & 34.28 & \cellcolor{tabthird}71.51 & 52.57 & 31.20 & 66.50 & 48.54 & 27.24 & 61.39 & 44.02 \\
        NeSS-ST~\cite{pakulev2023ness} & 61.18 & 60.74 & 60.97 & 60.07 & 59.53 & 59.80 & 59.42 & 59.00 & 59.21 \\
        ALIKED~\cite{Zhao2023ALIKED} & \cellcolor{tabthird}64.48 & 59.82 & 62.19 & \cellcolor{tabthird}61.64 & 57.28 & 59.49 & 58.90 & 56.19 & 57.57 \\
        DeDoDe v2~\cite{edstedt2024dedodev2} & 61.88 & 58.41 & 60.17 & 59.40 & 55.31 & 57.39 & 56.28 & 53.66 & 54.99 \\
        XFeat~\cite{potje2024cvpr} & 58.79 & 60.64 & 59.70 & 57.30 & 59.32 & 58.29 & 56.77 & 58.76 & 57.74 \\
        BALF~\cite{zhao2024balf} & \cellcolor{tabfirst}69.44 & \cellcolor{tabsecond}71.56 & \cellcolor{tabsecond}70.48 & \cellcolor{tabfirst}67.13 & \cellcolor{tabsecond}70.22 & \cellcolor{tabsecond}68.43 & \cellcolor{tabfirst}65.90 & \cellcolor{tabsecond}69.60 & \cellcolor{tabsecond}67.71 \\
        \midrule
        SSMB (Ours) & \cellcolor{tabsecond}64.60 & \cellcolor{tabfirst}84.14 & \cellcolor{tabfirst}74.20 & \cellcolor{tabsecond}64.66 & \cellcolor{tabfirst}84.34 & \cellcolor{tabfirst}74.33 & \cellcolor{tabsecond}64.75 & \cellcolor{tabfirst}84.55 & \cellcolor{tabfirst}74.48 \\
        \bottomrule
    \end{tabular}
	\label{tab:blur_HPatches_blur_to_blur}
\end{table*}

\PAR{Results on Blur-HPatches.}
\cref{tab:blur_Hpatches_blur_to_sharp,tab:blur_HPatches_blur_to_blur} report the detailed repeatability scores under \emph{blur-to-sharp} and \emph{blur-to-blur} settings, respectively.
SSMB's viewpoint repeatability remains stable around $64$--$67\%$ across all difficulty levels and both settings, while its illumination repeatability stays above $88\%$ under the \emph{blur-to-sharp} setting (peaking at $88.57\%$ on \textsc{Easy}) and above $84\%$ under the \emph{blur-to-blur} setting (peaking at $84.55\%$ on \textsc{Tough}).
This consistency confirms that the illumination-viewpoint trade-off discussed above originates from the blur consistency training objective itself, rather than from any particular difficulty level or evaluation setting.
Compared against BALF, the only other deblur-free method in this evaluation, SSMB outperforms it across all difficulty levels under both the \emph{blur-to-sharp} and \emph{blur-to-blur} settings on Blur-HPatches.
On Blur-HPatches \emph{blur-to-blur}, SSMB outperforms BALF by $3.72$ on \textsc{Easy}, $5.90$ on \textsc{Hard}, and $6.77$ on \textsc{Tough}, a margin that widens with increasing blur severity.
This widening margin reflects the benefit of training directly on cross-domain consistency between sharp and blurred images, rather than regressing fixed SIFT pseudo-labels that do not adapt to blur severity.

\begin{table*}
    \scriptsize
    \centering
    \renewcommand\arraystretch{1.3}
    \setlength{\tabcolsep}{8pt}
    \caption{\textbf{Keypoint detection repeatability on deblurred images from SRN-DeblurNet~\cite{Tao2018ScaleRecurrentNF} under deblur-to-sharp setting.}
    The bottom row shows the results of SSMB on the corresponding blurred images.}
    \begin{tabular}{l|ccc|ccc|ccc}
        \toprule
        & \multicolumn{3}{c|}{E\begin{tiny}ASY\end{tiny}} & \multicolumn{3}{c|}{H\begin{tiny}ARD\end{tiny}} & \multicolumn{3}{c}{T\begin{tiny}OUGH\end{tiny}}\\
        Method & Viewpoint $\uparrow$ & Illumination $\uparrow$ & Total $\uparrow$ & Viewpoint $\uparrow$ & Illumination $\uparrow$ & Total $\uparrow$ & Viewpoint $\uparrow$ & Illumination $\uparrow$ & Total $\uparrow$\\
        \midrule
        SIFT~\cite{LoweDavid2004DistinctiveIF} & 53.98 & 59.35 & 56.62 & 52.44 & 58.38 & 55.36 & 47.95 & 59.92 & 53.83\\
        Key.Net~\cite{Laguna2019KeyNetKD} & 62.86 & 63.72 & 63.28 & 56.37 & 59.71 & 58.01 & 42.08 & 52.30 & 47.10\\
        SuperPoint~\cite{DeTone2018SuperPointSI} & \cellcolor{tabsecond}68.65 & 66.76 & \cellcolor{tabthird}67.72 & 65.33 & 62.73 & \cellcolor{tabthird}64.05 & 54.18 & 56.37 & 55.26\\
        LF-Net~\cite{Ono2018LFNetLL} & 54.40 & 70.31 & 62.22 & 52.52 & 67.53 & 59.90 & 47.77 & 61.93 & 54.73\\
        D2-Net~\cite{Dusmanu2019D2NetAT} & 46.72 & 57.08 & 51.81 & 44.30 & 54.87 & 49.49 & 40.00 & 52.07 & 45.94 \\
        R2D2~\cite{Revaud2019R2D2RA} & 58.36 & 62.31 & 60.31 & 52.58 & 58.38 & 55.43 & 37.40 & 49.32 & 43.26\\
        DISK~\cite{tyszkiewicz2020disk} & 64.06 & 62.79 & 63.44 & 62.13 & 61.32 & 61.74 & \cellcolor{tabthird}57.30 & 57.05 & \cellcolor{tabthird}57.18 \\
        REKD~\cite{lee2022self} & 36.08 & \cellcolor{tabthird}74.96 & 55.18 & 34.88 & \cellcolor{tabthird}73.52 & 53.87 & 29.96 & \cellcolor{tabthird}69.73 & 49.50 \\
        NeSS-ST~\cite{pakulev2023ness} & 61.41 & 61.13 & 61.27 & 59.75 & 59.46 & 59.61 & 55.64 & 55.10 & 55.37 \\
        ALIKED~\cite{Zhao2023ALIKED} & \cellcolor{tabthird}68.29 & 64.11 & 66.24 & \cellcolor{tabthird}65.77 & 61.18 & 63.51 & 57.14 & 52.71 & 54.96 \\
        DeDoDe v2~\cite{edstedt2024dedodev2} & 57.51 & 56.07 & 56.80 & 55.80 & 54.47 & 55.15 & 48.26 & 47.48 & 47.88 \\
        XFeat~\cite{potje2024cvpr} & 57.33 & 59.89 & 58.59 & 56.56 & 58.47 & 57.50 & 51.58 & 53.40 & 52.47 \\
        \midrule
        BALF~\cite{zhao2024balf} & \cellcolor{tabfirst}72.58 & \cellcolor{tabsecond}75.74 & \cellcolor{tabsecond}74.12 & \cellcolor{tabfirst}72.93 & \cellcolor{tabsecond}76.07 & \cellcolor{tabsecond}74.45 & \cellcolor{tabfirst}67.26 & \cellcolor{tabsecond}76.54 & \cellcolor{tabsecond}71.84 \\
        SSMB (Ours) & 66.29 & \cellcolor{tabfirst}88.57 & \cellcolor{tabfirst}77.24 & \cellcolor{tabsecond}66.23 & \cellcolor{tabfirst}88.51 & \cellcolor{tabfirst}77.18 & \cellcolor{tabsecond}66.22 & \cellcolor{tabfirst}88.48 & \cellcolor{tabfirst}77.16 \\
        \bottomrule
    \end{tabular}
	\label{tab:deblur_blur_hpatches_srn_deblur_to_sharp}
\end{table*}

\begin{table*}
    \scriptsize
    \centering
    \renewcommand\arraystretch{1.3}
    \setlength{\tabcolsep}{8pt}
    \caption{\textbf{Keypoint detection repeatability on deblurred images from SRN-DeblurNet~\cite{Tao2018ScaleRecurrentNF} under deblur-to-deblur setting.}
    The bottom row shows the results of SSMB on the corresponding blurred images.}
    \begin{tabular}{l|ccc|ccc|ccc}
        \toprule
        & \multicolumn{3}{c|}{E\begin{tiny}ASY\end{tiny}} & \multicolumn{3}{c|}{H\begin{tiny}ARD\end{tiny}} & \multicolumn{3}{c}{T\begin{tiny}OUGH\end{tiny}}\\
        Method & Viewpoint $\uparrow$ & Illumination $\uparrow$ & Total $\uparrow$ & Viewpoint $\uparrow$ & Illumination $\uparrow$ & Total $\uparrow$ & Viewpoint $\uparrow$ & Illumination $\uparrow$ & Total $\uparrow$\\
        \midrule
        SIFT~\cite{LoweDavid2004DistinctiveIF} & 60.31 & 59.16 & 59.75 & 57.98 & 58.27 & 58.13 & 48.56 & 52.78 & 50.63\\
        Key.Net~\cite{Laguna2019KeyNetKD} & 62.71 & 63.01 & 62.86 & 59.69 & 61.22 & 60.44 & 46.57 & 55.06 & 50.74\\
        SuperPoint~\cite{DeTone2018SuperPointSI} & \cellcolor{tabthird}67.77 & 64.93 & 66.38 & 64.61 & 61.67 & 63.16 & 49.04 & 50.02 & 49.52\\
        LF-Net~\cite{Ono2018LFNetLL} & 55.36 & 71.03 & 63.06 &  54.60 & 69.72 & 62.03 & 49.17 & \cellcolor{tabthird}65.68 & 57.28\\
        D2-Net~\cite{Dusmanu2019D2NetAT} & 49.05 & 58.32 & 53.60 & 48.58 & 57.57 & 53.00 & 45.70 & 56.33 & 50.93\\
        R2D2~\cite{Revaud2019R2D2RA} & 55.71 & 60.60 & 58.11 & 52.30 & 57.38 & 54.80 & 40.83 & 50.88 & 45.77\\
        DISK~\cite{tyszkiewicz2020disk} & 65.02 & 63.15 & 64.10 & 64.13 & 62.76 & 63.46 & 59.52 & 59.40 & \cellcolor{tabthird}59.46 \\
        REKD~\cite{lee2022self} & 35.65 & \cellcolor{tabsecond}73.44 & 54.22 & 34.01 & \cellcolor{tabsecond}70.59 & 51.99 & 27.89 & 63.42 & 45.35 \\
        NeSS-ST~\cite{pakulev2023ness} & 63.36 & 62.86 & 63.11 & 62.63 & 61.72 & 62.18 & 59.90 & 58.74 & 59.33 \\
        ALIKED~\cite{Zhao2023ALIKED} & \cellcolor{tabfirst}71.35 & 65.45 & \cellcolor{tabthird}68.45 & \cellcolor{tabfirst}69.43 & 62.99 & \cellcolor{tabthird}66.26 & \cellcolor{tabthird}60.62 & 55.68 & 58.19 \\
        DeDoDe v2~\cite{edstedt2024dedodev2} & 63.73 & 59.54 & 61.67 & 63.24 & 58.45 & 60.89 & 55.10 & 50.65 & 52.91 \\
        XFeat~\cite{potje2024cvpr} & 60.72 & 62.67 & 61.68 & 60.66 & 61.65 & 61.14 & 57.54 & 58.51 & 58.01 \\
        \midrule
        BALF~\cite{zhao2024balf} & \cellcolor{tabsecond}69.44 & \cellcolor{tabthird}71.56 & \cellcolor{tabsecond}70.48 & \cellcolor{tabsecond}67.13 & \cellcolor{tabthird}70.22 & \cellcolor{tabsecond}68.43 & \cellcolor{tabfirst}65.90 & \cellcolor{tabsecond}69.60 & \cellcolor{tabsecond}67.71 \\
        SSMB (Ours) & 64.60 & \cellcolor{tabfirst}84.14 & \cellcolor{tabfirst}74.20 & \cellcolor{tabthird}64.66 & \cellcolor{tabfirst}84.34 & \cellcolor{tabfirst}74.33 & \cellcolor{tabsecond}64.75 & \cellcolor{tabfirst}84.55 & \cellcolor{tabfirst}74.48 \\
        \bottomrule
    \end{tabular}
	\label{tab:deblur_blur_hpatches_srn_deblur_to_deblur}
\end{table*}
\begin{table*}
    \scriptsize
    \centering
    \renewcommand\arraystretch{1.3}
    \setlength{\tabcolsep}{8pt}
    \caption{\textbf{Keypoint detection repeatability on deblurred images from DeblurGAN-v2~\cite{Kupyn2019DeblurGANv2D} under deblur-to-sharp setting.}
    The bottom row shows the results of SSMB on the corresponding blurred images.}
    \begin{tabular}{l|ccc|ccc|ccc}
        \toprule
        & \multicolumn{3}{c|}{E\begin{tiny}ASY\end{tiny}} & \multicolumn{3}{c|}{H\begin{tiny}ARD\end{tiny}} & \multicolumn{3}{c}{T\begin{tiny}OUGH\end{tiny}}\\
        Method & Viewpoint $\uparrow$ & Illumination $\uparrow$ & Total $\uparrow$ & Viewpoint $\uparrow$ & Illumination $\uparrow$ & Total $\uparrow$ & Viewpoint $\uparrow$ & Illumination $\uparrow$ & Total $\uparrow$\\
        \midrule
        SIFT~\cite{LoweDavid2004DistinctiveIF} & 56.28 & 59.03 & 57.63 & 54.02 & 59.10 & 56.52 & 51.29 & 61.88 & 56.50\\
        Key.Net~\cite{Laguna2019KeyNetKD} & 63.78 & 64.20 & 63.99 & 57.46 & 60.91 & 59.16 & 44.92 & 53.92 & 49.35\\
        SuperPoint~\cite{DeTone2018SuperPointSI} & \cellcolor{tabsecond}68.85 & 67.02 & \cellcolor{tabthird}67.95 & \cellcolor{tabsecond}66.83 & 64.84 & \cellcolor{tabthird}65.86 & 57.37 & 59.10 & \cellcolor{tabthird}58.22\\
        LF-Net~\cite{Ono2018LFNetLL} & 55.01 & 70.43 & 62.59 & 53.03 & 67.70 & 60.24 & 47.73 & 62.14 & 54.81\\
        D2-Net~\cite{Dusmanu2019D2NetAT} & 47.68 & 57.77 & 52.64 & 44.96 & 55.64 & 50.21 & 40.17 & 51.80 & 45.88\\
        R2D2~\cite{Revaud2019R2D2RA} & 58.60 & 62.38 & 60.46 & 52.72 & 58.74 & 55.68 & 40.55 & 50.38 & 45.38\\
        DISK~\cite{tyszkiewicz2020disk} & 63.65 & 62.85 & 63.25 & 62.08 & 61.42 & 61.76 & 58.61 & 57.61 & 58.12 \\
        REKD~\cite{lee2022self} & 35.87 & \cellcolor{tabthird}75.34 & 55.27 & 34.29 & \cellcolor{tabthird}74.03 & 53.82 & 30.46 & \cellcolor{tabthird}71.54 & 50.65 \\
        NeSS-ST~\cite{pakulev2023ness} & 61.24 & 61.56 & 61.40 & 59.87 & 59.89 & 59.88 & 56.33 & 56.22 & 56.27 \\
        ALIKED~\cite{Zhao2023ALIKED} & \cellcolor{tabthird}68.47 & 65.09 & 66.81 & 66.05 & 62.21 & 64.16 & \cellcolor{tabthird}59.21 & 55.29 & 57.28 \\
        DeDoDe v2~\cite{edstedt2024dedodev2} & 58.14 & 57.35 & 57.75 & 56.36 & 55.99 & 56.18 & 51.47 & 50.75 & 51.12 \\
        XFeat~\cite{potje2024cvpr} & 57.69 & 60.63 & 59.13 & 56.60 & 58.98 & 57.77 & 52.47 & 54.24 & 53.34 \\
        \midrule
        BALF~\cite{zhao2024balf} & \cellcolor{tabfirst}72.58 & \cellcolor{tabsecond}75.74 & \cellcolor{tabsecond}74.12 & \cellcolor{tabfirst}72.93 & \cellcolor{tabsecond}76.07 & \cellcolor{tabsecond}74.45 & \cellcolor{tabfirst}67.26 & \cellcolor{tabsecond}76.54 & \cellcolor{tabsecond}71.84 \\
        SSMB (Ours) & 66.29 & \cellcolor{tabfirst}88.57 & \cellcolor{tabfirst}77.24 & \cellcolor{tabthird}66.23 & \cellcolor{tabfirst}88.51 & \cellcolor{tabfirst}77.18 & \cellcolor{tabsecond}66.22 & \cellcolor{tabfirst}88.48 & \cellcolor{tabfirst}77.16 \\
        \bottomrule
    \end{tabular}
	\label{tab:deblur_blur_hpatches_deblurganv2_deblur_to_sharp}
\end{table*}

\begin{table*}
    \scriptsize
    \centering
    \renewcommand\arraystretch{1.3}
    \setlength{\tabcolsep}{8pt}
    \caption{\textbf{Keypoint detection repeatability on deblurred images from DeblurGAN-v2~\cite{Kupyn2019DeblurGANv2D} under deblur-to-deblur setting.}
    The bottom row shows the results of SSMB on the corresponding blurred images.}
    \begin{tabular}{l|ccc|ccc|ccc}
        \toprule
        & \multicolumn{3}{c|}{E\begin{tiny}ASY\end{tiny}} & \multicolumn{3}{c|}{H\begin{tiny}ARD\end{tiny}} & \multicolumn{3}{c}{T\begin{tiny}OUGH\end{tiny}}\\
        Method & Viewpoint $\uparrow$ & Illumination $\uparrow$ & Total $\uparrow$ & Viewpoint $\uparrow$ & Illumination $\uparrow$ & Total $\uparrow$ & Viewpoint $\uparrow$ & Illumination $\uparrow$ & Total $\uparrow$\\
        \midrule
        SIFT~\cite{LoweDavid2004DistinctiveIF} & 59.17 & 59.73 & 59.44 & 57.33 & 58.66 & 57.98 & 50.12 & 52.33 & 51.21\\
        Key.Net~\cite{Laguna2019KeyNetKD} & 62.38 & 63.08 & 62.73 & 59.39 & 61.81 & 60.58 & 49.12 & 56.93 & 52.96\\
        SuperPoint~\cite{DeTone2018SuperPointSI} & \cellcolor{tabthird}67.49 & 65.49 & 66.50 & \cellcolor{tabthird}64.89 & 62.48 & 63.71 & 51.85 & 52.35 & 52.09\\
        LF-Net~\cite{Ono2018LFNetLL} & 55.57 & 70.69 & 63.00 & 54.58 & 69.25 & 61.79 & 50.48 & \cellcolor{tabthird}65.49 & 57.85\\
        D2-Net~\cite{Dusmanu2019D2NetAT} & 49.40 & 58.62 & 53.93 & 48.65 & 58.08 & 53.29 & 45.69 & 55.97 & 50.74\\
        R2D2~\cite{Revaud2019R2D2RA} & 55.66 & 60.33 & 57.95 & 52.31 & 57.84 & 55.03 & 43.12 & 52.76 & 47.86\\
        DISK~\cite{tyszkiewicz2020disk} & 64.33 & 63.37 & 63.86 & 62.98 & 62.74 & 62.86 & 60.06 & 59.66 & 59.86 \\
        REKD~\cite{lee2022self} & 35.17 & \cellcolor{tabsecond}73.21 & 53.86 & 33.62 & \cellcolor{tabsecond}70.23 & 51.61 & 28.74 & 63.03 & 45.59 \\
        NeSS-ST~\cite{pakulev2023ness} & 62.72 & 62.65 & 62.69 & 62.45 & 61.55 & 62.01 & 59.95 & 58.80 & 59.39 \\
        ALIKED~\cite{Zhao2023ALIKED} & \cellcolor{tabfirst}70.38 & 65.59 & \cellcolor{tabthird}68.03 & \cellcolor{tabfirst}68.62 & 63.87 & \cellcolor{tabthird}66.29 & \cellcolor{tabthird}62.70 & 58.79 & \cellcolor{tabthird}60.78 \\
        DeDoDe v2~\cite{edstedt2024dedodev2} & 62.48 & 60.17 & 61.34 & 62.20 & 59.25 & 60.75 & 58.64 & 55.23 & 56.97 \\
        XFeat~\cite{potje2024cvpr} & 59.95 & 62.96 & 61.43 & 59.81 & 61.96 & 60.87 & 58.10 & 59.08 & 58.59 \\
        \midrule
        BALF~\cite{zhao2024balf} & \cellcolor{tabsecond}69.44 & \cellcolor{tabthird}71.56 & \cellcolor{tabsecond}70.48 & \cellcolor{tabsecond}67.13 & \cellcolor{tabthird}70.22 & \cellcolor{tabsecond}68.43 & \cellcolor{tabfirst}65.90 & \cellcolor{tabsecond}69.60 & \cellcolor{tabsecond}67.71 \\
        SSMB (Ours) & 64.60 & \cellcolor{tabfirst}84.14 & \cellcolor{tabfirst}74.20 & 64.66 & \cellcolor{tabfirst}84.34 & \cellcolor{tabfirst}74.33 & \cellcolor{tabsecond}64.75 & \cellcolor{tabfirst}84.55 & \cellcolor{tabfirst}74.48 \\
        \bottomrule
    \end{tabular}
	\label{tab:deblur_blur_hpatches_deblurganv2_deblur_to_deblur}
\end{table*}

\PAR{Results on Deblur-HPatches.}
\cref{tab:deblur_blur_hpatches_srn_deblur_to_sharp,tab:deblur_blur_hpatches_srn_deblur_to_deblur} report results using SRN-DeblurNet~\cite{Tao2018ScaleRecurrentNF} for restoration, and \cref{tab:deblur_blur_hpatches_deblurganv2_deblur_to_sharp,tab:deblur_blur_hpatches_deblurganv2_deblur_to_deblur} report results using DeblurGAN-v2~\cite{Kupyn2019DeblurGANv2D}.
The bottom row in each table shows SSMB's repeatability on the corresponding blurred images, without any deblurring preprocessing, for direct comparison against the deblur-then-detect pipeline.
The gap between SSMB and the deblur-then-detect pipeline widens as blur severity increases, since deblurring artifacts become more pronounced under \textsc{Hard} and \textsc{Tough} conditions and increasingly degrade the restored image structure that downstream detectors rely on.
This is most evident on \textsc{Tough} sequences, where SSMB's overall repeatability ($77.16\%$ \emph{blur-to-sharp}, $74.48\%$ \emph{blur-to-blur}) exceeds every deblur-then-detect combination, including the strongest ones, DeblurGAN-v2 with SuperPoint ($58.22\%$ under \emph{deblur-to-sharp} setting) and DeblurGAN-v2 with ALIKED ($60.78\%$ under \emph{deblur-to-deblur} setting).

\subsection{Image Matching}\label{subsec:image_matching_appendix}

\begin{figure*}
    \begin{minipage}{0.55\textwidth}
      \includegraphics[width=\textwidth]{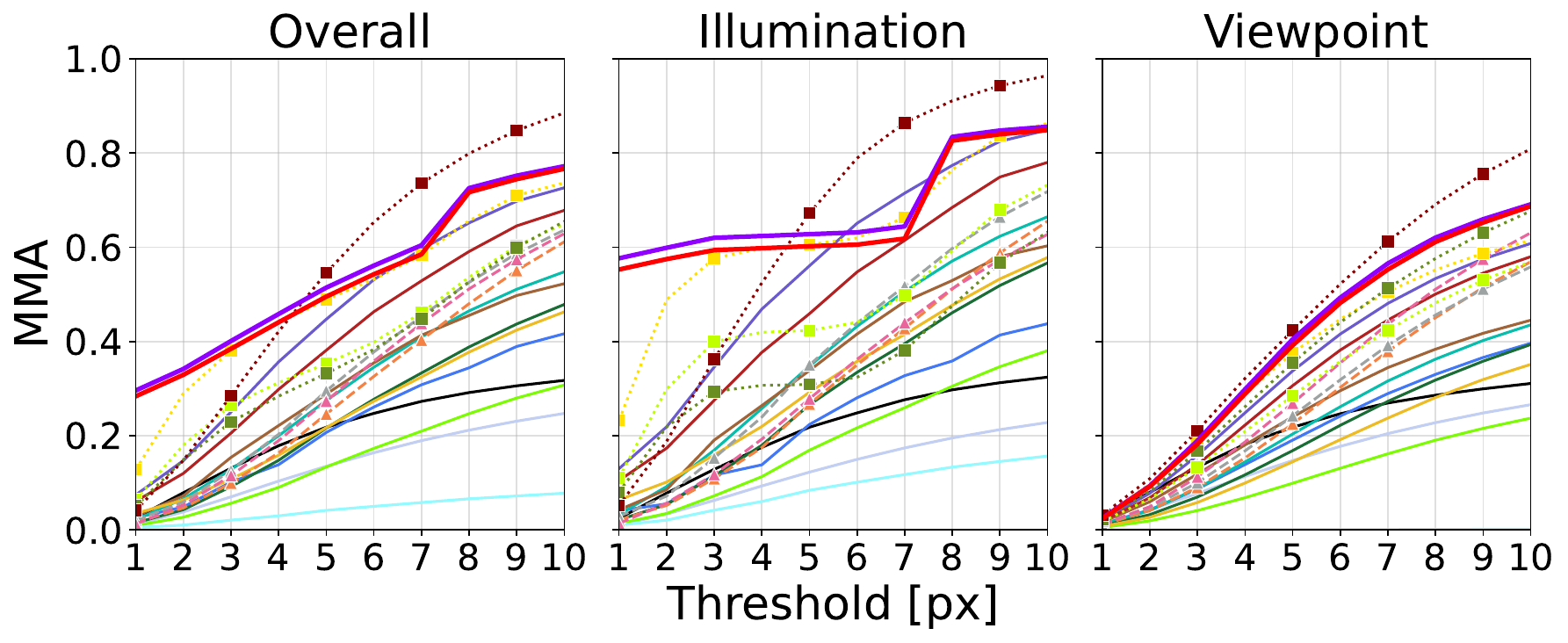}
    \end{minipage}
    \hfill
    \begin{minipage}{0.45\textwidth}
    \resizebox{8.2cm}{!}{
        \begin{tabular}{l c c c}
            \toprule
            & \multirow{2}{*}{\begin{tabular}[c]{@{}c@{}}Overall\\ MMA\end{tabular}}
            & \multirow{2}{*}{\begin{tabular}[c]{@{}c@{}}Illumination\\ MMA\end{tabular}}
            & \multirow{2}{*}{\begin{tabular}[c]{@{}c@{}}Viewpoint\\ MMA\end{tabular}} \\
            Method & & & \\
            \midrule
            \multicolumn{4}{l}{\textit{Detector-free (Dense/Semi-dense)}} \\
            \cstickdotted{clime}      LoFTR~\cite{sun2021loftr}                                                             & 26.40 & 39.97 & 13.29 \\
            \cstickdotted{cgold}      MatchFormer~\cite{wang2022matchformer}                                                & 38.08 & 57.69 & 19.14 \\
            \cstickdotted{colivedrab} ASpanFormer~\cite{chen2022aspanformer}                                                & 22.91 & 29.35 & 16.69 \\
            \cstickdotted{cdarkred}   RoMa v2~\cite{edstedt2025roma}                                                       & 28.51 & 36.24 & 21.04 \\
            \midrule
            \multicolumn{4}{l}{\textit{Sparse (detector + learned matcher)}} \\
            \cstickdashed{corange}    SuperPoint~\cite{DeTone2018SuperPointSI} + LightGlue~\cite{lindenberger2023lightglue} & 9.93  & 10.88 & 9.02  \\
            \cstickdashed{cgrey}      DISK~\cite{tyszkiewicz2020disk} + LightGlue~\cite{lindenberger2023lightglue}          & 12.58 & 15.24 & 10.00 \\
            \cstickdashed{crose}      ALIKED~\cite{Zhao2023ALIKED} + LightGlue~\cite{lindenberger2023lightglue}             & 11.51 & 11.77 & 11.27 \\
            \midrule
            \multicolumn{4}{l}{\textit{Sparse (detector + non-learned matching)}} \\
            \csticksolid{cblack}     SIFT~\cite{LoweDavid2004DistinctiveIF} + MNN                                         & 13.08 & 12.85 & 13.31 \\
            \csticksolid{cpurblue}   DoG + AffNet~\cite{AffNet2017} + HardNet~\cite{Mishchuk2017WorkingHT} + MNN          & 7.01  & 6.26  & 7.73  \\
            \csticksolid{cblue}      Key.Net~\cite{Laguna2019KeyNetKD} + AffNet~\cite{AffNet2017} + HardNet~\cite{Mishchuk2017WorkingHT} + MNN & 10.12 & 11.62 & 8.68  \\
            \csticksolid{cgrass}     SuperPoint~\cite{DeTone2018SuperPointSI} + MNN                                       & 5.63  & 7.21  & 4.09  \\
            \csticksolid{ccyan}      D2-Net~\cite{Dusmanu2019D2NetAT} + MNN                                               & 12.79 & 16.92 & 8.81  \\
            \csticksolid{cbrown}     R2D2~\cite{Revaud2019R2D2RA} + MNN                                                   & 15.39 & 19.05 & 11.84 \\
            \csticksolid{cgreen}     DISK~\cite{tyszkiewicz2020disk} + MNN                                                 & 9.18  & 11.47 & 6.97  \\
            \csticksolid{csky}       SiLK~\cite{gleize2023silk} + MNN                                                     & 2.06  & 4.17  & 0.02  \\
            \csticksolid{cyellow}    DeDoDe v2~\cite{edstedt2024dedodev2} + Dual-Softmax                                  & 10.76 & 15.90 & 5.80  \\
            \csticksolid{cslateblue}    BALF~\cite{zhao2024balf} + HardNet~\cite{Mishchuk2017WorkingHT} + MNN                & \cellcolor{tabthird}25.03 & \cellcolor{tabthird}34.58 & \cellcolor{tabthird}15.80 \\
            \csticksolid{cfirebrick}    BALF~\cite{zhao2024balf} + HyNet~\cite{hynet2020} + MNN                              & 20.55 & 27.51 & 13.82 \\
            \cmidrule{2-4}
            \csticksolid{cviolet}    SSMB (Ours) + HardNet~\cite{Mishchuk2017WorkingHT} + MNN & \cellcolor{tabfirst}40.10 & \cellcolor{tabfirst}62.02 & \cellcolor{tabfirst}18.91 \\
            \csticksolid{cred}       SSMB (Ours) + HyNet~\cite{hynet2020} + MNN               & \cellcolor{tabsecond}38.45 & \cellcolor{tabsecond}59.93 & \cellcolor{tabsecond}18.21 \\
            \bottomrule
        \end{tabular}
    }
    \end{minipage}
    \caption{\textbf{Image matching on the Blur-HPatches dataset~\cite{zhao2024balf}.}
    \textit{Left:} Mean Matching Accuracy (MMA) curves at different thresholds.
    Solid lines denote sparse detector + non-learned matching methods, dashed lines with triangle markers ($\blacktriangle$) denote sparse detector + learned matcher methods, and dotted lines with square markers ($\blacksquare$) denote detector-free methods.
    SSMB (thick solid lines) is highlighted for clarity.
    \textit{Right:} MMA at a 3-pixel threshold.
    \colorbox{tabfirst}{first}, \colorbox{tabsecond}{second}, and \colorbox{tabthird}{third} best results are highlighted within sparse methods only.
    SSMB with HardNet achieves the highest MMA among all sparse methods, and at this strict threshold also surpasses every detector-free method.}
    \label{fig:image_matching_appendix}
\end{figure*}

\cref{fig:image_matching_appendix} further validates the effectiveness of SSMB for image matching under motion blurred conditions.
The plot on the left shows the Mean Matching Accuracy (MMA) curves for different methods at various matching thresholds, providing a comprehensive view of each method's performance.
SSMB with HardNet consistently outperforms all sparse methods, excelling in overall, illumination, and viewpoint MMA.
At the strict 3-pixel threshold, SSMB with HardNet even surpasses the strongest detector-free methods, RoMa v2 and MatchFormer, despite operating on a single sparse keypoint set rather than full dense correspondences.

\subsection{Relative Pose Estimation}\label{subsec:relpose_appendix}

\begin{figure*}[!hb]
    \centering
    {\scriptsize
    \begin{minipage}[t]{0.48\linewidth}
        \centering
        \includegraphics[width=\linewidth]{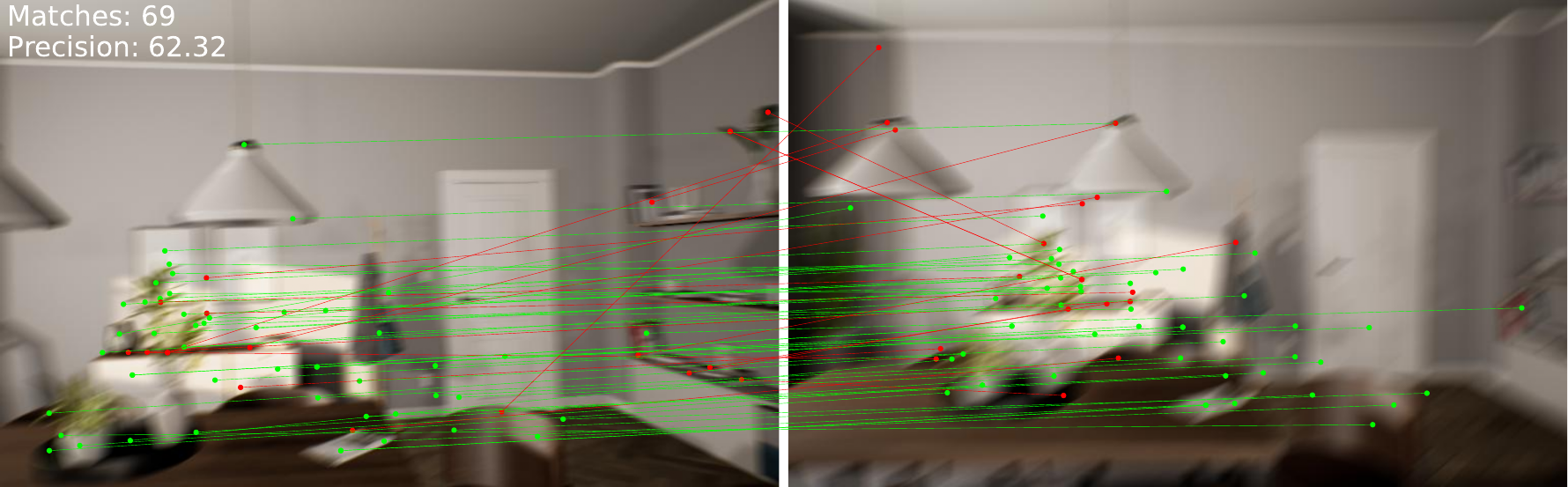}\\[1mm]
        SIFT~\cite{LoweDavid2004DistinctiveIF} + MNN
    \end{minipage}
    \hspace{0.5mm}
    \begin{minipage}[t]{0.48\linewidth}
        \centering
        \includegraphics[width=\linewidth]{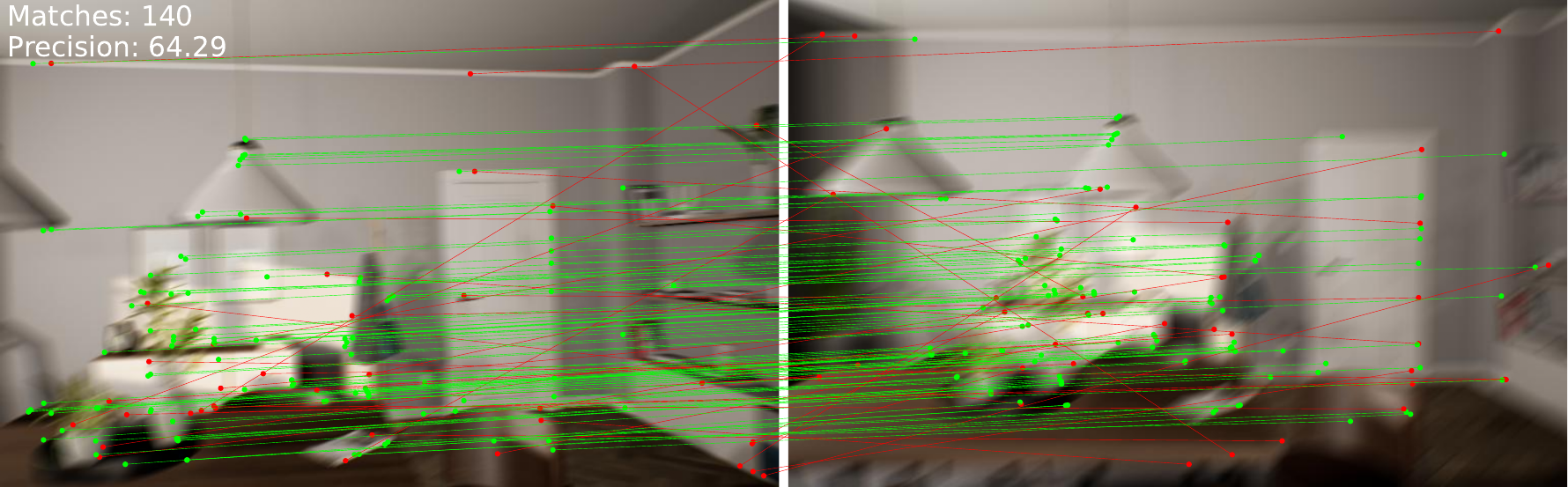}\\[1mm]
        Key.Net~\cite{Laguna2019KeyNetKD} + AffNet~\cite{AffNet2017} + HardNet~\cite{Mishchuk2017WorkingHT} + MNN
    \end{minipage}
    
    \vspace{4mm}
    
    \begin{minipage}[t]{0.48\linewidth}
        \centering
        \includegraphics[width=\linewidth]{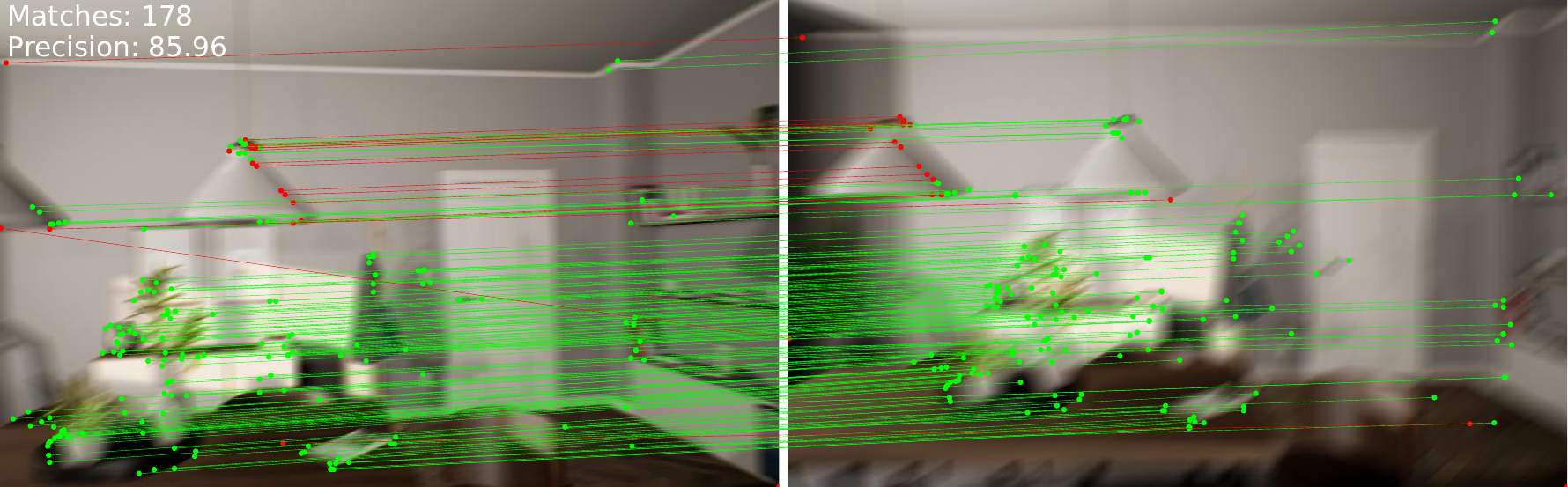}\\[1mm]
        DeDoDe v2~\cite{edstedt2024dedodev2} + Dual-Softmax
    \end{minipage}
    \hspace{0.5mm}
    \begin{minipage}[t]{0.48\linewidth}
        \centering
        \includegraphics[width=\linewidth]{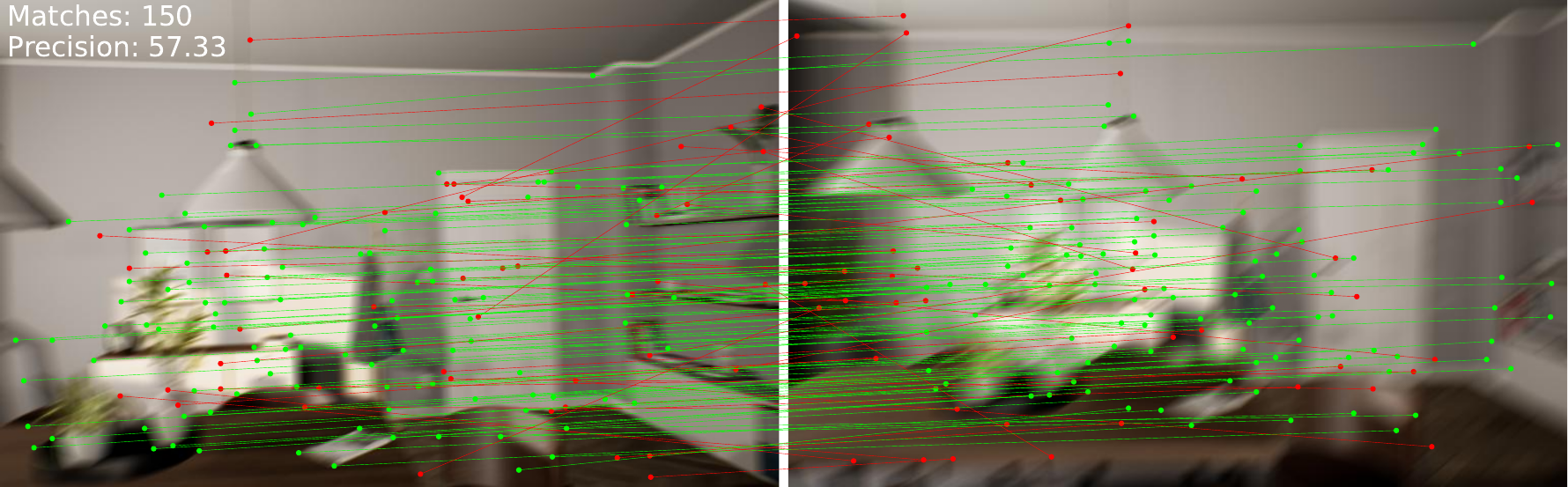}\\[1mm]
        SuperPoint~\cite{DeTone2018SuperPointSI} + LightGlue~\cite{lindenberger2023lightglue}
    \end{minipage}
    
    \vspace{4mm}
    
    \begin{minipage}[t]{0.48\linewidth}
        \centering
        \includegraphics[width=\linewidth]{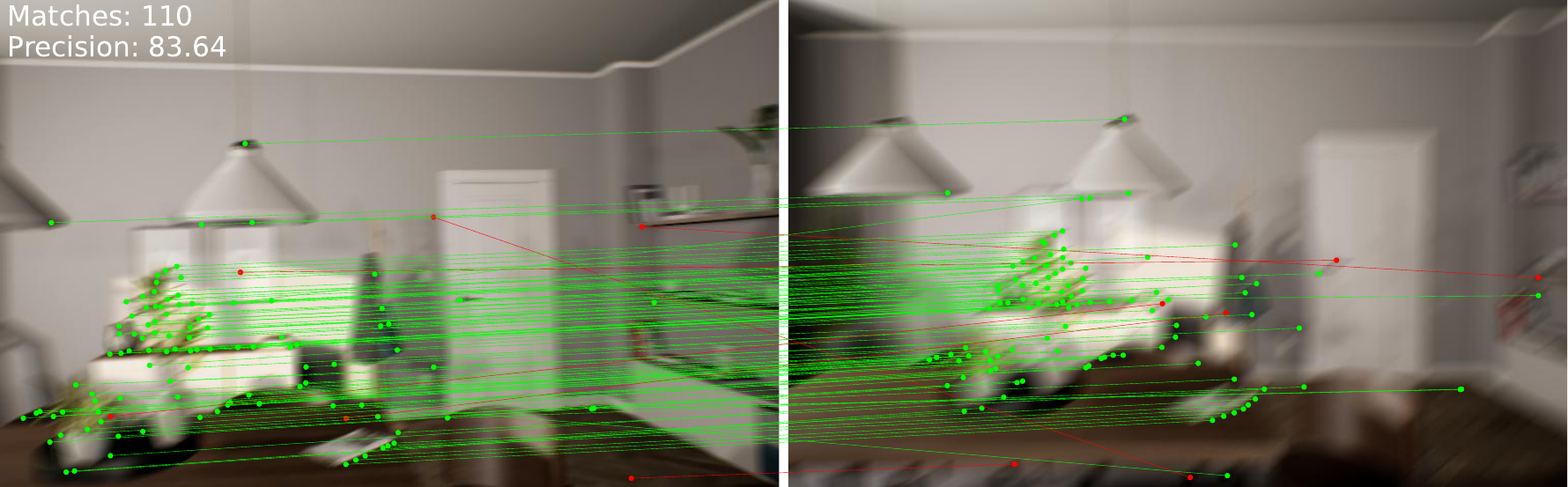}\\[1mm]
        ALIKED~\cite{Zhao2023ALIKED} + LightGlue~\cite{lindenberger2023lightglue}
    \end{minipage}
    \hspace{0.5mm}
    \begin{minipage}[t]{0.48\linewidth}
        \centering
        \includegraphics[width=\linewidth]{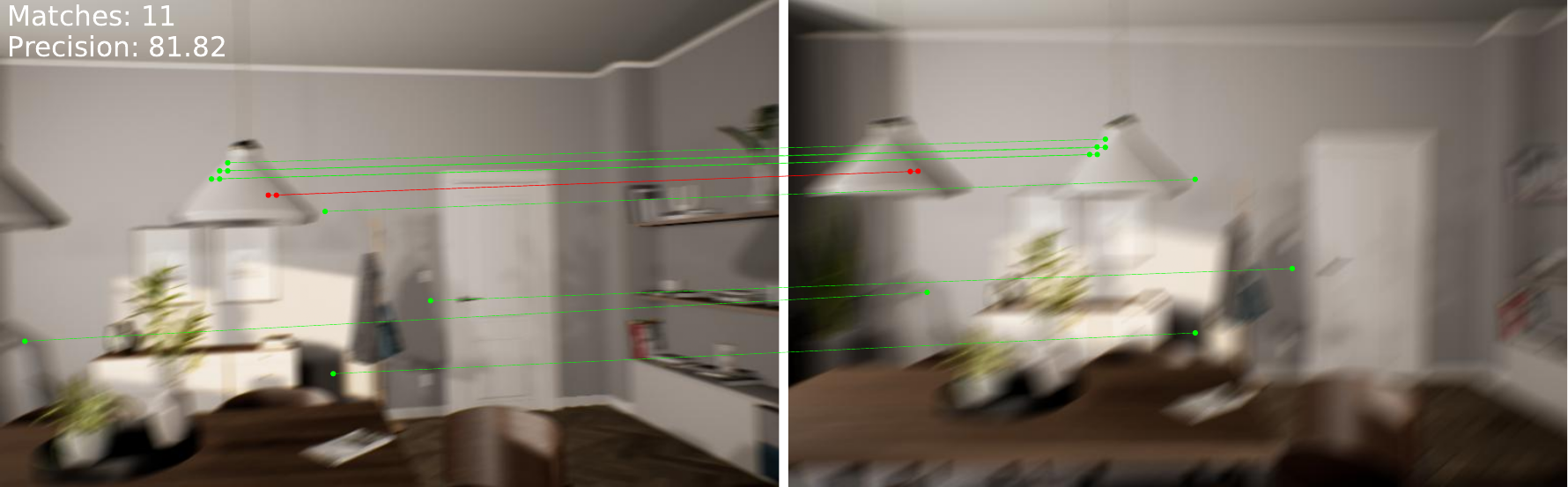}\\[1mm]
        LoFTR~\cite{sun2021loftr}
    \end{minipage}
    
    \vspace{4mm}
    
    \begin{minipage}[t]{0.48\linewidth}
        \centering
        \includegraphics[width=\linewidth]{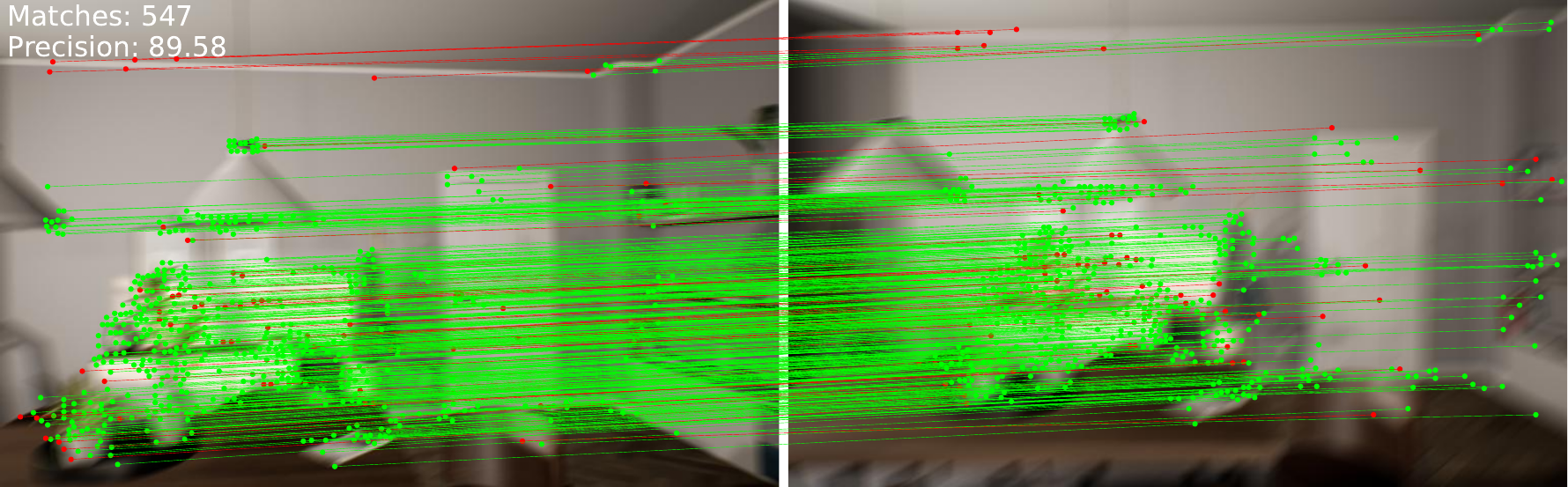}\\[1mm]
        BALF~\cite{zhao2024balf} + HardNet~\cite{Mishchuk2017WorkingHT} + MNN
    \end{minipage}
    \hspace{0.5mm}
    \begin{minipage}[t]{0.48\linewidth}
        \centering
        \includegraphics[width=\linewidth]{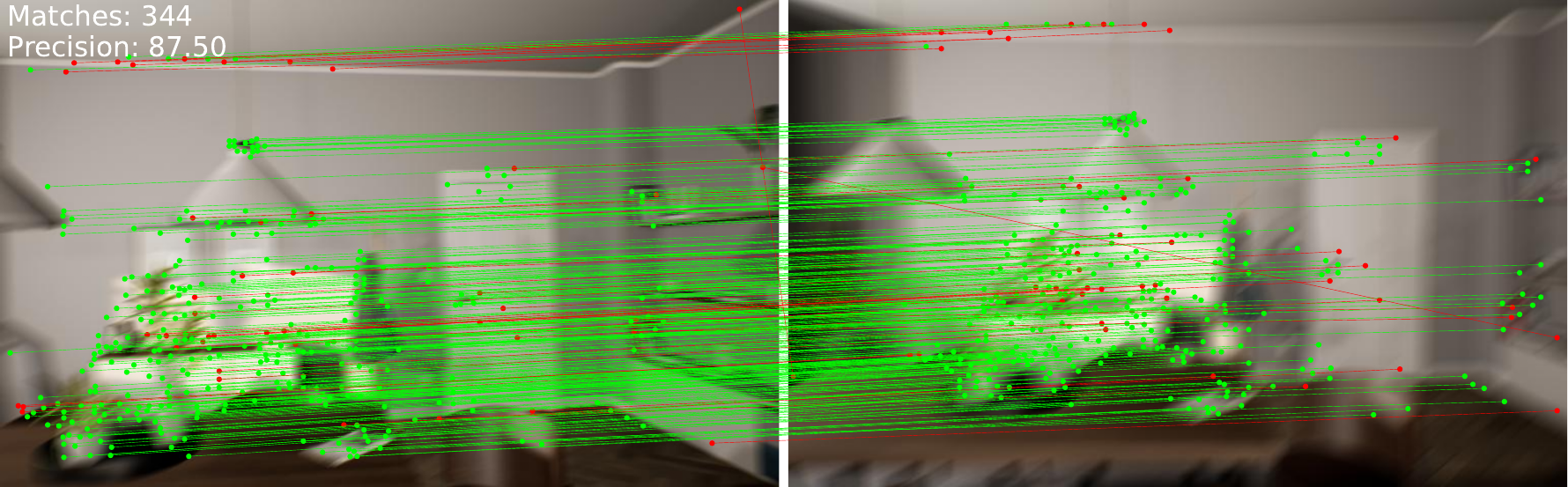}\\[1mm]
        BALF~\cite{zhao2024balf} + HyNet~\cite{hynet2020} + MNN
    \end{minipage}
    
    \vspace{4mm}
    
    \begin{minipage}[t]{0.48\linewidth}
        \centering
        \includegraphics[width=\linewidth]{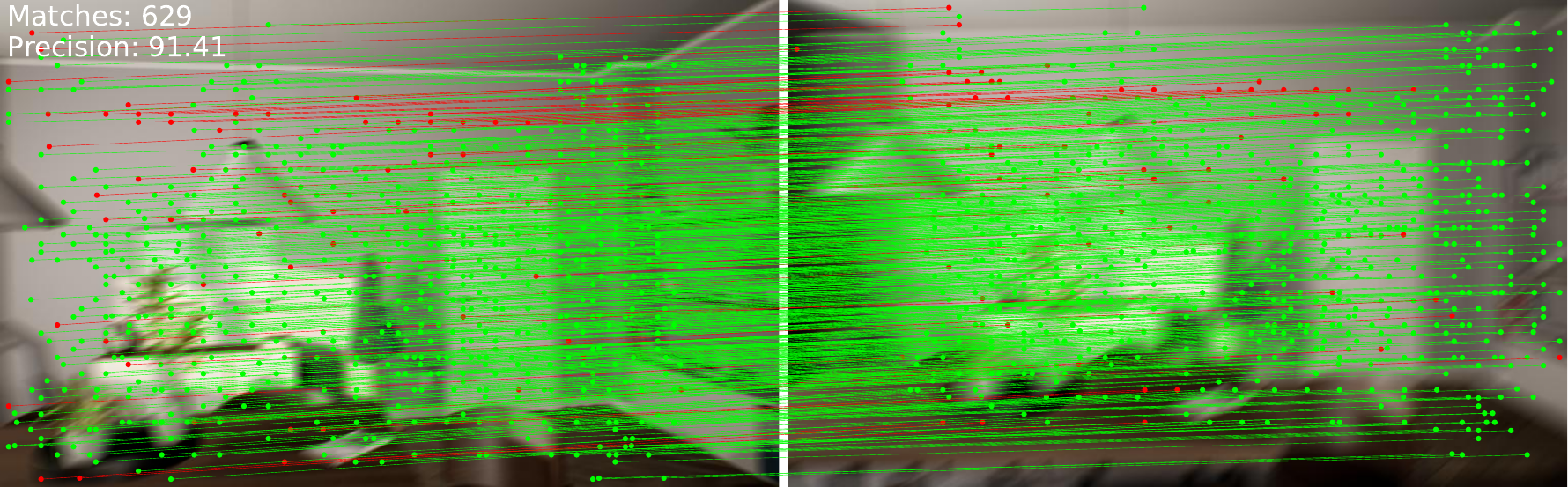}\\[1mm]
        \textbf{SSMB (Ours)} + HardNet~\cite{Mishchuk2017WorkingHT} + MNN
    \end{minipage}
    \hspace{0.5mm}
    \begin{minipage}[t]{0.48\linewidth}
        \centering
        \includegraphics[width=\linewidth]{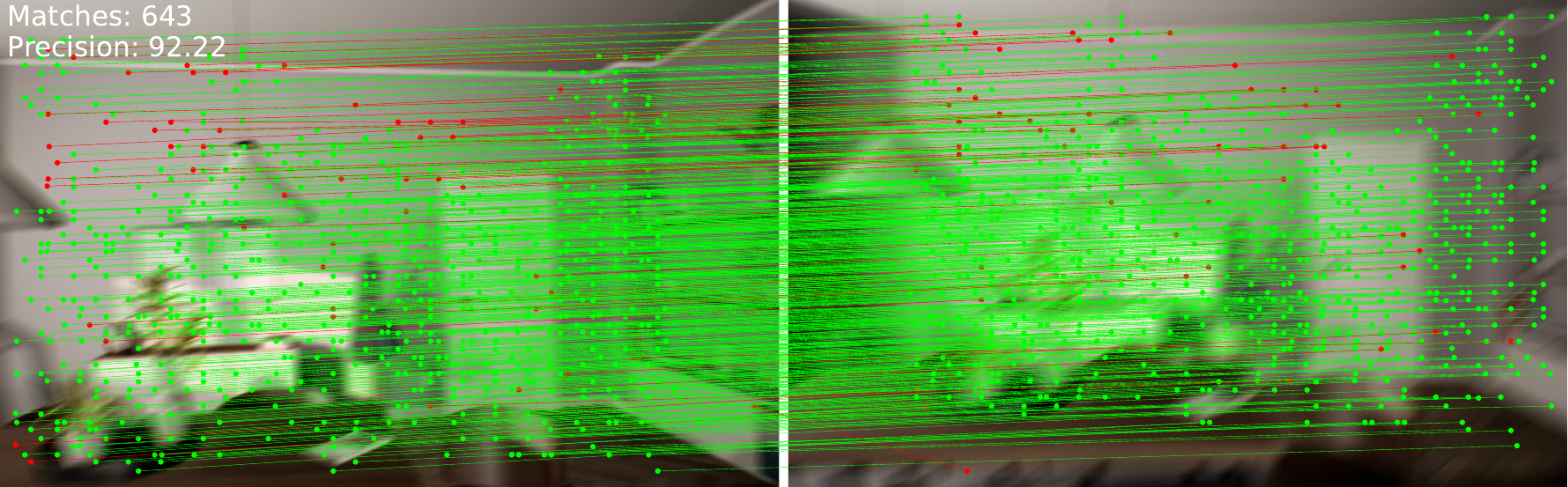}\\[1mm]
        \textbf{SSMB (Ours)} + HyNet~\cite{hynet2020} + MNN
    \end{minipage}
    }
    \caption{\textbf{Qualitative matching results on the ArchViz dataset~\cite{liu2021mba} under the blur-to-blur setting.}
    Each subfigure displays the number of correct matches and precision in the top-left corner.
    SSMB, when paired with HardNet~\cite{Mishchuk2017WorkingHT} or HyNet~\cite{hynet2020}, achieves more correct matches and fewer mismatches, effectively handling motion blur and viewpoint changes.
    Green lines indicate correct matches (epipolar error below $5 \times 10^{-4}$) and red lines indicate incorrect matches.
    Best viewed in color.}
\label{fig:qualitative_relpose}
\end{figure*}

This section presents additional qualitative results for relative pose estimation on the ArchViz dataset~\cite{liu2021mba}, following the \emph{blur-to-blur} setting discussed in the main paper.
The figure compares qualitative matching results across multiple methods, highlighting correct matches and mismatches under motion blur and viewpoint changes.
Correct matches are plotted in green, and incorrect matches (\textit{i.e.,} epipolar errors beyond $5 \times 10^{-4}$) in red.
 
\cref{fig:qualitative_relpose} illustrates matching results for different methods on the same image pair.
These results show that SSMB, when paired with HardNet~\cite{Mishchuk2017WorkingHT} or HyNet~\cite{hynet2020}, achieves more correct matches and fewer mismatches compared to existing methods, demonstrating superior robustness to motion blur and viewpoint changes.
These reinforce the conclusions in the main paper, confirming SSMB's effectiveness for relative pose estimation under challenging blur conditions.

\subsection{More Qualitative Results}\label{subsec:more_real_results}

\begin{figure*}[t]
    \centering
    {\scriptsize
    \begin{minipage}[t]{0.235\linewidth}
        \centering
        \includegraphics[width=\linewidth]{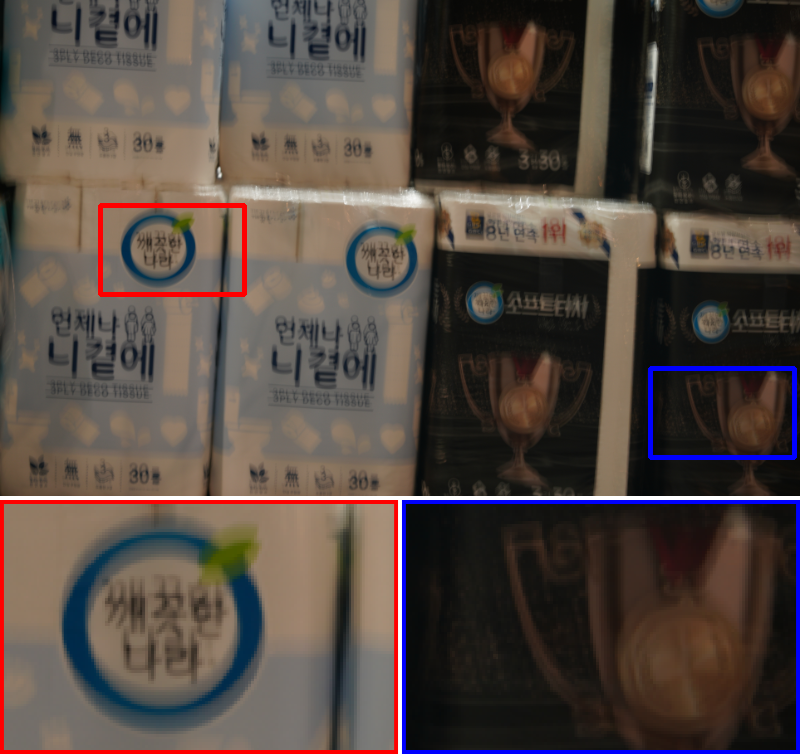}\\[1mm]
        Input
    \end{minipage}
    \hspace{0.5mm}
    \begin{minipage}[t]{0.235\linewidth}
        \centering
        \includegraphics[width=\linewidth]{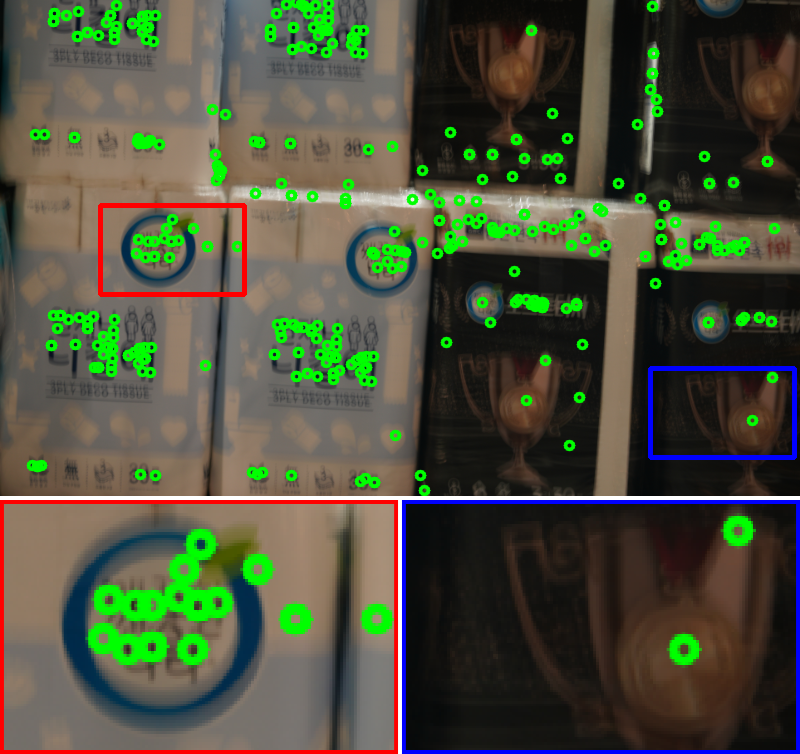}\\[1mm]
        SIFT~\cite{LoweDavid2004DistinctiveIF}
    \end{minipage}
    \hspace{0.5mm}
    \begin{minipage}[t]{0.235\linewidth}
        \centering
        \includegraphics[width=\linewidth]{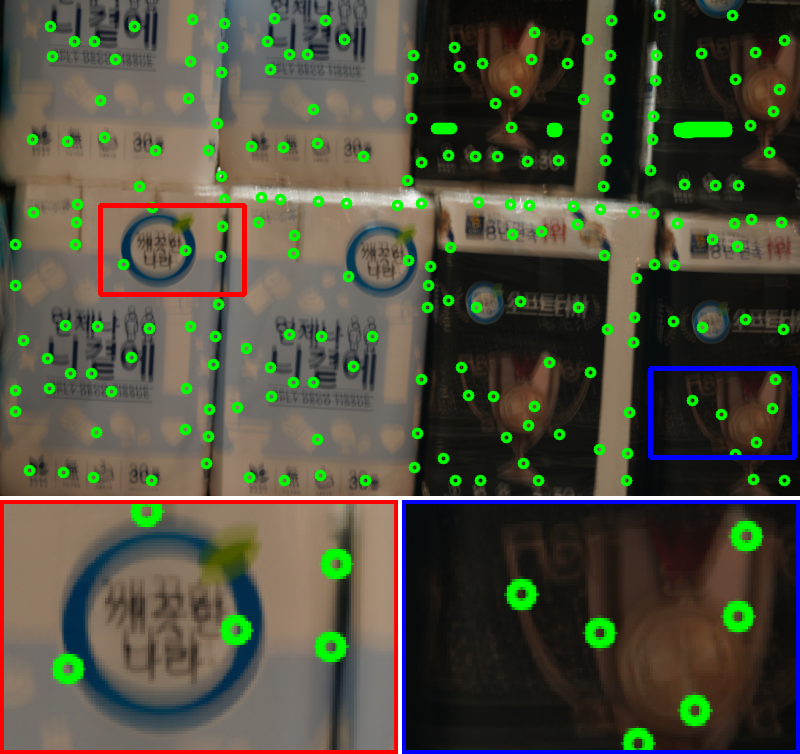}\\[1mm]
        Key.Net~\cite{Laguna2019KeyNetKD}
    \end{minipage}
    \hspace{0.5mm}
    \begin{minipage}[t]{0.235\linewidth}
        \centering
        \includegraphics[width=\linewidth]{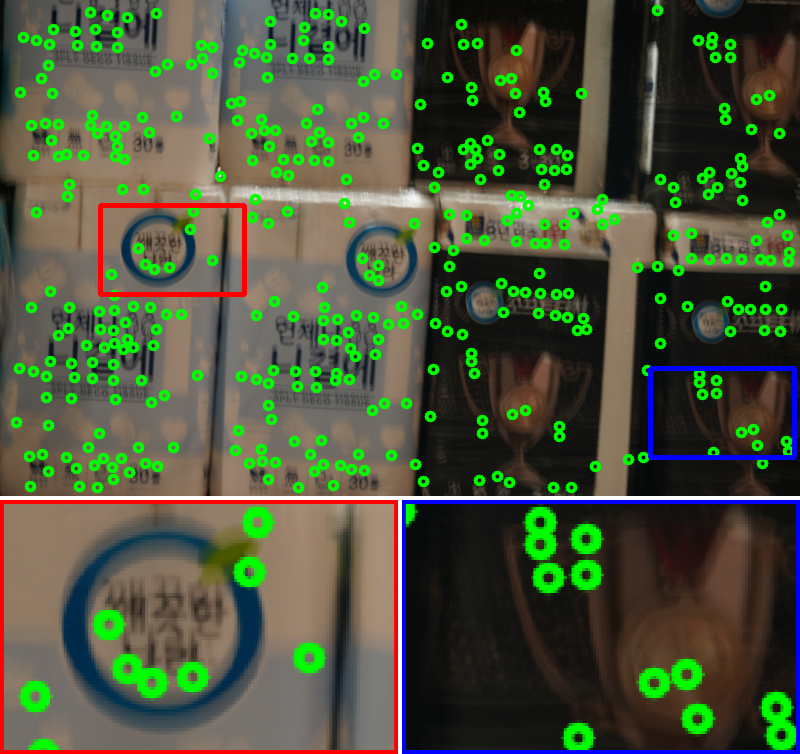}\\[1mm]
        SuperPoint~\cite{DeTone2018SuperPointSI}
    \end{minipage}
    
    \vspace{4mm}
    
    \begin{minipage}[t]{0.235\linewidth}
        \centering
        \includegraphics[width=\linewidth]{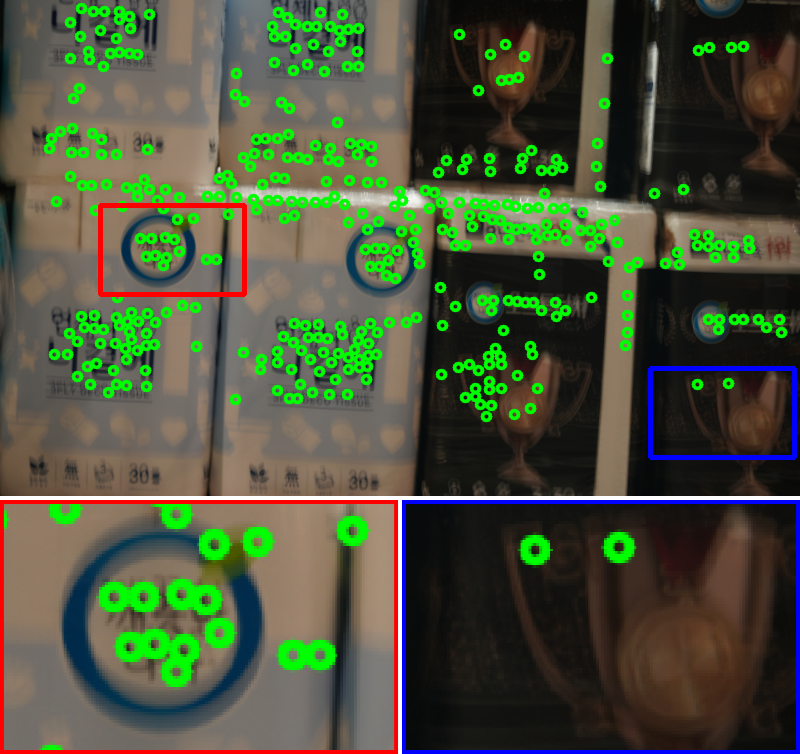}\\[1mm]
        DISK~\cite{tyszkiewicz2020disk}
    \end{minipage}
    \hspace{0.5mm}
    \begin{minipage}[t]{0.235\linewidth}
        \centering
        \includegraphics[width=\linewidth]{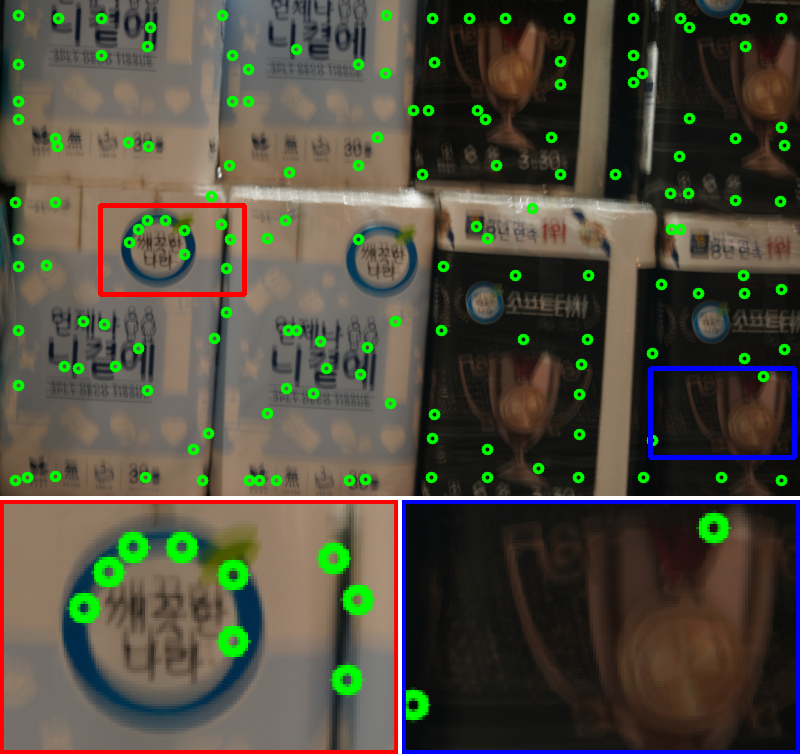}\\[1mm]
        REKD~\cite{lee2022self}
    \end{minipage}
    \hspace{0.5mm}
    \begin{minipage}[t]{0.235\linewidth}
        \centering
        \includegraphics[width=\linewidth]{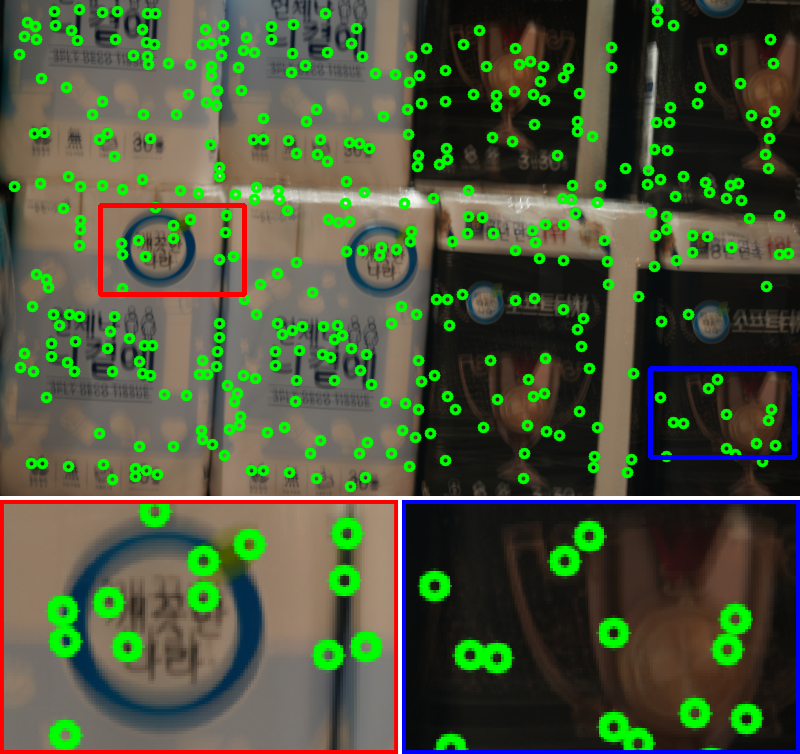}\\[1mm]
        NeSS-ST~\cite{pakulev2023ness}
    \end{minipage}
    \hspace{0.5mm}
    \begin{minipage}[t]{0.235\linewidth}
        \centering
        \includegraphics[width=\linewidth]{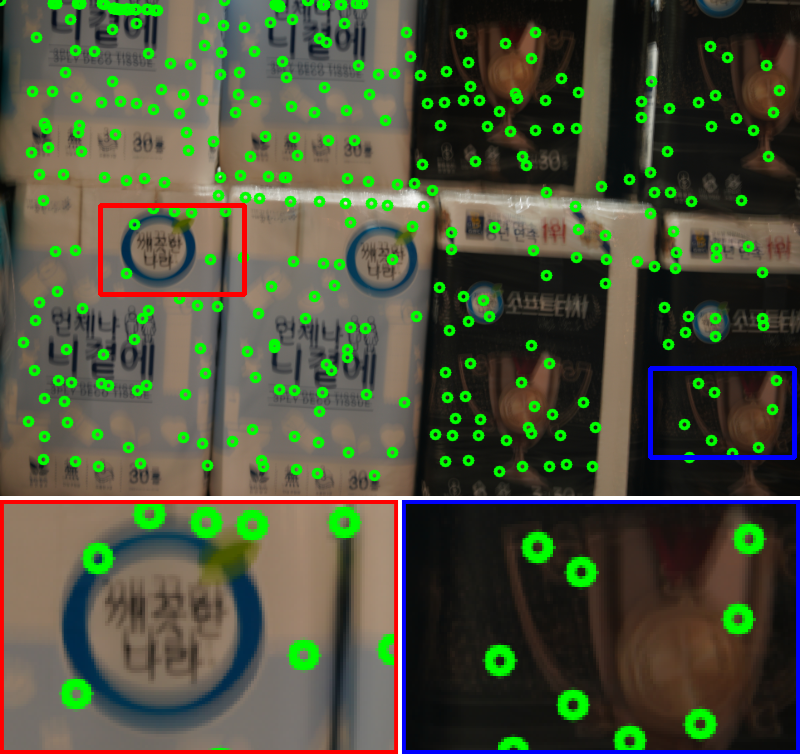}\\[1mm]
        ALIKED~\cite{Zhao2023ALIKED}
    \end{minipage}
    
    \vspace{4mm}
    
    \begin{minipage}[t]{0.235\linewidth}
        \centering
        \includegraphics[width=\linewidth]{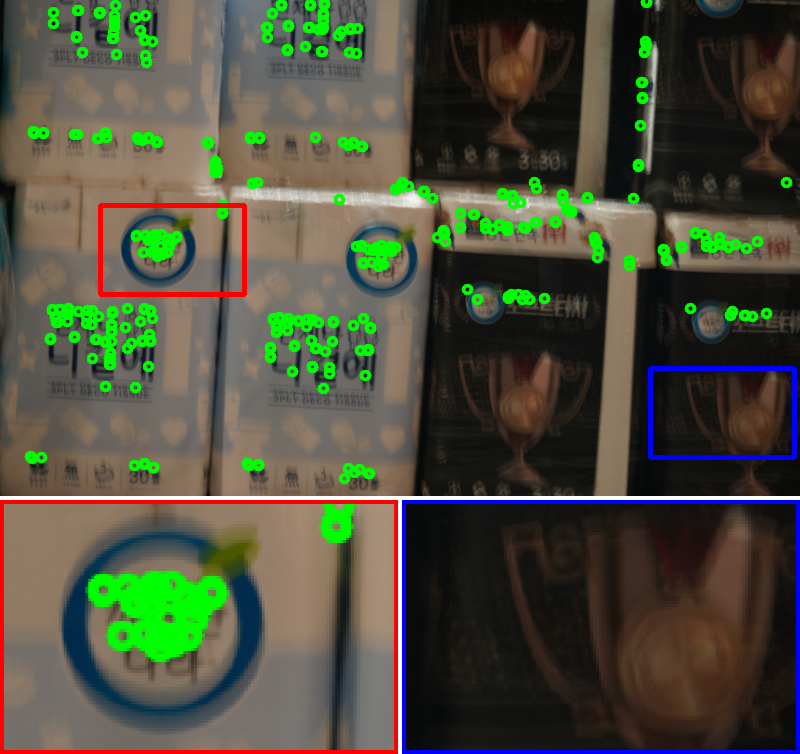}\\[1mm]
        DeDoDe v2~\cite{edstedt2024dedodev2}
    \end{minipage}
    \hspace{0.5mm}
    \begin{minipage}[t]{0.235\linewidth}
        \centering
        \includegraphics[width=\linewidth]{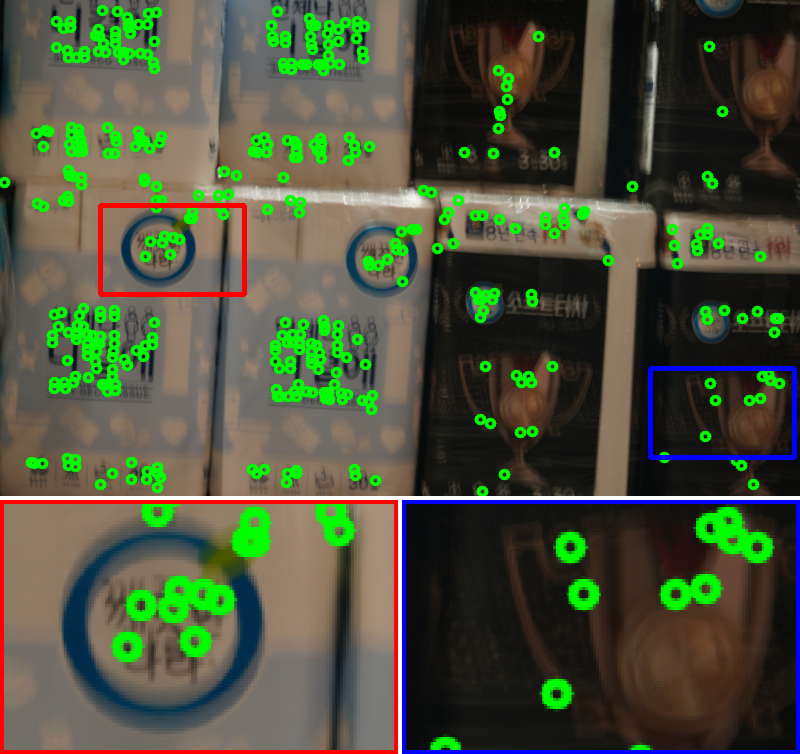}\\[1mm]
        XFeat~\cite{potje2024cvpr}
    \end{minipage}
    \hspace{0.5mm}
    \begin{minipage}[t]{0.235\linewidth}
        \centering
        \includegraphics[width=\linewidth]{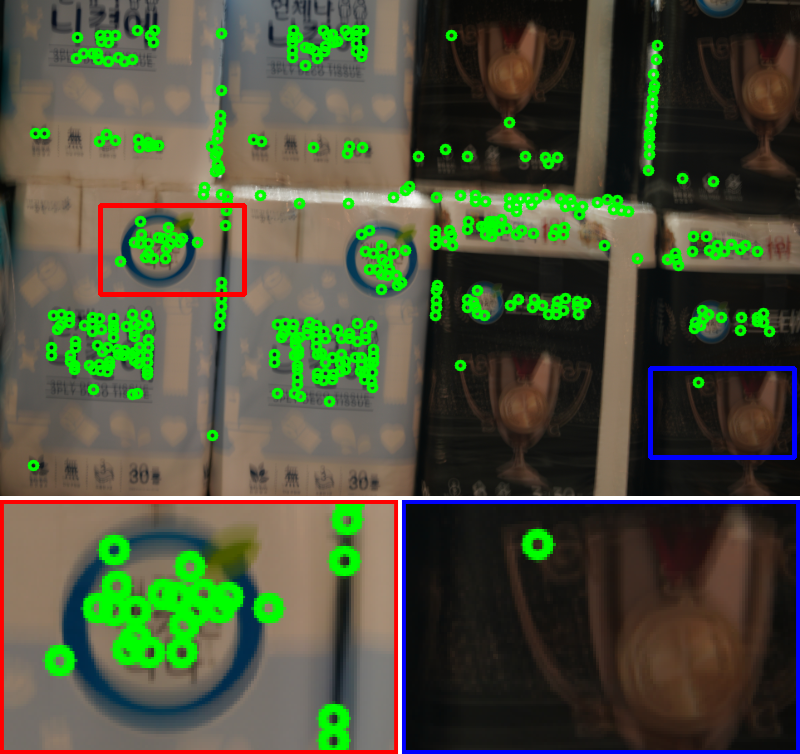}\\[1mm]
        BALF~\cite{zhao2024balf}
    \end{minipage}
    \hspace{0.5mm}
    \begin{minipage}[t]{0.235\linewidth}
        \centering
        \includegraphics[width=\linewidth]{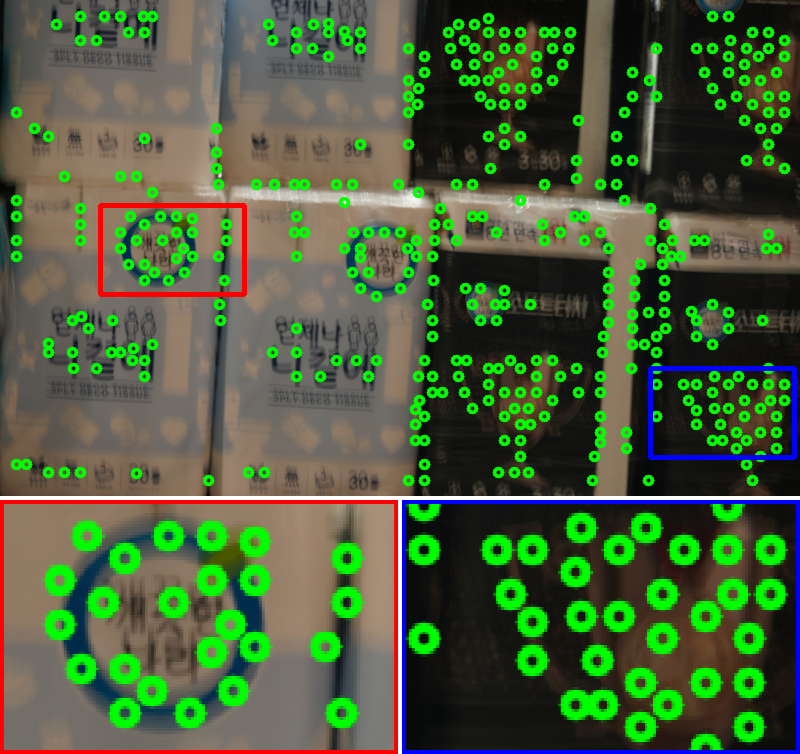}\\[1mm]
        \textbf{SSMB (Ours)}
    \end{minipage}
    }
    \caption{\textbf{Qualitative detection results on real-world blurred images from the RealBlur dataset~\cite{Rim2020RealWorldBD}.}
    SSMB detects well-localized and consistently distributed keypoints under real camera motion blur, while several baselines either miss salient structure or produce noisy responses in heavily blurred regions.
    The \textcolor{red}{red} and \textcolor{blue}{blue} boxes mark two representative regions, shown enlarged below each image for closer comparison.
    Best viewed in color.}
\label{fig:real_detection_appendix}
\end{figure*}

\PAR{Detection on RealBlur.}
\cref{fig:real_detection_appendix} presents additional keypoint detection results on a real-world blurred image from the RealBlur dataset~\cite{Rim2020RealWorldBD}, complementing the RWBI results in \cref{fig:real_detection} of the main paper with a broader set of baseline comparisons.
Eleven methods are shown, including four methods not in the main figure: Key.Net, REKD, DeDoDe v2, and XFeat.
SSMB consistently produces well-localized keypoints concentrated on salient image structure, while most baselines either respond sparsely in blurred regions or distribute detections without aligning to meaningful structure.
The enlarged regions in \cref{fig:real_detection_appendix} further highlight this advantage, where SSMB maintains dense and well-aligned detections in areas that most baselines fail to cover.
The results confirm that SSMB's blur-robust behavior generalizes across different real-world blur sources and scene types.

\begin{figure*}[t]
    \centering
    \renewcommand{\arraystretch}{0}
    \setlength{\tabcolsep}{0pt}
    \begin{tabular}{c @{\hspace{3pt}} !{\vrule width 0.8pt} @{\hspace{4pt}} l}
        \multirow{2}{*}{
            \rotatebox[origin=c]{90}{\normalsize\textit{Indoor}}
        } &
        \includegraphics[width=0.96\linewidth]{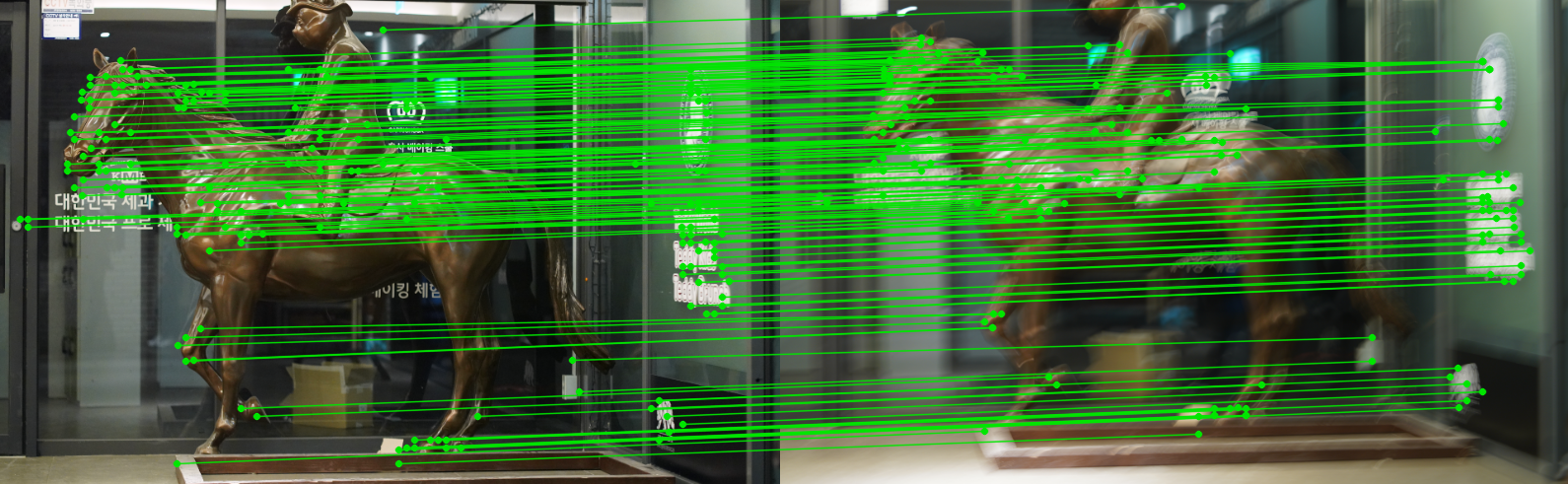} \\[1mm]
        &
        \includegraphics[width=0.96\linewidth]{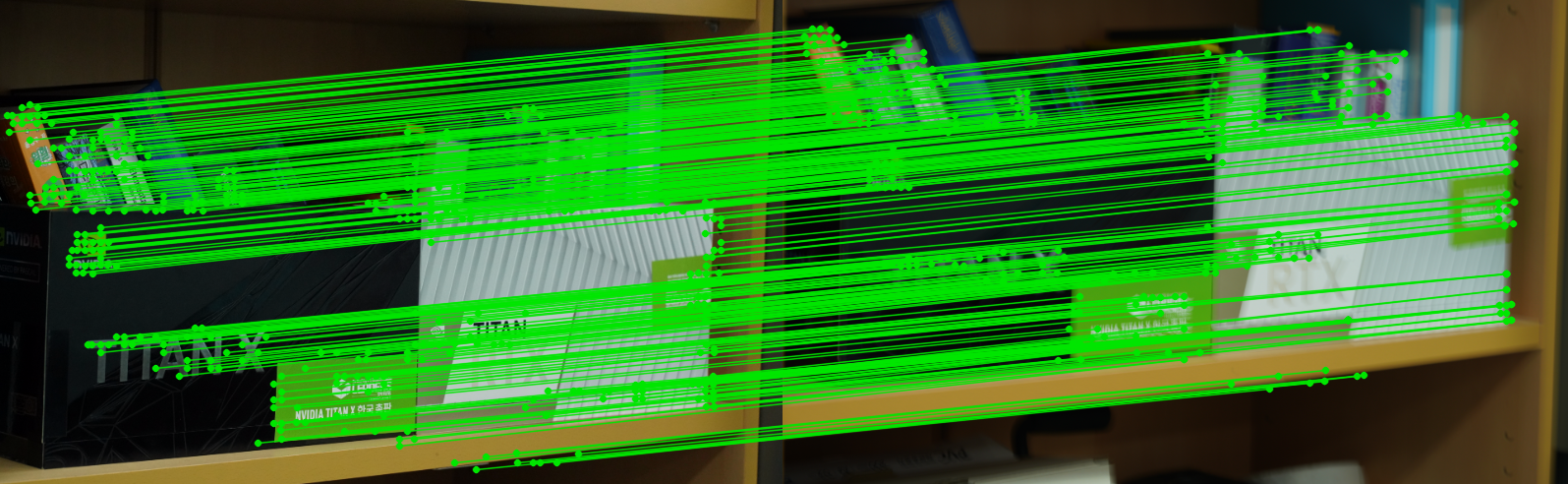} \\
        \noalign{\vskip 4mm}
        \multirow{2}{*}{
            \rotatebox[origin=c]{90}{\normalsize\textit{Outdoor}}
        } &
        \includegraphics[width=0.96\linewidth]{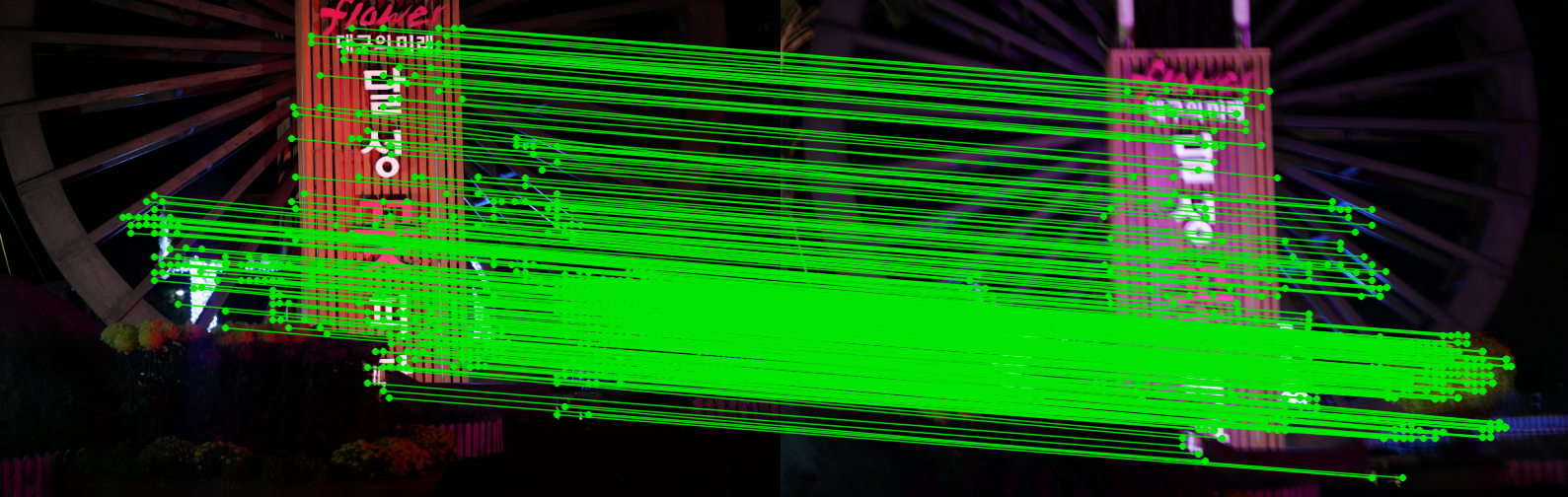} \\[1mm]
        &
        \includegraphics[width=0.96\linewidth]{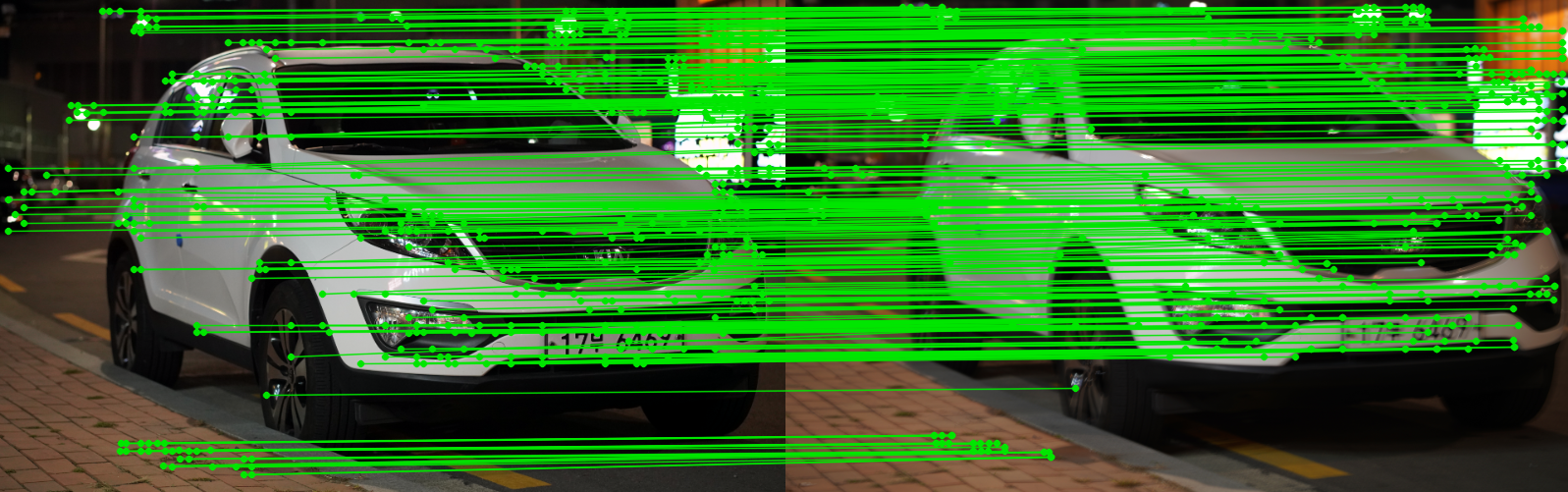} \\
    \end{tabular}
    \caption{\textbf{Qualitative matching results on real-world sharp-blur image pairs from the RealBlur dataset~\cite{Rim2020RealWorldBD}.}
    Each row shows a matched sharp-blur pair (left: sharp; right: blurred) with real camera motion blur and viewpoint changes.
    SSMB detects well-localized and repeatable keypoints across diverse indoor and outdoor scenes despite the cross-domain appearance gap.
    Green lines indicate correct matches.
    Best viewed in color.}
    \label{fig:real_match_appendix}
\end{figure*}
 
\PAR{Matching on RealBlur.}
\cref{fig:real_match_appendix} presents additional qualitative feature matching results on real-world sharp-blur image pairs from the RealBlur dataset~\cite{Rim2020RealWorldBD}, complementing the result in \cref{fig:real_match} of the main paper with coverage of two indoor and two outdoor scenes.
Each pair contains real camera motion blur and viewpoint changes between the sharp and blurred images, following the same setting as the main paper.
SSMB is combined with HardNet~\cite{Mishchuk2017WorkingHT} and MNN matching.
Across all four scenes, SSMB consistently detects well-localized and repeatable keypoints from both images despite the cross-domain appearance gap, yielding dense and geometrically consistent matches under challenging real-world blur conditions.

\bibliographystyle{IEEEtran}
\bibliography{egbib}

@String(CVPR = {Proc. IEEE/CVF Conf. Comput. Vis. Pattern Recog.})

@String(ICCV = {Proc. IEEE/CVF Int. Conf. Comput. Vis.})

@String(ECCV = {Proc. Eur. Conf. Comput. Vis.})

@String(BMVC = {Proc. Brit. Mach. Vis. Conf.})

@String(WACV = {Proc. IEEE/CVF Winter Conf. Appl. Comput. Vis.})

@String(ACCV = {Proc. Asian Conf. Comput. Vis.})

@String(CVPRW = {Proc. IEEE/CVF Conf. Comput. Vis. Pattern Recog. Worksh.})

@String(ICPR = {Proc. Int. Conf. Pattern Recog.})

@String(NIPS = {Proc. Adv. Neural Inf. Process. Syst.})

@String(PAMI = {IEEE Trans. Pattern Anal. Mach. Intell.})

@String(IJCV = {Int. J. Comput. Vis.})

@String(TRO = {IEEE Trans. Robot.})

@String(TIM = {IEEE Trans. Instrum. Meas.})

@inproceedings{Sun2015LearningAC,
	title={Learning a convolutional neural network for non-uniform motion blur removal},
	author={Jian Sun and Wenfei Cao and Zongben Xu and Jean Ponce},
	booktitle=CVPR,
	year={2015},
	pages={769--777}
}

@inproceedings{Nah2017DeepMC,
	title={Deep Multi-scale Convolutional Neural Network for Dynamic Scene Deblurring},
	author={Seungjun Nah and Tae Hyun Kim and Kyoung Mu Lee},
	booktitle=CVPR,
	year={2017},
	pages={257--265}
}

@inproceedings{Tao2018ScaleRecurrentNF,
	title={Scale-Recurrent Network for Deep Image Deblurring},
	author={Xin Tao and Hongyun Gao and Yi Wang and Xiaoyong Shen and Jue Wang and Jiaya Jia},
	booktitle=CVPR,
	year={2018},
	pages={8174--8182}
}

@inproceedings{Kupyn2019DeblurGANv2D,
	title={DeblurGAN-v2: Deblurring (Orders-of-Magnitude) Faster and Better},
	author={Orest Kupyn and T. Martyniuk and Junru Wu and Zhangyang Wang},
	booktitle=ICCV,
	year={2019},
	pages={8877--8886}
}

@inproceedings{Kupyn2018DeblurGANBM,
	title={DeblurGAN: Blind Motion Deblurring Using Conditional Adversarial Networks},
	author={Orest Kupyn and Volodymyr Budzan and Mykola Mykhailych and Dmytro Mishkin and Jiri Matas},
	booktitle=CVPR,
	year={2018},
	pages={8183--8192}
}

@inproceedings{Harris1988ACC,
	title={A Combined Corner and Edge Detector},
	author={Christopher G. Harris and M. J. Stephens},
	booktitle={Alvey Vision Conference},
	year={1988},
	pages = {147--151}
}

@article{LoweDavid2004DistinctiveIF,
  title={Distinctive image features from scale-invariant keypoints},
  author={Lowe, David G},
  journal=IJCV,
  volume={60},
  number={2},
  pages={91--110},
  year={2004}
}

@article{Mikolajczyk2003APE,
  title={A performance evaluation of local descriptors},
  author={Mikolajczyk, Krystian and Schmid, Cordelia},
  journal=PAMI,
  volume={27},
  number={10},
  pages={1615--1630},
  year={2005},
}

@inproceedings{Laguna2019KeyNetKD,
	title={Key.Net: Keypoint Detection by Handcrafted and Learned CNN Filters},
	author={Axel Barroso Laguna and Edgar Riba and Daniel Ponsa and Krystian Mikolajczyk},
	booktitle=ICCV,
	year={2019},
	pages={5835--5843}
}

@inproceedings{zhang2017learning,
  title={Learning discriminative and transformation covariant local feature detectors},
  author={Zhang, Xu and Yu, Felix X and Karaman, Svebor and Chang, Shih-Fu},
  booktitle=CVPR,
  pages={6818--6826},
  year={2017}
}

@inproceedings{DeTone2018SuperPointSI,
	title={SuperPoint: Self-Supervised Interest Point Detection and Description},
	author={Daniel DeTone and Tomasz Malisiewicz and Andrew Rabinovich},
	booktitle=CVPRW,
	year={2018},
	pages={224--236}
}

@inproceedings{Ono2018LFNetLL,
	title={LF-Net: Learning Local Features from Images},
	author={Yuki Ono and Eduard Trulls and Pascal V. Fua and Kwang Moo Yi},
	booktitle=NIPS,
    volume={31},
	year={2018}
}

@inproceedings{Tu2022MAXIMMM,
	title={MAXIM: Multi-Axis MLP for Image Processing},
	author={Zhengzhong Tu and Hossein Talebi and Han Zhang and Feng Yang and Peyman Milanfar and Alan Conrad Bovik and Yinxiao Li},
	booktitle=CVPR,
	year={2022},
	pages={5759--5770}
}

@inproceedings{Balntas2017HPatchesAB,
	title={HPatches: A Benchmark and Evaluation of Handcrafted and Learned Local Descriptors},
	author={Vassileios Balntas and Karel Lenc and Andrea Vedaldi and Krystian Mikolajczyk},
	booktitle=CVPR,
	year={2017},
	pages={3852--3861}
}

@inproceedings{Dusmanu2019D2NetAT,
	title={D2-Net: A Trainable CNN for Joint Description and Detection of Local Features},
	author={Mihai Dusmanu and Ignacio Rocco and Tom{\'a}s Pajdla and Marc Pollefeys and Josef Sivic and Akihiko Torii and Torsten Sattler},
	booktitle=CVPR,
	year={2019},
	pages={8084--8093}
}

@inproceedings{Revaud2019R2D2RA,
	title={R2D2: Repeatable and Reliable Detector and Descriptor},
	author={J{\'e}r{\^o}me Revaud and Philippe Weinzaepfel and C{\'e}sar Roberto de Souza and No'e Pion and Gabriela Csurka and Yohann Cabon and M. Humenberger},
	booktitle=NIPS,
    volume={32},
	year={2019}
}

@inproceedings{zhang2020deblurring,
	title={Deblurring by realistic blurring},
	author={Zhang, Kaihao and Luo, Wenhan and Zhong, Yiran and Ma, Lin and Stenger, Bjorn and Liu, Wei and Li, Hongdong},
	booktitle=CVPR,
	pages={2737--2746},
	year={2020}
}

@inproceedings{Rim2020RealWorldBD,
	title={Real-World Blur Dataset for Learning and Benchmarking Deblurring Algorithms},
	author={Jaesung Rim and Hoon Sung Chwa and Sunghyun Cho},
	booktitle=ECCV,
	year={2020}
}

@inproceedings{Mishchuk2017WorkingHT,
	title={Working hard to know your neighbor's margins: Local descriptor learning loss},
	author={Anastasiya Mishchuk and Dmytro Mishkin and Filip Radenovi{\'c} and Jiri Matas},
	booktitle=NIPS,
    volume={30},
	year={2017}
}

@inproceedings{sun2021loftr,
  title={LoFTR: Detector-free local feature matching with transformers},
  author={Sun, Jiaming and Shen, Zehong and Wang, Yuang and Bao, Hujun and Zhou, Xiaowei},
  booktitle=CVPR,
  pages={8922--8931},
  year={2021}
}

@inproceedings{liu2021mba,
  title={MBA-VO: Motion blur aware visual odometry},
  author={Liu, Peidong and Zuo, Xingxing and Larsson, Viktor and Pollefeys, Marc},
  booktitle=ICCV,
  pages={5550--5559},
  year={2021}
}

@article{wang2025mba,
  title={Mba-slam: Motion blur aware dense visual slam with radiance fields representation},
  author={Wang, Peng and Zhao, Lingzhe and Zhang, Yin and Zhao, Shiyu and Liu, Peidong},
  journal=PAMI,
  year={2025}
}

@article{hynet2020,
  title={HyNet: Learning local descriptor with hybrid similarity measure and triplet loss},
  author={Tian, Yurun and Barroso Laguna, Axel and Ng, Tony and Balntas, Vassileios and Mikolajczyk, Krystian},
  journal=NIPS,
  volume={33},
  pages={7401--7412},
  year={2020}
}

@article{Fischler1981RandomSC,
	title={Random sample consensus: a paradigm for model fitting with applications to image analysis and automated cartography},
	author={Martin Fischler and Robert Bolles},
	journal={Communications of the ACM},
	year=1981,
	volume=24,
	pages={381--395}
}

@inproceedings{zhao2024balf,
  title={Balf: Simple and efficient blur aware local feature detector},
  author={Zhao, Zhenjun},
  booktitle=WACV,
  pages={3362--3372},
  year={2024}
}

@inproceedings{tyszkiewicz2020disk,
  title={DISK: Learning local features with policy gradient},
  author={Tyszkiewicz, Micha{\l} and Fua, Pascal and Trulls, Eduard},
  booktitle=NIPS,
  pages={14254--14265},
  volume={33},
  year={2020}
}

@inproceedings{gleize2023silk,
  title={Silk: Simple learned keypoints},
  author={Gleize, Pierre and Wang, Weiyao and Feiszli, Matt},
  booktitle=ICCV,
  pages={22499--22508},
  year={2023}
}

@inproceedings{edstedt2024dedodev2,
  title={DeDoDe v2: Analyzing and Improving the DeDoDe Keypoint Detector},
  author={Edstedt, Johan and B{\"o}kman, Georg and Zhao, Zhenjun},
  booktitle=CVPRW,
  pages={4245--4253},
  year={2024}
}

@inproceedings{Sattler2018CVPR,
  title={Benchmarking 6dof outdoor visual localization in changing conditions},
  author={Sattler, Torsten and Maddern, Will and Toft, Carl and Torii, Akihiko and Hammarstrand, Lars and Stenborg, Erik and Safari, Daniel and Okutomi, Masatoshi and Pollefeys, Marc and Sivic, Josef and Kahl, Fredrik and Pajdla, Tomas},
  booktitle=CVPR,
  pages={8601--8610},
  year={2018}
}

@inproceedings{lindenberger2023lightglue,
  title={Lightglue: Local feature matching at light speed},
  author={Lindenberger, Philipp and Sarlin, Paul-Edouard and Pollefeys, Marc},
  booktitle=ICCV,
  pages={17581--17592},
  year={2023}
}

@article{Zhao2023ALIKED,
  title={Aliked: A lighter keypoint and descriptor extraction network via deformable transformation},
  author={Zhao, Xiaoming and Wu, Xingming and Chen, Weihai and Chen, Peter CY and Xu, Qingsong and Li, Zhengguo},
  journal=TIM,
  volume={72},
  pages={1--16},
  year={2023}
}

@inproceedings{AffNet2017,
  title={Repeatability is not enough: Learning affine regions via discriminability},
  author={Mishkin, Dmytro and Radenovic, Filip and Matas, Jiri},
  booktitle=ECCV,
  pages={284--300},
  year={2018}
}

@inproceedings{chen2022aspanformer,
  title={ASpanFormer: Detector-Free Image Matching with Adaptive Span Transformer},
  author={Chen, Hongkai and Luo, Zixin and Zhou, Lei and Tian, Yurun and Zhen, Mingmin and Fang, Tian and McKinnon, David and Tsin, Yanghai and Quan, Long},
  booktitle = ECCV,
  pages={20--36},
  year={2022}
}

@article{campos2021orb,
  title={Orb-slam3: An accurate open-source library for visual, visual--inertial, and multimap slam},
  author={Campos, Carlos and Elvira, Richard and Rodr{\'\i}guez, Juan J G{\'o}mez and Montiel, Jos{\'e} MM and Tard{\'o}s, Juan D},
  journal=TRO,
  volume={37},
  number={6},
  pages={1874--1890},
  year={2021}
}

@inproceedings{schonberger2016structure,
  title={Structure-from-motion revisited},
  author={Schonberger, Johannes L and Frahm, Jan-Michael},
  booktitle=CVPR,
  pages={4104--4113},
  year={2016}
}

@article{toft2020long,
  title={Long-term visual localization revisited},
  author={Toft, Carl and Maddern, Will and Torii, Akihiko and Hammarstrand, Lars and Stenborg, Erik and Safari, Daniel and Okutomi, Masatoshi and Pollefeys, Marc and Sivic, Josef and Pajdla, Tomas and Kahl, Fredrik and Sattler, Torsten},
  journal=PAMI,
  volume={44},
  number={4},
  pages={2074--2088},
  year={2020}
}

@inproceedings{sattler2012image,
  title={Image retrieval for image-based localization revisited.},
  author={Sattler, Torsten and Weyand, Tobias and Leibe, Bastian and Kobbelt, Leif},
  booktitle=BMVC,
  volume={1},
  number={2},
  pages={4},
  year={2012}
}

@inproceedings{wang2022matchformer,
  title={MatchFormer: Interleaving Attention in Transformers for Feature Matching},
  author={Wang, Qing and Zhang, Jiaming and Yang, Kailun and Peng, Kunyu and Stiefelhagen, Rainer},
  booktitle=ACCV,
  pages={2746--2762},
  year={2022}
}

@inproceedings{sarlin20superglue,
  title={Superglue: Learning feature matching with graph neural networks},
  author={Sarlin, Paul-Edouard and DeTone, Daniel and Malisiewicz, Tomasz and Rabinovich, Andrew},
  booktitle=CVPR,
  pages={4938--4947},
  year={2020}
}

@article{oquab2023dinov2,
  title={DINOv2: Learning Robust Visual Features without Supervision},
  author={Oquab, Maxime and Darcet, Timothée and Moutakanni, Theo and Vo, Huy V. and Szafraniec, Marc and Khalidov, Vasil and Fernandez, Pierre and Haziza, Daniel and Massa, Francisco and El-Nouby, Alaaeldin and Howes, Russell and Huang, Po-Yao and Xu, Hu and Sharma, Vasu and Li, Shang-Wen and Galuba, Wojciech and Rabbat, Mike and Assran, Mido and Ballas, Nicolas and Synnaeve, Gabriel and Misra, Ishan and Jegou, Herve and Mairal, Julien and Labatut, Patrick and Joulin, Armand and Bojanowski, Piotr},
  journal={arXiv preprint arXiv:2304.07193},
  year={2023}
}

@article{bellavia2024image,
  title={Image matching filtering and refinement by planes and beyond},
  author={Bellavia, Fabio and Zhao, Zhenjun and Morelli, Luca and Remondino, Fabio},
  journal={arXiv preprint arXiv:2411.09484},
  year={2024}
}

@article{zhu2026mygo,
  title={MyGO-Splat: Multi-Objective Closed-Loop Geometric Feedback for RGB-Only Gaussian SLAM},
  author={Zhu, Fan and Chen, Ziyu and Zhao, Zhenjun and Xu, Zhisong and Zhu, Hui and Li, Mingrui and Jiang, Chunmao and Civera, Javier},
  journal={arXiv preprint arXiv:2606.29738},
  year={2026}
}

@inproceedings{Vaswani2017AttentionIA,
  title={Attention is all you need},
  author={Ashish Vaswani and Noam M. Shazeer and Niki Parmar and Jakob Uszkoreit and Llion Jones and Aidan N. Gomez and Lukasz Kaiser and Illia Polosukhin},
  booktitle=NIPS,
  year={2017}
}

@article{gu2026ulf,
  title={ULF-Loc: Unbiased Landmark Feature for Robust Visual Localization with 3D Gaussian Splatting},
  author={Gu, Yingdong and Yan, Shaocheng and Zhao, Zhenjun and Kou, Yuan and Luo, Jianxin and Shi, Pengcheng and Li, Jiayuan},
  journal={arXiv preprint arXiv:2605.04730},
  year={2026}
}

@article{meng2026dream,
  title={Dream-slam: Dreaming the unseen for active slam in dynamic environments},
  author={Meng, Xiangqi and Hou, Pengxu and Zhao, Zhenjun and Civera, Javier and Cremers, Daniel and Wang, Hesheng and Li, Haoang},
  journal={arXiv preprint arXiv:2602.21967},
  year={2026}
}

@inproceedings{pakulev2023ness,
  title={Ness-st: Detecting good and stable keypoints with a neural stability score and the shi-tomasi detector},
  author={Pakulev, Konstantin and Vakhitov, Alexander and Ferrer, Gonzalo},
  booktitle=ICCV,
  pages={9544--9554},
  year={2023}
}

@inproceedings{lee2022self,
  title={Self-supervised equivariant learning for oriented keypoint detection},
  author={Lee, Jongmin and Kim, Byungjin and Cho, Minsu},
  booktitle=CVPR,
  pages={4847--4857},
  year={2022}
}

@article{li2023hong,
  title={Hong kong world: Leveraging structural regularity for line-based slam},
  author={Li, Haoang and Zhao, Ji and Bazin, Jean-Charles and Kim, Pyojin and Joo, Kyungdon and Zhao, Zhenjun and Liu, Yun-Hui},
  journal=PAMI,
  volume={45},
  number={11},
  pages={13035--13053},
  year={2023}
}

@article{nister2004efficient,
  title={An efficient solution to the five-point relative pose problem},
  author={Nist{\'e}r, David},
  journal=PAMI,
  volume={26},
  number={6},
  pages={756--770},
  year={2004}
}

@inproceedings{sarlin2019coarse,
  title={From coarse to fine: Robust hierarchical localization at large scale},
  author={Sarlin, Paul-Edouard and Cadena, Cesar and Siegwart, Roland and Dymczyk, Marcin},
  booktitle=CVPR,
  pages={12708--12717},
  year={2019}
}

@inproceedings{arandjelovic2016netvlad,
  title={NetVLAD: CNN architecture for weakly supervised place recognition},
  author={Arandjelovic, Relja and Gronat, Petr and Torii, Akihiko and Pajdla, Tomas and Sivic, Josef},
  booktitle=CVPR,
  pages={5297--5307},
  year={2016}
}

@inproceedings{edstedt2024roma,
  title={RoMa: Robust dense feature matching},
  author={Edstedt, Johan and Sun, Qiyu and B{\"o}kman, Georg and Wadenb{\"a}ck, M{\aa}rten and Felsberg, Michael},
  booktitle=CVPR,
  pages={19790--19800},
  year={2024}
}

@article{edstedt2025roma,
  title={RoMa v2: Harder Better Faster Denser Feature Matching},
  author={Edstedt, Johan and Nordstr{\"o}m, David and Zhang, Yushan and B{\"o}kman, Georg and Astermark, Jonathan and Larsson, Viktor and Heyden, Anders and Kahl, Fredrik and Wadenb{\"a}ck, M{\aa}rten and Felsberg, Michael},
  journal={arXiv preprint arXiv:2511.15706},
  year={2025}
}

@article{zhao2026advances,
  title={Advances in global solvers for 3d vision},
  author={Zhao, Zhenjun and Yang, Heng and Liao, Bangyan and Zeng, Yingping and Yan, Shaocheng and Gu, Yingdong and Liu, Peidong and Zhou, Yi and Li, Haoang and Civera, Javier},
  journal={arXiv preprint arXiv:2602.14662},
  year={2026}
}

@inproceedings{potje2024cvpr,
  author={Potje, Guilherme and Cadar, Felipe and Araujo, André and Martins, Renato and Nascimento, Erickson R.},
  title={XFeat: Accelerated Features for Lightweight Image Matching},
  booktitle=CVPR,
  year={2024},
  pages={2682-2691}
}

@inproceedings{li2025slam,
  title={Slam-x: Generalizable dynamic removal for nerf and gaussian splatting slam},
  author={Li, Mingrui and Li, Dong and Hu, Sijia and Wang, Kangxu and Zhao, Zhenjun and Wang, Hongyu},
  booktitle={Proceedings of the 33rd ACM International Conference on Multimedia},
  pages={1132--1140},
  year={2025}
}

@inproceedings{beaudet1978rotational,
  title={Rotational invariant image operators},
  author={Beaudet, Paul R},
  booktitle=ICPR,
  pages={579--583},
  year={1978}
}

\end{document}